\documentclass[acmsmall]{acmart}

\AtBeginDocument{%
  \providecommand\BibTeX{{%
    \normalfont B\kern-0.5em{\scshape i\kern-0.25em b}\kern-0.8em\TeX}}}
\setcopyright{acmcopyright}
\copyrightyear{2018}
\acmYear{2018}
\acmDOI{XXXXXXX.XXXXXXX}
\acmJournal{JACM}
\acmVolume{37}
\acmNumber{4}
\acmArticle{111}
\acmMonth{8}

\usepackage{xspace}
\newcommand{\ie}{\emph{i.e.}\xspace} 
\newcommand{\etc}{\emph{etc.}\xspace} 
\newcommand{\eg}{\emph{e.g.}\xspace} 

\newtheorem{prop}{Proposition}
\newtheorem{Definition}{Definition}

\usepackage{threeparttable}
\usepackage{multirow}
\usepackage{booktabs}

\usepackage{tikz}
\usepackage[edges]{forest}

\usepackage{amsmath}

\usepackage[ruled, vlined, linesnumbered]{algorithm2e}

\begin{document}

\title{Causal Discovery from Temporal Data: An Overview and New Perspectives}

\author{Chang Gong}
\affiliation{%
  \institution{Institute of Computing Technology, Chinese Academy of Sciences}
  \city{Beijing}
  \country{China}}
\email{gongchang21z@ict.ac.cn}

\author{Di Yao}
\affiliation{%
  \institution{Institute of Computing Technology, Chinese Academy of Sciences}
  \city{Beijing}
  \country{China}}
\email{yaodi@ict.ac.cn}

\author{Chuzhe Zhang}
\affiliation{%
  \institution{School of Mathematical Sciences, Fudan University}
  \city{Shanghai}
  \country{China}}
\email{felixzh01@outlook.com}

\author{Wenbin Li}
\affiliation{%
  \institution{Institute of Computing Technology, Chinese Academy of Sciences}
  \city{Beijing}
  \country{China}}
\email{liwenbin20z@ict.ac.cn}

\author{Jingping Bi}
\affiliation{%
  \institution{Institute of Computing Technology, Chinese Academy of Sciences}
  \city{Beijing}
  \country{China}}
\email{bjp@ict.ac.cn}

\begin{abstract}
  Temporal data, representing chronological observations of complex systems, has always been a typical data structure that can be widely generated by many domains, such as industry, medicine and finance. Analyzing this type of data is extremely valuable for various applications. Thus, different temporal data analysis tasks, \eg, classification, clustering and prediction, have been proposed in the past decades. Among them, causal discovery, learning the causal relations from temporal data, is considered an interesting yet critical task and has attracted much research attention. Existing causal discovery works can be divided into two highly correlated categories according to whether the temporal data is calibrated, \ie, multivariate time series causal discovery, and event sequence causal discovery. However, most previous surveys are only focused on the time series causal discovery and ignore the second category. In this paper, we specify the correlation between the two categories and provide a systematical overview of existing solutions. Furthermore, we provide public datasets, evaluation metrics and new perspectives for temporal data causal discovery. 
\end{abstract}




\received{20 February 2007}
\received[revised]{12 March 2009}
\received[accepted]{5 June 2009}

\maketitle

\section{Introduction}

Temporal data recording the status changing of complex systems is widely collected by different application domains, such as social networks, bioinformatics, neuroscience and finance, \etc. As one of the most popular data structural, temporal data consists of attribute sequences ordered by time. Owing to the rapid development of sensors and computing devices, research works on temporal data analysis are emerging in recent years. Different approaches have been proposed for different tasks such as classification\cite{ismail2019deep, ratanamahatana2004making}, clustering\cite{aghabozorgi2015time, liao2005clustering}, prediction\cite{weigend2018time}, causal discovery\cite{intro/ts_surveys/bivariate_comparative_edinburgh2021causality, intro/ts_surveys/bivariate_comparative_PRE_krakovska2018comparison}, \etc. 

Among these tasks, causal discovery recognizing the causal relations between many temporal components has become a challenging yet critical task for temporal data analysis. The learned causal structures could be beneficial for explaining the data generation process and guiding the design of data analysis methods. According to whether the data is calibrated, the temporal data for causal discovery can be categorized into two groups, \ie, multivariate time series (MTS) and event sequences. Therefore, existing causal discovery methods can also be divided into two groups respectively. In this survey, we aim to provide a thoughtful overview and summarize the frontiers of temporal data causal discovery.

MTS data, describing the calibrated states of multiple variables changing over time, is a general kind of temporal data in many domains. Discovering causal relations from MTS could be beneficial to the explainability and robustness of data analysis models. However, the definitions of causal relations are not unique, leading to different solutions.
Accordingly, existing works can be grouped into four categories, \ie, constraint-based methods, score-based methods, functional causal model (FCM)-based methods and Granger causality-based methods. Besides, there also exist some perspectives such as Takens' causality and differential equations. In this paper, we will specify the main idea and recent advances for each category.

Another task discussed in this survey is the causal discovery from event sequences, which infers causal relationships within irregularly and asynchronously observed time series. Specifically, it takes a sequence of different events as the input and outputs a causal graph representing the causal interactions between different events. This task is of great importance since most real-world events cannot emerge within a fixed time interval. In accordance with the MTS task, we classify the corresponding methods into three main categories: constraint-based, score-based, and Granger causality-based methods. Among these three categories, Granger causality-based methods, especially Granger causality-based Hawkes process models, are well-developed since a natural match-up exists between Granger causality and Hawkes processes. We will further describe these approaches in detail within this review.




\renewcommand{\thefootnote}{\fnsymbol{footnote}}

\begin{table*}[h]
    \tiny 
    \caption{Highlights of existing reviews on causal discovery.}
    \label{tab:surveys_overview}
    \centering
    \begin{threeparttable} 
    \resizebox{\textwidth}{14mm}{ 
        \begin{tabular}{c|ccccc|c|l}
           \toprule
           \multirow{2}{*}{Reviews} & \multicolumn{5}{c|}{Multivariate Time-series} & \multirow{2}{*}{Event Sequence} & \multirow{2}{*}{Highlights} \\
            & Constrain-based & Score-based & FCM-based & Granger & Deep Learning & & \\
    
           \midrule
            \cite{intro/nonts_surveys/glymour2019review} & No\footnotemark[1] & No\footnotemark[1] & No\footnotemark[1] & Yes & No & No & An overview for causal discovery methods with practical issues and insightful guidelines \\  
            \cite{intro/surveys_guoruocheng/csur/GuoCLH020} & No\footnotemark[1] & No\footnotemark[1] & No\footnotemark[1] & No & No\footnotemark[1] & No & Causal discovery methods dealing with big data (high-dimensional, mixed data) are reviewed   \\   
            \cite{intro/nonts_surveys/Dya21} & No\footnotemark[1]& No\footnotemark[1] & No\footnotemark[1] & Yes & No\footnotemark[1] & No & A more extensive coverage of continuous optimization approaches compared to other surveys \\  
            \cite{intro/nonts_surveys/XJTU22} & No\footnotemark[1] & No\footnotemark[1] & No\footnotemark[1] & No & No\footnotemark[1] & No & A wider concept of deep learning causal discovery methods is introduced \\  
            \cline{1-8}
             \cite{intro/ts_surveys/Moraffah21} & Yes & No & Yes & Yes & No & No & The first survey covers the current progress to analyze time series from a causal perspective  \\ 
            \cite{intro/ts_surveys/AliGranger21} & No & No & No & Yes & Yes & No\footnotemark[2] & Recent advances including network-form and more general notions of Granger causality  \\
            \cite{intro/ts_surveys/AssaadDG22} & Yes & Yes& Yes& Yes& No& No & A recent and comprehensive review for causal discovery in time series with comparative evaluations \\
            Ours & Yes & Yes & Yes& Yes& Yes& Yes&  A systematic review of causal discovery in both MTS and event sequence, with new perspectives \\
           \bottomrule
        \end{tabular}
    }
    \end{threeparttable}  
\end{table*}

Recently, many surveys \cite{intro/nonts_surveys/glymour2019review, intro/surveys_guoruocheng/csur/GuoCLH020, intro/nonts_surveys/Dya21, intro/nonts_surveys/XJTU22, intro/ts_surveys/Moraffah21, intro/ts_surveys/AliGranger21, intro/ts_surveys/AssaadDG22, intro/nonts_surveys/BN21, intro/DL_surveys/corr/abs-2211-03374, intro/surveys/CSL/heinze2018causal} have been published to summarize the progress of causal discovery. 
We compared the representative reviews and their highlight points in Table \ref{tab:surveys_overview}. As shown, these surveys fall into two lines. Research works in the first line \cite{intro/nonts_surveys/glymour2019review, intro/nonts_surveys/Dya21, intro/surveys_guoruocheng/csur/GuoCLH020, intro/nonts_surveys/XJTU22} discuss the general causal discovery problem in different perspectives. For example,
\cite{intro/nonts_surveys/glymour2019review} provide a brief review of the computational causal discovery methods. 
\cite{intro/nonts_surveys/Dya21} focus on the flurry developments of continuous optimization approaches. 
To handle big data, both causal inference and causal discovery methods based on machine learning are introduced in \cite{intro/surveys_guoruocheng/csur/GuoCLH020}. 
Moreover, deep learning causal discovery methods are reviewed in different variable paradigms \cite{intro/nonts_surveys/XJTU22}, where the causal relations in data are discussed from a broader perspective. In these papers, temporal data was taken as one special application and many data-specified methods are not included.
The surveys in the second line focus on temporal data causal discovery. As illustrated in Table \ref{tab:surveys_overview}, causal discovery methods for bivariate time series are reviewed in \cite{intro/ts_surveys/bivariate_comparative_edinburgh2021causality, intro/ts_surveys/bivariate_comparative_PRE_krakovska2018comparison}. 
The approaches for causal inference in time series are recently reviewed in \cite{intro/ts_surveys/Moraffah21, intro/ts_surveys/AliGranger21}.
The recent work \cite{intro/ts_surveys/AssaadDG22} discusses and comparatively evaluates the existing solutions of time series causal discovery. Nevertheless, causal discovery methods for event sequences are ignored in these reviews.  In this paper, we not only provide a thoughtful overview of causal discovery methods of the two kinds of temporal data but also give an analysis of the connections and differences between them.

Next, we first introduce the background and preliminary of the causal discovery problem in Section \ref{sec:prem}. The recent progress of causal discovery from MTS and event sequences are specified in Section \ref{sec:mts} and Section~\ref{sec:event} respectively. After that, we provide an overview of the applications of temporal data causal discovery in Section \ref{sec:app} and summarize the available resources in Section \ref{sec:res}. At last, we discuss the limitations and new perspectives of recent temporal data causal discovery methods in Section \ref{sec:discuss}.
The whole framework of this survey is shown in Figure~\ref{fig:frame}.

\footnotetext[1]{Entries correspond to methods reviewed which are mainly for non-temporal settings.}
\footnotetext[2]{Mainly about causalities related to the Hawkes process. }

\renewcommand{\thefootnote}{\arabic{footnote}}

\tikzset{
    parent/.style={align=center,text width=3cm,rounded corners=3pt, 
    },
    child/.style={
        align=center,text width=3cm,rounded corners=3pt,
        }}

\colorlet{col1}{white}
\colorlet{col2}{blue!15}
\colorlet{col3}{red!15}
\colorlet{col4}{green!15}
\colorlet{col5}{yellow!15}
\colorlet{colline1}{orange!60}
\colorlet{colline2}{blue!60}
\colorlet{colline3}{brown!60}
\colorlet{colline4}{green!60}
\colorlet{colline5}{gray!60}

\begin{figure*}[!b]
\resizebox*{.98\linewidth}{!}{
\begin{forest}
        for tree={
            grow'=east,
            anchor=west,
            node options={draw, thick, font=\sffamily, align=center, },
            edge={semithick},
            forked edges,
            l sep=8mm, 
            s sep=8mm, 
            text width=2.3cm,
            fork sep = 2mm,           
        },
        [\textbf{Temporal Causal Discovery}, fill=col1, parent, rotate=90,font=\LARGE, 
        for tree={s sep=2.0mm},
        [{\textbf{MTS\\Causal Discovery} \\ \emph{(\S~\ref{sec:mts}, Table~\ref{tab:ts_category_overview})}}, font=\large,
            for tree={child, fill=col3, text width = 4.0cm}, 
            [\textbf{Constraint-Based Approaches} \\\emph{(\S~\ref{subsection:CB})}, text width = 3.6cm,
                [With Causal Sufficiency,fill = col3,text width = 2.8cm
                    [
                    {oCSE~\cite{MTS/CB/oCSE/siamads/0007TB15}, PCGCE~\cite{MTS/others/information_criterion/uai/AssaadDG22}, PCMCI~\cite{MTS/CB/PCMCI_runge2019detecting, MTS/CB/insPCMCI_Runge20}},
                    text width=5.5 cm,draw=colline1,line width=1.2pt,fill=col1,]
                ]
                [Without Causal Sufficiency,fill = col3,text width = 2.8cm
                    [
                    {ANLTSM~\cite{MTS/CB/jmlr_ChuG08}, tsFCI~\cite{MTS/CB/FCI_tsFCI_entner2010causal}, SVAR-FCI~\cite{MTS/CB_FCI_SVAR_FCI_MalinskyS18}, LPCMCI~\cite{MTS/CB/LPCMCI_GerhardusR20}},
                    text width=5.5 cm,draw=colline1,line width=1.2pt,fill=col1,]
                ]
            ]
            [\textbf{Score-Based Approaches} \\\emph{(\S~\ref{subsection:SB})},  text width = 3.6cm,
                for tree = {child, fill = col3,text width = 3.6 cm}
                [Combinatorial Search,fill = col3,text width = 2.8cm
                    [
                    {Structural EM~\cite{MTS/SB/learnDBN/uai/FriedmanMR98}, Greedy Hill-climbling Search~\cite{MTS/SB/CV_DBN_prl/PenaBT05}, Structural Constraints~\cite{MTS/SB/learnDBN/jmlr/CamposJ11}, etc.},
                    text width=5.5 cm,draw=colline1,line width=1.2pt,fill=col1,]
                ]
                [Continuous Optimization,fill = col3,text width = 2.8cm
                    [
                    {DYNOTEARS~\cite{MTS/SB/Dynotears_aistats_PamfilSDPGBA20}, NTS-NOTEARS~\cite{MTS/SB/NTS_NOTEARS}, IDYNO~\cite{MTS/SB/Dynotears_interventionalDATA/icml/GaoBNLY22}},
                    text width=5.5 cm,draw=colline1,line width=1.2pt,fill=col1,]
                ]
            ]
            [\textbf{FCM-Based Approaches} \\\emph{(\S~\ref{subsection:FCM})},
                for tree = {child, fill = col3,text width = 3.6 cm}
                [Independent Component Analysis,fill = col3,text width = 2.8cm
                    [
                    {VAR-LiNGAM~\cite{MTS/FCM/VAR_LINGAM_icml_HyvarinenSH08, MTS/FCM/VAR_LINGAM_v2_jmlr_HyvarinenZSH10}, MCD~\cite{MTS/FCM/ijcai2013/SchaechtleSB13}, NCDH~\cite{MTS/FCM/cikm_WuWWLC22}},
                    text width=5.5 cm,draw=colline1,line width=1.2pt,fill=col1,]
                ]
                [Additive Noise Model,fill = col3,text width = 2.8cm
                    [
                    {TiMINo~\cite{MTS/FCM/nips_PetersJS13},  NBCB~\cite{MTS/FCM_maybe/pkdd/AssaadDGA21}},
                    text width=5.5 cm,draw=colline1,line width=1.2pt,fill=col1,]
                ]
            ]
            [\textbf{Granger Causality \\Based Approaches} \\\emph{(\S~\ref{subsection:Granger_base})}, 
                for tree = {child, fill = col3,text width = 3.6cm}
                [{HSIC-Lasso-GC~\cite{MTS/Granger/Kernel_ren2020novel}, (R)NN-GC~\cite{MTS/Granger/NN_GC_MontaltoSFTPM15, MTS/Granger/RNN_GC_WangLQLFWP18}, MPIR~\cite{MTS/Granger/MPIR_ICML19TS_workshop}, NGC~\cite{MTS/Granger/pamiNGC22}, eSRU~\cite{MTS/Granger/iclr20_esru}, SCGL~\cite{MTS/Granger/SCGL_CIKM_XuHY19}, GVAR~\cite{MTS/Granger/iclr21_GVAR_MarcinkevicsV}, TCDF~\cite{MTS/Attention/TCDF_NautaBS19}, CR-VAE~\cite{MTS/Granger/CR_VAE2023}, InGRA~\cite{MTS/Attention/icdm_InGRA_ChuWMJZY20}, ACD~\cite{Discussion/NewForm/ACD_LoweMSW22}, etc.},
                text width=9.4 cm,draw=colline1 ,line width=1.0pt,fill=col1,
                ]
            ]
            [\textbf{Others} \\\emph{(\S~\ref{subsection:MTS_others})}, 
                for tree = {child, fill = col3,text width = 3.6cm}
                [{Information-theoretic Statistics~\cite{TE_origin/schreiber2000measuring, MTS/others/information/PRE_runge2012quantifying, MTS/Others/concepts/causationEntropy/sun2014causation}, Differential Equation Based Methods~\cite{MTS/others/difference_based/uai/VoortmanDD10, MTS/Others/NGM_neuralode_iclr_BellotBS22}, Nonlinear State-space Methods~\cite{MTS/CCM/work2_main_science_sugihara2012detecting}, Logic-based Methods~\cite{MTS/logic/uai/KleinbergM09}, Hybrid Methods~\cite{MTS/SB/MMHO_DBN/tcbb/LiCZN16}, etc.},
                text width=9.4 cm,draw=colline1 ,line width=1.0pt,fill=col1,
                ]
            ]
        ]
        [\textbf{Event Sequence Causal Discovery} \\\emph{(\S~\ref{sec:event})}, font=\large,  for tree={child, fill=col2, text width = 4.0cm} 
            [\textbf{Multivariate Point Process} \\\emph{(\S~\ref{subsec:MPP})},
                for tree = {child, fill = col2, text width=3.6cm,}
                [{Basics: Intensity Function, Log-likelihood},
                text width=9.4 cm,draw=colline1 ,line width=1.0pt,fill=col1,
                ]
            ]
            [\textbf{Granger Causality\\Based Approaches} \\\emph{(\S~\ref{subsec:event_Granger})},
                for tree = {child, fill = col2,text width=3.6cm}
                [GLM Point Process, text width=2.8cm,
                    [
                    {GLM Model~\cite{10.1371/journal.pcbi.1001110}},
                    text width=5.5cm,draw=colline2,line width=1.2pt,fill=col1]
                    ]
                [Hawkes Process, text width=2.8cm,
                    [{MLE-SGLP~\cite{pmlr-v48-xuc16}, THP~\cite{Cai2021THPTH}, $L_0$Hawkes~\cite{NEURIPS2021_15cf7646}, HGEM~\cite{pmlr-v138-yu20a}, NPHC~\cite{pmlr-v70-achab17a}, GC-nsHP~\cite{CHEN202222}, MDLH~\cite{Schindler_Plant_2022}, etc.},
                    text width=5.5cm,draw=colline2,line width=1.0pt,fill=col1]
                ]
                [Wold Process, text width=2.8cm,
                    [{Granger-Busca~\cite{NEURIPS2018_aff0a6a4}, VI-MWP~\cite{pmlr-v130-etesami21a}},
                    text width=5.5cm,draw=colline2,line width=1.0pt,fill=col1]
                ]
                [Neural Point Process, text width=2.8cm,
                    [{CAUSE~\cite{pmlr-v119-zhang20v}},
                    text width=5.5cm,draw=colline2,line width=1.0pt,fill=col1]
                ]
            ]
            [\textbf{Others} \\\emph{(\S~\ref{subsec43})},
                for tree = {child, fill = col2,text width=3.6cm}
                [Constraint-Based Approaches, text width=2.8cm,
                    [
                    {MMP-LR/NI~\cite{Bhattacharjya2022ProcessIT}, CA~\cite{article}},
                    text width=5.5cm,draw=colline2,line width=1.2pt,fill=col1]
                    ]
                [Score-Based Approaches, text width=2.8cm,
                    [{PGEM~\cite{NEURIPS2018_f1ababf1}},
                    text width=5.5cm,draw=colline2,line width=1.0pt,fill=col1]
                ]
            ]
        ]
        [{\textbf{Applications} \\\emph{(\S~\ref{sec:app}, Table~\ref{tab:application_overview})}},font=\large, for tree={child, fill=col5, text width = 4.0cm}, 
            [\textbf{Scientific Endeavors}, 
            for tree = {child, fill = col5,text width = 3.6cm},
                [
                {Earth Science, Neuroscience, Bioinformatics, etc.},
                 text width=9.4cm,draw=colline3,line width=1.0pt,fill=col1]
                ]
            [\textbf{Industrial\\Implementations}, 
                for tree = {child, fill = col5,text width = 3.6cm},
                [{Anomaly Detection, Root Cause Analysis, Business Intelligence in Online Systems, Video Analysis, Urban Data Analysis, Clinical Data Analysis, etc.},
                 text width=9.4cm,draw=colline3,line width=1.0pt,fill=col1]
            ]
        ]
        [{\textbf{Discussions}\ \&\\\textbf{New Perspectives} \\ \emph{(\S~\ref{sec:discuss})} }, font=\large, for tree={child, fill=col4, text width = 4.0cm}, 
            [\textbf{Challenges \& Practical Considerations} \\\emph{(\S~\ref{subsec:challenges})},
                for tree = {child, fill = col4, text width=3.6cm,}
                [{1) Non-stationarity, 2) Heterogeneity, 3) Unobserved Confounders, 4) Subsampling, 5) Expert Knowledge},
                text width=9.4 cm,draw=colline4 ,line width=1.0pt,fill=col1,
                ]
            ]
            [\textbf{New Perspectives}  \\\emph{(\S~\ref{subsec:newper})},
                for tree = {child, fill = col4,text width = 3.6cm}
                [{1) Amortized Paradigm, 2) Supervised Paradigm, 3) Causal Representation Learning},
                 text width=9.4cm,draw=colline4,line width=1.0pt,fill=col1]
            ]
        ]
        ]
        \end{forest}
} 
\caption{Framework for causal discovery from temporal data. }
\label{fig:frame}
\end{figure*}
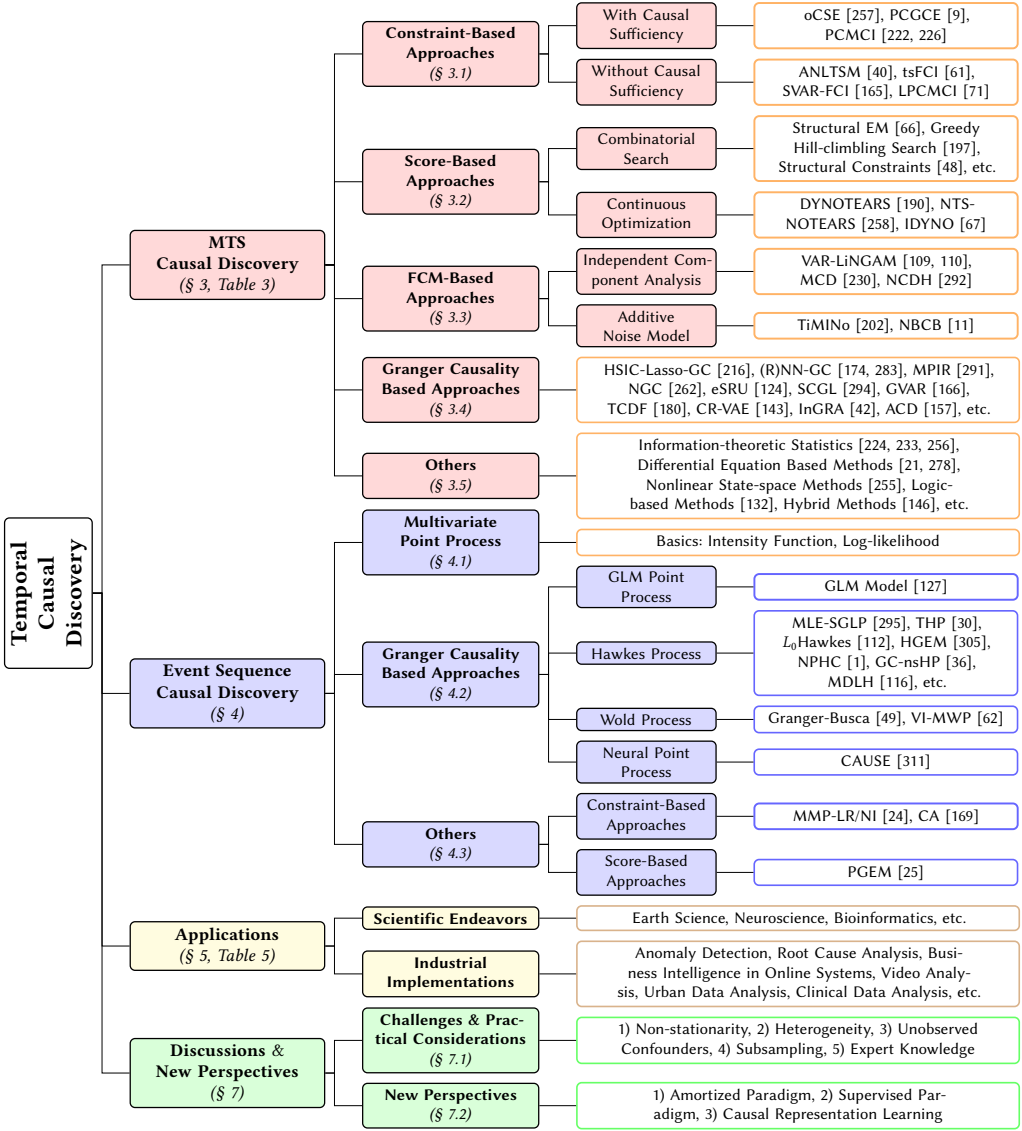

\section{Background \& preliminaries}\label{sec:prem}

This section begins with the definition of key concepts and assumptions in causal discovery, followed by an overview of three causal graph representations applicable to temporal data. Finally, the problem definitions for causal discovery from MTS and event sequences will be presented.

\begin{table}[h]
    \caption{Main notations used in this survey.}
    \label{tab:overall_notation}
    \centering
    \begin{tabular}{cc}
       \toprule
       Notation & Description \\
       \midrule
        $d, E$ & number of time-series variate, and of event types, respectively \\
        $x_i^t$ & the $i$-th time series at time $t$ in multivariate time series \\
        $N_e(t)$ & the number of the event $e$ occurrences before time $t$ \\
        $\perp \!\!\! \perp, \not \! \perp \!\!\! \perp$ & independent, and not independent\\
        $V,U$ & the set of endogenous variables, and of exogenous variables, respectively  \\
        $\mathcal{G}$ & causal graph \\
        $Pa(x_i)$ & the parent nodes of $x_i$ \\  
       \bottomrule
    \end{tabular}
\end{table}

\subsection{Key concepts and assumptions in causal discovery}
\label{subsection:key_concepts}
Some key concepts serve as the foundation for inferring causal relationships from temporal data. We establish this common ground before discussing research works. Afterward, we present formal definitions for the structural causal model, $d$-separation, causal Markov condition, causal identifiability and causal minimality with notations detailed in Table \ref{tab:overall_notation}.

\begin{figure*}
    \centering
	\includegraphics[width=1.0\textwidth]{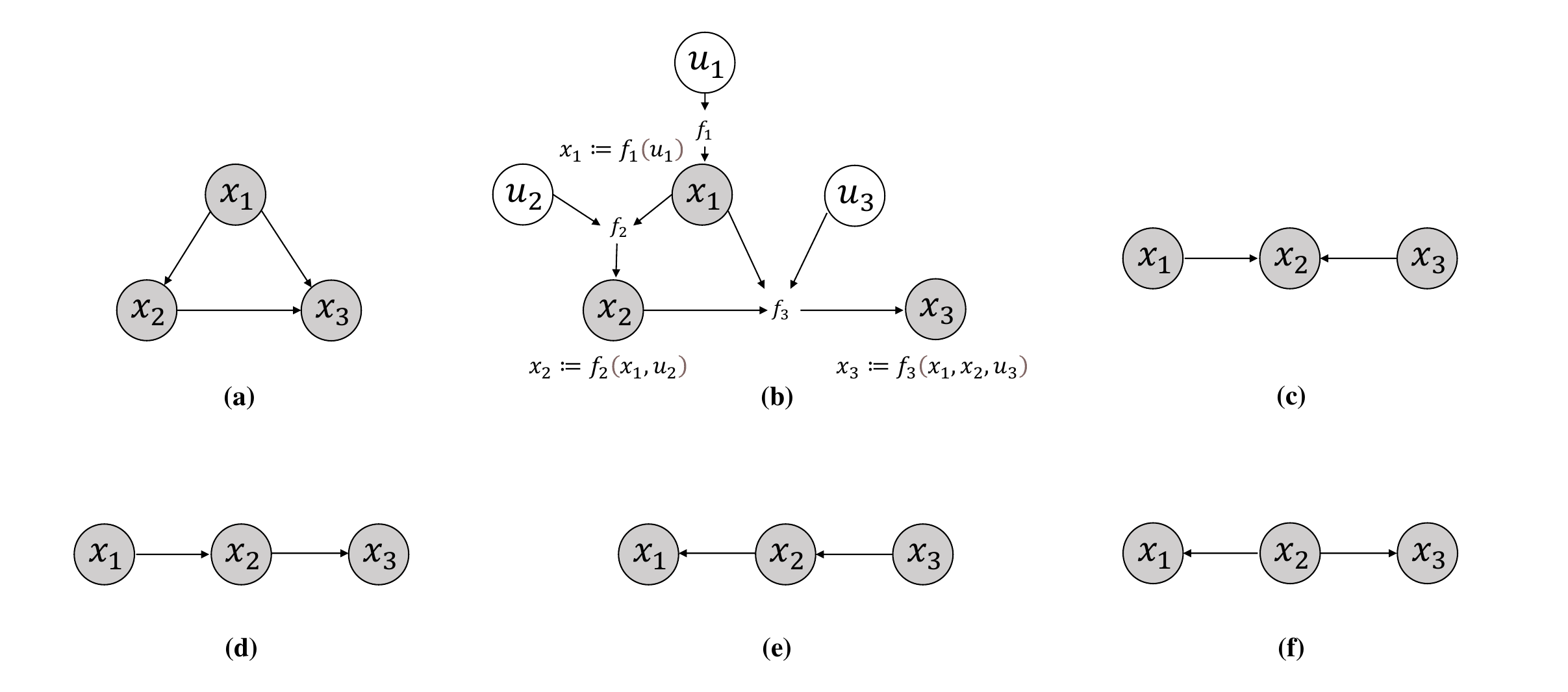}
	\caption{Basic DAGs and a simple structural causal model.}
	\label{fig:6figs}
\end{figure*}

\textbf{Structural Causal Model (SCM).} Pearl's comprehensive theory of causality, as presented in \cite{pearl2009causality}, enables us to draw causal conclusions from observations using causal hierarchy (PCH) \cite{General/book_of_why_pearl2018book}. From that, the structural causal model is defined as a graphical representation of causal relationships that captures how interventions on one or more variables affect the values of other variables in the data generation mechanism.
Formally, SCM can be represented in a 4-tuple $<V,U,F,P(U)>$, where $V,U$ denote the set of endogenous and exogenous variables respectively, $P(U)$ is the distribution of exogenous variables, and $F$ represents the set of the mapping function.
Specifically, for $f_i \in F$, the model $x_i := f_i(Pa(x_i), u_i), i=1,...,d$ indicates the assignment of the value $x_i$ to a function of its structural parents $Pa(x_i)$ and exogenous variable $u_i$.  
For each SCM, we can yield a causal graph DAG $G$ by adding one vertex for each $x_i$ and directing edges from each parent variable in $Pa(x_i)$ (the causes) to child $x_i$ (the effect).
The relationship of the SCM and the corresponding DAG is shown in figure \ref{fig:6figs} (a)(b).

\textbf{\textit{d}-separation.} $d$-separation is a criterion for determining the absence of causal effects between two sets of variables in a graphical model. Two sets of variables are said to be $d$-separated if every path between them is blocked. 
In formal, a set of variables $\mathbf{S}$ \textit{d}-separates two variables if $\mathbf{S}$ blocks all paths between them. 
For the given causal graph in figure \ref{fig:6figs} (d)(e)(f), two vertices $x_1,x_3$ are \textit{d}-separated by the set of vertices $\mathbf{S}$ if $x_2 \in \mathbf{S}$. 
As for the relations in figure \ref{fig:6figs} (c) (a.k.a., a \textit{v-structure} or \textit{collider}), $x_1,x_3$ are also \textit{d}-separated if $x_2$ and none of the descendants of $x_2$ are in set $\mathbf{S}$.

$d$-separation is a fundamental concept in causal discovery because it provides a criterion for determining whether two sets of variables are causally related. If two sets of variables are $d$-separated, then there is no direct or indirect causal effect between them, and they can be considered independent given the observed variables. Conversely, if two sets of variables are not $d$-separated, then there may be a direct or indirect causal effect between them that needs to be accounted for when inferring causal relationships from data. Thus, $d$-separation is an essential tool for identifying causal relationships in graphical models.

\textbf{Causal Markov Condition.} In the causal graph of SCM, each variable is independent of its non-effects given its direct causes \cite{pearl2009causality}. In other words, a variable is conditionally independent of its non-effects (i.e., variables that do not directly cause it) given its parents (i.e., variables that directly cause it). This condition plays an essential role in causal inference. It enables the identification of causal effects from non-experimental data. Formally, the causal Markov condition implies the joint distribution can be factorized according to the following decomposition:
\begin{equation}
    P(\mathbf{x}) = \prod_i^d P(x_i | Pa(x_i)) \nonumber
\end{equation}

\textbf{Markov Equivalence Class(MEC).} Two graphical models belong to the same MEC if they entail the same set of conditional independence relations among the observed variables, regardless of the specific structure of the graph. 
For example, the causal diagrams in Figure \ref{fig:6figs} (d)(e)(f) imply the same \textit{d}-seperation information $x_1 \perp \!\!\! \perp  x_3 | x_2$ and belong to the same MEC. MEC is important because it enables us to identify the minimal set of conditions necessary for inferring causal relationships from non-experimental data.

\textbf{Causal Identifiability.}
A causal effect is identifiable if it can be estimated without making any untestable assumptions or invoking additional information beyond the observed variables. This means that all causal graphs in the same MEC represent equivalent causal structures from an observational viewpoint. In general, causal identifiability requires that the causal graph is acyclic and that all backdoor paths between the treatment and outcome variables are blocked. If these conditions are met, the causal effect can be identified using the $do$-calculus or other causal inference techniques. Thus, the prerequisite of causal discovery is that causal relationships can be identifiable.

\textbf{Causal Minimality.} Consider a DAG $\mathcal{G}$ and a probability distribution $P$, $P$ is said to satisfy the causal minimality with respect to  $\mathcal{G}$ if $P$ is Markovian with respect to $\mathcal{G}$ but not to any proper subgraph of $\mathcal{G}$. It indicates that all the variables are necessary and sufficient to accurately represent the causal relationships while excluding any variables that do not contribute to the causal mechanism. A distribution is minimal with respect to the causal graph if and only if there is no node that is conditionally independent of any of its parents, given the remaining parents. In other words, all the parents are “active”.

Building on the aforementioned concepts, we introduce three assumptions, causal sufficiency, faithfulness, and temporal priority, which are the untestable foundations of causal discovery.

\textbf{Causal Sufficiency.}
A set of variables is causally sufficient if all common causes of all variables are observed \cite{MTS/CB/PC_origin}. This assumption indicates that the causal graph in SCM can reflect the truth data generation process and there is no hidden confounder. Under the assumption of causal sufficiency, the majority of causal discovery algorithms presume that the causal structure can be depicted as a DAG. 

\textbf{Faithfulness.}
Faithfulness asserts that all conditional independence relations of $P$ that hold in the observed data are entailed by the causal model $\mathcal{G}$, and conversely, all conditional independence relations implied by the causal model are also held in the observed data. Note that faithfulness implies causal minimality. If $P$ is faithful and Markovian with respect to $\mathcal{G}$, then the causal minimality is satisfied.

\begin{figure*}
    \centering
	\includegraphics[width=0.7\textwidth]{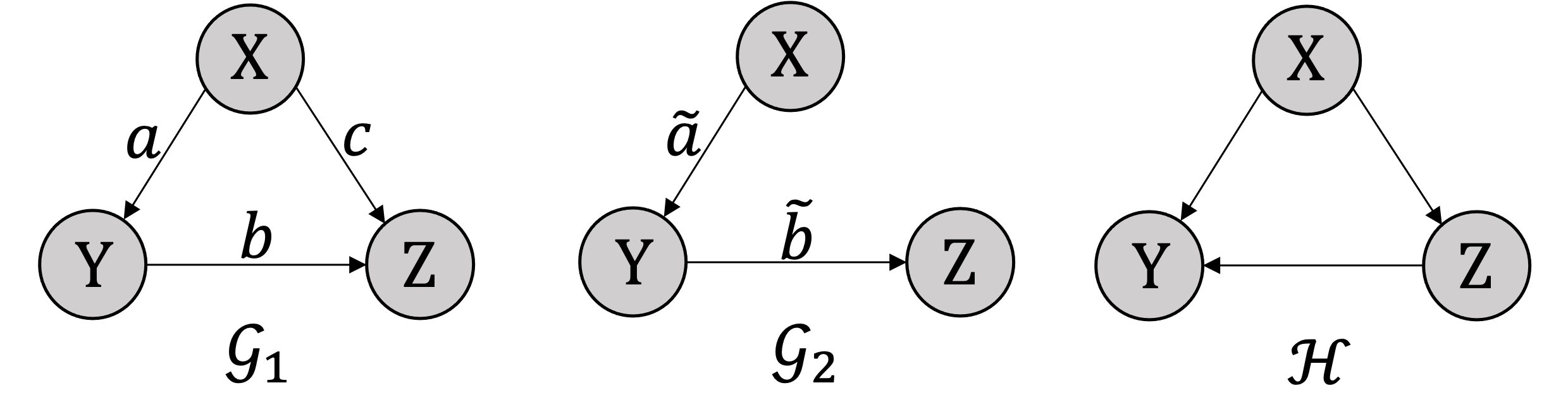}
	\caption{Example of the faithfulness and causal minimality.}
	\label{fig:faithfulness}
\end{figure*}

Intuitively, faithfulness is not easy to understand. We try to clarify it with an example \cite{peters2017elements}. As shown in Figure \ref{fig:faithfulness}, we assume the generation process of $\mathcal{G}_1$ as a linear Gaussian SCM:
\begin{align*}
    X &:= N_X\\
    Y &:= aX + N_Y\\
    Z &:= bY + cX + N_Z
\end{align*}
The noise variables $N_X \sim \mathcal{N}(0, \sigma_x^2)$, $N_Y \sim \mathcal{N}(0, \sigma_y^2)$ and $N_Z \sim \mathcal{N}(0, \sigma_z^2)$ are jointly independent. Let us consider a special case that $a \cdot b + c = 0$. In this setting, the variables $X$ and $Z$ are independent. The direction of $Y \rightarrow Z$ would be inverted and the causal model $\mathcal{G}_1$ is degraded to $\mathcal{G}_2$. According to the definition, $\mathcal{G}_1$ and $\mathcal{G}_2$ satisfy the causal minimality. But the faithfulness is violated in this special case, \ie $\mathcal{G}_2$ is not a proper subgraph of $\mathcal{G}_1$.  Thus, the probability of this linear Gaussian model is not faithfulness with respective to $\mathcal{G}_1$. Although $\mathcal{G}_2$ is a proper subgraph of $\mathcal{H}$, the distribution does not satisfy causal minimality because the probability is not Markovian with respect to $\mathcal{H}$.

While faithfulness is untestable in practice, it is crucial for deriving valid causal inferences from data because it ensures that the model correctly represents the data-generating mechanisms. If this assumption is violated, the causal relationships are uncertain which is a disaster for causal discovery methods~\cite{MTS/CB/PC_origin}.

\textbf{Temporal priority.} For two variables, temporal priority means that the cause must have occurred before its effect. It is a foundation assumption of causal discovery from temporal data and creates an asymmetric time relationship in causal processes. The temporal priority helps us to establish the direction of a causal relationship when two variables are causally linked. However, if the sampling frequencies of time series are high, the time difference between events associated with the time series may be indistinguishable. In such cases, two events that occurred at different times could be perceived as instantaneous in the observational time series, leading to contemporaneous causal relationships between causes and effects occurring at different time instants.

\subsection{Causal Structure for Temporal Data}

\begin{figure}
    \centering
	\includegraphics[width=0.9\textwidth]{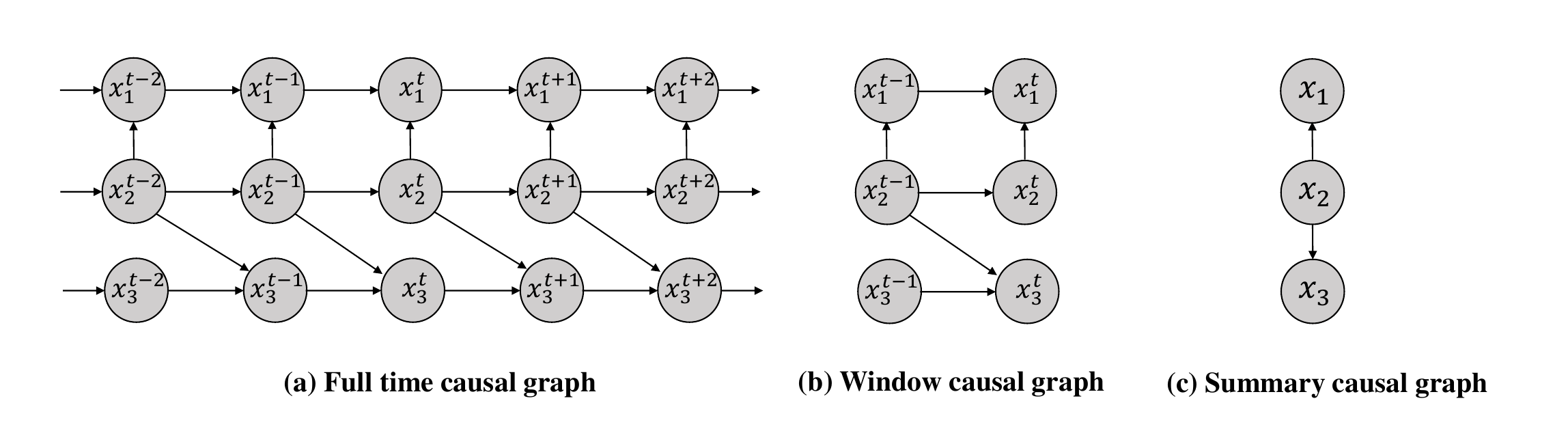}
	\caption{Causal graphs for discrete-time observations.}
	\label{fig:full_window_summary_graph}
\end{figure}

For temporal data, the causal relationship can be intuitively defined by the temporal precedence \cite{eichler2012causal} indicating the causes precede their effects. It reveals the causality asymmetric in time and can be used to orient a causal relation when two variables are known to be causally related. 
Based on the temporal precedence, there exist three graphical representations of causal structure, \ie, \textit{full-time causal graph}, \textit{window causal graph}, and \textit{summary causal graph}. 

As illustrated in Figure \ref{fig:full_window_summary_graph} (a), the \textit{full-time causal graph} represents a complete graph of the dynamic systems. For $d$-variate time series $\mathbf{x}$, the measurement at each time point $t$ is a vector $(x_1^t, ..., x_d^t)$. Vertices in full-time causal graphs consist of the set of component $x_1,...,x_d$ at each time point $t$ with lag-specific directed links such as $x_i^{t-k} \to x_j^t$. 
However, it's usually difficult to discover full-time causal graphs due to the single observation for each series at each time point.

To remedy this problem, \textit{window causal graph} is proposed. It assumes a time-homogeneous causal structure such that the dynamics of observation vector $\mathbf{x}$ are governed by $\mathbf{x}^t:=f(\mathbf{x}^{<t}, \mathbf{u}^t)$ where the function $f$ determines the following observation based on past $\mathbf{x}^{<t}$ and the noise $\mathbf{u}^t$.   
As illustrated in Figure \ref{fig:full_window_summary_graph} (b), the window causal graph is represented in a time window, the size of which amounts to the maximum lag in the full-time causal graph.

As shown in Figure \ref{fig:full_window_summary_graph} (c), each time series component is collapsed into a node to form the \textit{summary causal graph}.
The summary graph represents causal relations between time series without referring to time lags \cite{MTS/FCM/nips_PetersJS13}. In many applications,  it is sufficient to model the relations between temporal variables without precisely knowing the interaction between time instants.

For causal discovery from temporal data, most works aim to find the summary causal graph. Nevertheless, summary causal graphs do not always correspond to an SCM, which means they do not enable interventional predictions that are consistent with the underlying time-resolve SCM \cite{background/cg/janzing2018structural,background/cg/uai_RubensteinWBMJG17}.

\subsection{Problem Definitions}

\begin{figure*}
    \centering
	\includegraphics[width=1.0\textwidth]{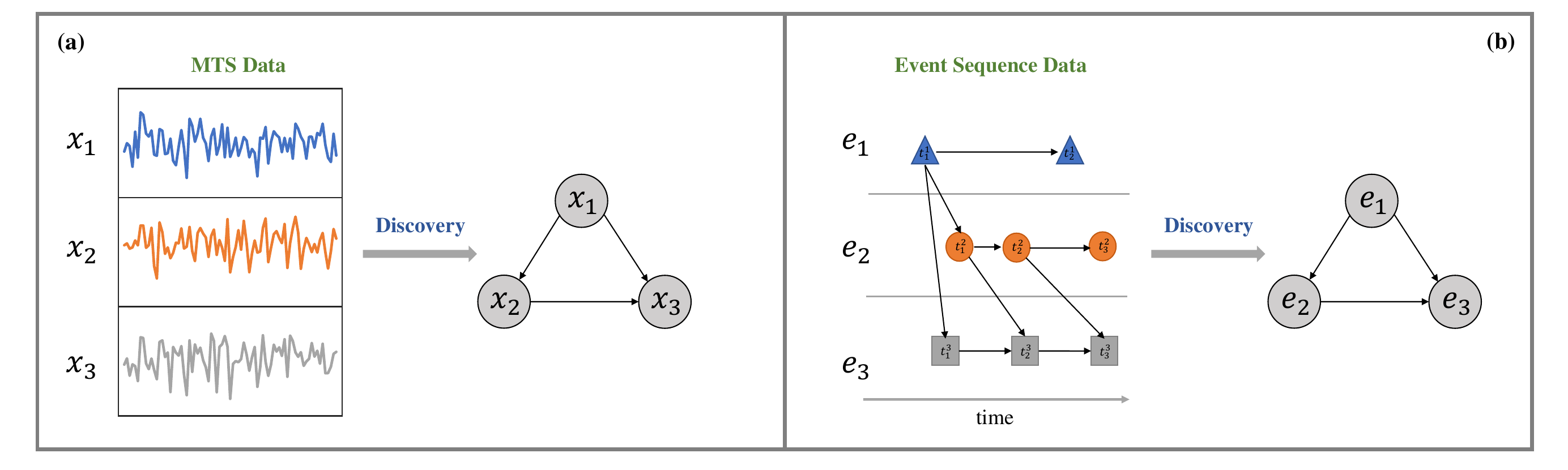}
	\caption{The problem description: (a) causal discovery from MTS data, (b) causal discovery from event sequence data.}
	\label{fig:problem_def}
\end{figure*}

As illustrated in Figure \ref{fig:problem_def}, causal discovery from temporal data can be divided into two problems, \ie, causal discovery from MTS and causal discovery from event sequence. Next, we formally define them respectively.

\textbf{Causal Discovery from MTS.} Consider a time series with $d$ variables: $\{ \mathbf{x}^t \}_{t \in \mathbb{Z}^+} = \{ ({x}^t_1\ {x}^t_2 \ ... \ {x}^t_d )^{\top}  \}_{t \in \mathbb{Z}^+} $. 
Assume that causal relationships between variables are given by the following structural equation model:
\begin{equation}
    x^t_i := f_i( Pa(x^t_i), u^t_i),\ i=1,...,d, \nonumber
\end{equation}
where for any $i \in \{1,...,d\}$ at time instance $t$, $Pa(x^t_i)$ is the set of direct parents of $x^t_i$ which can be both in the past and at the same time instance. 
$u^t_i$ denotes the independent noise and can denote either measurement noise or driving noise \cite{intro/ts_surveys/peters2022causal} without losing generality.
Causal discovery from MTS aims to find either of the two kinds of outputs, \ie, summary causal graph or window causal graph.
As for the summary causal graph, the output is the adjacency matrix $A \in \mathbb{Z}^{d \times d}$ which summarizes the causal structure, and the $(i,j)$-th entry of the matrix $A$ is $1$ if past observations of $x_i$ enter the structural equation of $x_j^t$ and $0$ otherwise. We say that `$x_i$ causes $x_j$' if $A_{ij}=1$.
As for window causal graph with maximum time lag $K$, the output matrices $W$ and $A^k\ (k \in \{1,...,K\})$ correspond to intra-slice and inter-slice edges, respectively. 
For example, $W_{ij}=1$ denotes the instantaneous dependency $x^t_i \to x^t_j$, while $A_{ij}^k=1$ denotes a lagged dependency $x^{t-k}_i \to x^t_j$ for $k>0$.

\textbf{Causal Discovery from Event Sequence.}  For an event sequence: $\{(t_1,e_1),(t_2,e_2),...\}$, $t_i,i=1,2,...$ indicates the time at which the event occurred, while $e_i,i=1,2,...$ stands for the corresponding event type. We aim to discover the causal relationships between different event types. In general, we can construct a causal graph $G=(g_i),i=1,2,...,n$, where each node represents a type of event sequence. Our mission is to discover the edge in the causal graph. For example, if there is a directed edge from node $g_j$ to node $g_i$, we say event-type $g_j$ is a cause of event-type $g_i$.

\section{Causal Discovery from Multivariate Time Series}\label{sec:mts}

In this section, we review causal discovery methods for multivariate time-series data, including constraint-based approaches, score-based approaches, functional causal model-based approaches, Granger causality, and others.
The representative algorithms combined with the characteristics are summarized in table \ref{tab:ts_category_overview}.

\renewcommand{\thefootnote}{\fnsymbol{footnote}}

\begin{table*}[t]
    \centering
    \caption{Characteristics of causal discovery algorithm reviewed for time-series data, arranged by category.}
    \label{tab:ts_category_overview}
    \tiny 
    \resizebox{\textwidth}{!}{
    \begin{tabular}{c|c|cccccccc}
    \toprule
    Section & Method & Causal Graph & Nonlinear & Instantaneous effects & Hidden confounders & Sufficiency Asm. & Markov Asm. & Faithfulness Asm. & Minimality Asm.  \\
    \midrule
    \multirow{9}{*}{Constraint-based}
    & oCSE (2015)~\cite{MTS/CB/oCSE/siamads/0007TB15} &  Summary & Yes & No & No & Yes & Yes & Yes &  \\
    & PCGCE (2022)~\cite{MTS/others/information_criterion/uai/AssaadDG22} &  Extended & Yes & Yes & No & Yes & Yes & Yes & \\
    & PCMCI (2019)~\cite{MTS/CB/PCMCI_runge2019detecting}  &  Window & Yes & No & No  & Yes & Yes & Yes &\\
    & PCMCI$^+$ (2020)~\cite{MTS/CB/insPCMCI_Runge20}  &  Window & Yes & Yes & No & Yes & Yes & Yes &\\
    & ANLTSM (2008)~\cite{MTS/CB/jmlr_ChuG08} &  Window & Yes & Yes & Yes & No & Yes & Yes &\\
    & tsFCI (2010)~\cite{MTS/CB/FCI_tsFCI_entner2010causal} &  Window & Yes & No & Yes & No & Yes & Yes &\\
    & SVAR-FCI (2018)~\cite{MTS/CB_FCI_SVAR_FCI_MalinskyS18} &  Window & No & Yes & Yes & No & Yes & Yes &  \\
    & FCIGCE (2022)~\cite{MTS/others/information_criterion/uai/AssaadDG22} &  Extended & Yes & Yes & Yes & No & Yes & Yes &\\
    & LPCMCI (2020)~\cite{MTS/CB/LPCMCI_GerhardusR20} &  Window & Yes & Yes & Yes & No & Yes & Yes &\\
    \midrule
    \multirow{3}{*}{Score-based}  
    & DYNOTEARS (2020)~\cite{MTS/SB/Dynotears_aistats_PamfilSDPGBA20} &  Window & No & Yes & No & Yes & Yes & No & No\\
    & NTS-NOTEARS (2021)~\cite{MTS/SB/NTS_NOTEARS} &  Window & Yes & Yes & No & Yes & Yes & No & No\\
    & IDYNO (2022)~\cite{MTS/SB/Dynotears_interventionalDATA/icml/GaoBNLY22} &  Window & Yes & Yes & No & Yes & Yes & No & No\\
    \midrule
    \multirow{4}{*}{FCM-Based} 
    &VAR-LiNGAM (2008)~\cite{MTS/FCM/VAR_LINGAM_icml_HyvarinenSH08} &  Window & No & Yes & No & Yes & Yes & No & Yes\\
    &NCDH (2022)~\cite{MTS/FCM/cikm_WuWWLC22} &  Summary & Yes & No & No & Yes & Yes & No & Yes \\
    &TiMINo (2013)~\cite{MTS/FCM/nips_PetersJS13} &  Summary & Yes & Yes & No & Yes & Yes & No & Yes\\
    &NBCB (2021)~\cite{MTS/FCM_maybe/pkdd/AssaadDGA21} &  Summary & Yes & Yes & No & Yes & Yes & Yes\footnotemark[3] & Yes\\
    \midrule
    \multirow{11}{*}{Granger Causality} 
    &HSIC-Lasso-GC (2020)~\cite{MTS/Granger/Kernel_ren2020novel} &  Summary & Yes & No & No & No & No & No & No\\
    &(R)NN-GC (2015,2018)~\cite{MTS/Granger/NN_GC_MontaltoSFTPM15, MTS/Granger/RNN_GC_WangLQLFWP18} &  Summary & Yes & Yes & No & No & No & No & No\\
    &MPIR (2019)~\cite{MTS/Granger/MPIR_ICML19TS_workshop} &  Summary & Yes & No & No  & No & No & No & No\\
    &NGC (2022)~\cite{MTS/Granger/pamiNGC22} &  Summary & Yes & No & No & No & No & No & No \\
    &eSRU (2020)~\cite{MTS/Granger/iclr20_esru} &  Summary & Yes & No & No & No & No & No & No\\
    &SCGL (2019)~\cite{MTS/Granger/SCGL_CIKM_XuHY19} &  Summary & Yes & No & No & No & No & No & No\\
    &GVAR (2021)~\cite{MTS/Granger/iclr21_GVAR_MarcinkevicsV} &  Summary & Yes & No & No & No & No & No & No\\
    &TCDF (2019)~\cite{MTS/Attention/TCDF_NautaBS19} &  Window & Yes & Yes & Yes  & No & No & No & No\\
    &CR-VAE (2023)~\cite{MTS/Granger/CR_VAE2023} &  Summary & Yes & Yes & No & No & No & No & No\\ 
    &InGRA (2020)~\cite{MTS/Attention/icdm_InGRA_ChuWMJZY20} &  Summary & Yes & No & No & No & No & No & No\\
    &ACD (2022)~\cite{Discussion/NewForm/ACD_LoweMSW22} &  Summary & Yes & No & Yes  & No & No & No & No\\
    \midrule
    \multirow{4}{*}{Others} 
    &DBCL (2010)~\cite{MTS/others/difference_based/uai/VoortmanDD10} &  Summary & Yes & Yes & Yes & No & Yes & Yes &  \\
    &NGM (2022)~\cite{MTS/Others/NGM_neuralode_iclr_BellotBS22} &  Summary & Yes & Yes & No & No & No & No & No\\
    &CCM (2012)~\cite{MTS/CCM/work2_main_science_sugihara2012detecting} &  Summary & Yes & No & No & No & No & No & No\\
    &PCTL(c) (2009,2011)~\cite{MTS/logic/uai/KleinbergM09,MTS/logic/ijcai/Kleinberg11} &  Summary & Yes & No & No  & No & No & No & No\\
    \bottomrule
    \end{tabular}}
    \vspace{-4ex}


\end{table*}

\subsection{Constraint-Based Approaches}
\label{subsection:CB} 

As a family of causal discovery algorithms, constraint-based approaches rely on statistical tests of conditional independence and are easy to understand and widely used.
We first give the main ideas of constraint-based approaches, including general steps and causal assumptions.
The detailed methodologies will be categorized into approaches with and without causal sufficiency assumption, and be introduced respectively.

The general steps are: Firstly, it builds a skeleton between variables based on conditional independence. Secondly, it orients the skeleton according to the orientation criterion in the rules.
The goal is to construct \textit{Completed Partially Directed Acyclic Graphs} (CPDAGs) representing the MEC of the true causal diagram.
Central to these approaches to derive MEC from observations are the causal assumptions.
These methods are usually under the assumptions of causal Markov property and faithfulness, and some also assume causal sufficiency (no unobserved confounders). 
In this section, we first review the main algorithms and their extensions to time-series data assuming causal sufficiency, then introduce the approaches for conditions when the causal sufficiency assumption is not guaranteed.

\footnotetext[3]{A lighter version of the faithfulness assumption, termed adjacency faithfulness, is needed.}

\renewcommand{\thefootnote}{\arabic{footnote}}

\subsubsection{Methods with causal sufficiency}
\label{subsection:CB_mwcs} 

In this part, we review methods with causal sufficiency.
To reveal the principles of these approaches, we first give a short introduction to methods in the non-temporal setting. 
Then several popular constraint-based approaches, which originate from the approaches for non-temporal data, for time series are reviewed on the basis of two types of extensions (transfer entropy and momentary conditional independence tests). 

As for extracting causal relations from non-temporal data, the Sprites-Glymour-Scheines (SGS) algorithm \cite{MTS/CB/SGS_origin} is one of the first constraint-based approaches, being proved to be consistent under independently, identically distributed (i.i.d) observations assuming causal sufficiency. However, it suffers from exhausting the test of independence between all nodes. The very large search problem makes it unsuitable in practice.
The Peter-Clark (PC) algorithm \cite{MTS/CB/PC_origin}, which also assumes causal sufficiency, is introduced to reduce unnecessary conditional independence tests and search procedures. 
Given $d$ non-temporal variables, the detailed procedure of PC algorithm is defined as follows in 3 steps: 
(1) Firstly, the algorithm starts with a completed undirected graph $G$.
(2) Secondly, the algorithm respectively retrieves whether there exist pairs of variables $i$ and $j$ are conditioned on other $n$ variables when $n=0,1,2,...,d-2$. If satisfied, remove undirected edges between $i$ and $j$, and update the conditioned variables to the separation set. It proceeds to the pruned skeleton.  
(3) Finally, it determines the collider (V-structure) to obtain the CPDAG and determines the remaining undirected edges based on other rules.

Although approaches such as SGS and PC are designed in non-temporal settings, constraint-based approaches for time-series data are usually extended from them. 
We will review recent four popular constraint-based methods, which also assume causal sufficiency, for time-series data. Among these methods, two extensions \cite{MTS/CB/oCSE/siamads/0007TB15, MTS/others/information_criterion/uai/AssaadDG22} are based on the causal concept of Transfer Entropy, another two \cite{MTS/CB/PCMCI_runge2019detecting, MTS/CB/insPCMCI_Runge20} of them are extended to time series via momentary conditional independence tests.

\textbf{Extension to time series based on Transfer Entropy.}
Traditional constraint-based approaches can be extended to the scenario of time series based on the concept of Transfer Entropy. 
The Transfer Entropy is a model-free measure of temporal causality, of which the definition and variants will be detailed in subsection \ref{subsection:TE}. 
Here we view the Transfer Entropy measure as an off-the-shelf part and review two representative approaches from the perspective of constraint-based methodology.

The Optimal Causation Entropy (\textbf{oCSE}) Principle  \cite{MTS/CB/oCSE/siamads/0007TB15} is proposed to guide computational and data-efficient causal discovery algorithms from MTS data.
It's based on the theoretical concept of Causation Entropy, a generalization of Transfer Entropy for measuring pairwise relations to network relations of many variables
The oCSE method takes a procedure slightly different from that in PC: instead of limiting as much as possible the size of its conditioning set, it conditions since the start on all potential causes which constitute the past of all available nodes.
The algorithm is summarized in Algorithm~\ref{alg:oCSE}, which consists of aggregative discovery of causal nodes, and progressive removal of non-causal nodes.
In detail, given node $j$, two procedures are conducted jointly to infer its direct causal neighbors: (1) Firstly, it discovers a superset $Pa(x_j)$ of $j$'s direct causal neighbors aggregately based on the maximization of Causation Entropy. (2) Secondly, it prunes away non-direct causal neighbors based on the Causation Entropy criterion, for example, $i$ is removed from $Pa(x_j)$ if $\mathrm{CE}(x_i^t \to x_j^{t+1} | Pa(x_j^t) \backslash \{x_i^t\} ) =0  $. It's a computational and sample-efficient algorithm.
However, it assumes that the hidden dynamics follow a stationary first-order Markov process as the Causation Entropy only models causal relations with time lags equal to one. 
Recently, the \textbf{PCGCE} \cite{MTS/others/information_criterion/uai/AssaadDG22} is proposed to extract extended summary causal graphs for time-series data based on the PC algorithm and the Greedy Causation Entropy, which is a variant of the Causation Entropy. 

    \begin{algorithm}[t]
        \caption{oCSE}
        \label{alg:oCSE}
        \KwIn{Multivariate time series $\mathbf{x}$ with $d$ dimensions, a significant threshold $\alpha$}
	\KwOut{The summary causal graph $G$}
        \BlankLine
        Initialize an empty graph $G$ with $d$ nodes $V$ \\
        \textbf{for} $j \in \{1,...,d\}$ \textbf{do} \\
        \quad \# Aggregative Discovery of Causal Nodes\\
        \quad $z=\infty$ \\
        \quad \textbf{while} $z>0$ and card($Pa(x_j)$)$<d$ \textbf{do} \\
        \quad\quad \textbf{for} $x_i\in V \backslash Pa(x_j)$ \textbf{do} \\
        \quad\quad\quad Compute the p-value ($z_p$) corresponding to the test  $\mathrm{CE}(x_i^t \to x_j^{t+1} | Pa (x_j^t) ) >0  $ \\
        \quad\quad \textbf{if} $z_p > \alpha$ \textbf{then} add edge $x_i \to x_j$ to $G$ \\
        \quad \# Progressive Removal of Non-Causal Nodes\\
        \quad \textbf{for} $x_i \in Pa(x_j)$ \textbf{do} \\
        \quad\quad Compute $z$ corresponding to the test $\mathrm{CE}(x_i^t \to x_j^{t+1} | Pa(x_j^t) \backslash \{x_i^t\} ) =0  $ \\
        \quad\quad \textbf{if} $z>\alpha$ \textbf{then} remove edge $x_i \to x_j$ from $G$ 
    \end{algorithm}

\textbf{Extension to time series via Momentary Conditional Independence Tests.}
The \textbf{PCMCI} algorithm \cite{MTS/CB/PCMCI_runge2019detecting} leverages a variant of the PC algorithm that flexibly combines linear or nonlinear conditional independence tests and extracts causal relations from time-series data. The goal of the algorithm is to discover the window causal graph.
Different from that of PC algorithm, PCMCI starts by constructing a partially connected graph, where all pairs of nodes $(x^{t-k}_i, x^t_j)$ are directed as $x^{t-k}_i \to x^t_j$ if $k>0$. This initialization also caters to temporal priority. The algorithm consists of two stages:
(1) As done in PC, PCMCI removes all unnecessary edges based on conditional independence. It furthermore removes homologous edges based on the assumption of consistency through time.
(2) Momentary Conditional Independence (MCI) is leveraged to deal with autocorrelation, which may lead to spurious correlation.
Here, MCI is a measurement, which conditions on the parents of $x^t_j$ and $x^{t-k}_i$ while testing $X^{t-k}_i \not \! \perp \!\!\! \perp X^t_j | Pa(X^t_j) \textbackslash \{ X^{t-k}_i\},  X^{t-k}_i$. It also provides an interpretable notion of causal strength from $x^{t-k}_i$ to $x^t_j$.
PCMCI has been shown to be consistent and can be flexibly combined with any kind of conditional independence test (linear or nonlinear), such as partial correlation and mutual information. In recent years, there is also a wealth of machine learning approaches on nonparametric tests that address a wide range of independence and dependence types~\cite{MTS/CB/CI_uai_ZhangPJS11, MTS/CB/CI_aistats_Runge18}.

The \textbf{PCMCI$^+$} algorithm \cite{MTS/CB/insPCMCI_Runge20} extends PCMCI to include the discovery of instantaneous causal relations.
Central to the PCMCI$^+$ algorithm are two basic ideas that deviate from the origin PC algorithm: First,  it conducts the edge removal process separately for lagged and contemporaneous conditioning sets. Second, it leverages MCI to calibrate CI tests under autocorrelation, which is similar to that in PCMCI.
The author in \cite{MTS/CB/insPCMCI_Runge20} also details the curse and blessing of autocorrelation.


\subsubsection{Methods without causal sufficiency}
Constraint-based approaches without causal sufficiency will be reviewed in this part.
In the beginning, we give a brief introduction to the Fast Causal Inference (FCI) Algorithm~\cite{MTS/CB/PC_origin} for non-temporal data.
Then, methods for MTS data consist of two categories: (1) Fast causal inference through time-series models, which is extended from FCI. (2) The methodology via momentary conditional independence tests.

The FCI algorithm is a generalization of the PC algorithm, which can be used in the presence of latent confounders and proven to be asymptotically correct. 
It utilizes independence tests on the observed data to extract (partial) information on ancestral relationships between the observed variables, thus the goal of the FCI algorithm is to infer the appropriate PAG. 
The FCI algorithm starts by constructing a complete graph consisting of undirected edges, similar to the PC algorithm.
Then iterative conditional independence tests are conducted for the removal of edges. As a result, the FCI algorithm removes edges that are independent, first when conditioning with Sepsets and the with Possible-Dsep sets. For the remaining undirected edges, ten orientation rules are applied recursively.
The detailed FCI algorithm, including theoretical analysis, demonstrates the algorithm is sound and complete and can be found in \cite{MTS/CB/FCI_completeness_Zhang08}.

\textbf{Fast Causal Inference Through Time-series Models.}
A constraint-based method called additive nonlinear time series model (\textbf{ANLTSM}) \cite{MTS/CB/jmlr_ChuG08} is proposed under the assumption that the effects of hidden confounders are linear and contemporaneous.
To escape the curse of dimensionality for nonparametric conditional independence tests, ANLTSM leverages additive regression model, which can be specified as follows :
\begin{equation}
    x_j^t = \sum_{ 1 \leq i \leq d, i \neq j} a_{j,i}x^t_i + \sum_{1 \leq i \leq d, 1 \leq l \leq \tau} f_{j,i,l}(x^{t-l}_i) + \sum_{r=1}^h b_{j,r}u^t_r + e^t. \nonumber
\end{equation}
Here, $a_{j,i}$ and $b_{j,r}$ are constant values, and $f_{j,i,l}(\cdot)$ denotes the smooth univariate function. The unobserved effects in the form of multi-dimensional Gaussian white noise can be categorized into two types: $ e^t $ reflects the latent direct causes of the observed variables, and $(u^t_r)_{1\leq r \leq h}$ denotes latent common causes. And the latent common causes affect the observed variables at the same instant. For $x^t_i$ and $x^t_j$, $u_r^t$ suffices to be stated as a latent common cause if and only if there exists $1 \leq r \leq h$ such that $b_{j,r}b_{i,r} \neq 0$.
Based on the aforementioned additive regression model, the FCI algorithm is leveraged to identify lagged and instantaneous causal relations. For detecting the instantaneous relations, the conditional independence between $x^t_i$ and $x^t_j$ is first tested given the set $S$ by estimating the conditional expectation $\mathbb{E}(x^t_i | x^t_j \cup  S)$, then the significance of prediction relationship between $x^t_i$ and $x^t_j$ is checked using statistical tests such as the F-test or the BIC scores, where the insignificance of the predictor implies the conditional independence between $x^t_i$ and $x^t_j$. The lagged causal relations are identified in a similar way. The remaining edges are oriented based on rules.
This method is shown to be consistent if the data generation caters to the additive nonlinear time series models.
However, the ANLTSM method restricts contemporaneous interactions to be linear, and latent confounders to be linear and contemporaneous.

Another extension of FCI to time-series data is the \textbf{tsFCI} \cite{MTS/CB/FCI_tsFCI_entner2010causal} algorithm, where the FCI algorithm is directly applied via a time window.
In detail, by assuming the observed time-series data comes from a system at equilibrium, the original time-series data is transformed into a set of samples of the random vector, via a sliding window of size $\tau$. 
Then considering every component of the transformed vector as a separate random variable, the original FCI algorithm is directly applicable. As the amount of information derived from standard FCI is quite restricted, the temporal priority and time invariance is further incorporated as background knowledge to make more inferences in the orientation phase.
However, the tsFCI ignores selection variables and contemporaneous causal relations.
Recently, a constrain-based approach named \textbf{SVAR-FCI} \cite{MTS/CB_FCI_SVAR_FCI_MalinskyS18} is proposed that allows for both instantaneous influences and arbitrary latent confounding in the data-generating process. 
Similar to tsFCI, it also uses time invariance to infer additional edge removals.

\textbf{Methodology Via Momentary Conditional Independence Tests.}
It's found that the original FCI algorithm and its temporal variants suffer from low recall in the autocorrelated time-series case due to the low effect size of conditional independence tests \cite{MTS/CB/LPCMCI_GerhardusR20}.
Some researchers aim to extend PCMCI in the presence of unobserved confounding variables to tackle the aforementioned issues. 
In \cite{MTS/CB/LPCMCI_GerhardusR20}, the Latent PCMCI (\textbf{LPCMCI}) algorithm is proposed.
Central to the LPCMCI algorithm are two ideas that: 
First, based on the analysis of the effect size in causal discovery, it uses parents of variables as default conditions and non-ancestors are not tested in the condioning sets, which not only avoids inflated false positives but also reduce the sets to be tested.
Second, it introduces the notions of middle marks and LPCMCI-PAGs as an unambiguous causal interpretation to facilitate the early orientation of edges.
And the LPCMCI algorithm is proven to be order-independent, sound and complete.

\subsection{Score-Based Approaches}
\label{subsection:SB} 

Another family of causal discovery approaches is based on score function.
The main ideas of score-based approaches will first be introduced, including (dynamic) Bayesian Network, characteristics of score-based approaches compared to their constraint-based counterpart, model scoring, and model search.
Then, we will review combinatorial search approaches and continuous optimization approaches for MTS, respectively.

\subsubsection{Basics of score-based approaches}

The score-based approaches are motivated by the idea that graph structures encoding the wrong (conditional) independence will also result in poor model fit. In the score-based approaches, the causal structure is attached to the concept of \textit{Bayesian Network (BN)} or \textit{Dynamic Bayesian Network (DBN)} \cite{DBN/dean1989model, DBN/murphy2002dynamic} dealing with temporal data. 
In light of this, the score-based methods can generate and probabilistically score multiple models, and then output the most probable one. This contrasts with the constrained-based approaches, which derive and output a single model without quantification regarding how likely it's to be correct.
And the faithfulness assumption is diluted in the scored-based approaches by applying a goodness-of-fit measurement instead of a conditional independence test.
The problem of learning a BN or DBN from observations can be therefore formulated as: given a set of instances, find the network that best matches them, i.e., optimize the objective functions.  
It consists of two elements: \textit{model scoring} and \textit{model search}.

\textbf{Model scoring.} Common objective functions fall under two categories: the Bayesian scores which focus on goodness-of-fit and allow the incorporation of prior knowledge, and information-theoretic scores which explicitly consider model complexity, aiming to avoid over-fitting, in addition to the goodness-of-fit \cite{intro/nonts_surveys/BN21}. The family of Bayesian score functions contains Bayesian Dirichlet equivalent (BDe) score \cite{scorefunctions/BDe/ml/HeckermanGC95}, K2 score \cite{scorefunctions/K2_score/corr/abs-1301-0576}, and so on.
The most widely used information-theoretic scores include the Bayesian Information Criterion (BIC) \cite{scorefunctions/BIC/neath2012bayesian} and the Akaike Information Criterion (AIC) \cite{scorefunctions/BIC_AIC/burnham2004multimodel}.

\textbf{Model search / Optimization.} The score-based approaches cast the problem of searching causal structure  $G$ into an optimization program using the aforementioned score functions $S$. The ultimate goal is therefore stated as \cite{peters2017elements}:  
\begin{equation}
    \hat{G} = \mathrm{argmin}_{G\ \mathrm{over} \ \mathbf{x}} S(D,G), \nonumber
\end{equation}
where $D$ represents the empirical data for variables $\mathbf{x}$.
Traditionally, it's a combinatoric graph-search problem, and the solution is generally sub-optimal as finding a globally optimal network is known to be NP-hard \cite{MTS/SB/NP_COMPLETE/aistats/Chickering95}. A line of works, such as Greedy Equivalence Search (GES) \cite{MTS/SB/GES_start_jmlr_Chickering02} involve local heuristics owing to the large search space of graphs. However, they still suffer from the curse of dimensionality and suboptimal problems. 
Recently, an algebraic result characterizing the acyclicity constraint is leveraged in structure learning, which turns the combinatoric problems into a continuously optimizing problem \cite{MTS/SB/linear_NOTEARS/nips/ZhengARX18, MTS/SB/nonlinear_NOTEARS/aistats/ZhengDARX20}, which can be reformulated as: 
    \begin{equation}
    \begin{aligned}
        \mathrm{min}_{ \mathbf{A} \in  \mathbb{R}^{d \times d}          } &S(\mathbf{A})   \nonumber     \\
    	\mathrm{subject\ to\ } G(\mathbf{A} ) &\in \mathrm{DAGs}     \nonumber \\
    \end{aligned}
    \qquad                 
    \begin{aligned}
        \mathrm{min}_{ \mathbf{A} \in  \mathbb{R}^{d \times d}          } &S(\mathbf{A})  \nonumber \\
        \mathrm{subject\ to\ } h(&\mathbf{A}) = 0  \nonumber \\
    \end{aligned}
\end{equation}
where $\mathbf{A}$ denotes the adjacency matrix, and $h$ is the function used to enforce acyclicity in the inferred structure. The original acyclicity constraint function is implemented as $h(\mathbf{A}) = \mathrm{tr}(e^{\mathbf{A} \odot  \mathbf{A}   }) - d$ in \cite{MTS/SB/linear_NOTEARS/nips/ZhengARX18}. 
It relies on the augmented Lagrangian method (ALM) \cite{ALM/networks/Yurkiewicz85} to solve the continuous constrained optimization problem.
Various works have further adopted the continuous constrained formulation in neural networks to extract nonlinear causal relations \cite{MTS/SB/nonlinear_NOTEARS/aistats/ZhengDARX20, NOTMTS/GNN/DAG_GNN/icml/YuCGY19, NOTMTS/GNN/DAG_CAN/icassp/GaoSX21}.

In the context of time series, the ultimate goal of score-based approaches is to learn the structure of DBN. A DBN is a probabilistic network where variables are time series, and it can be decomposed into a prior network and a transition network. A prior network provides dependencies between variables in a given time stamp, and a transition network provides dependencies over time. Therefore, a DBN represents contemporaneous and time-delayed effects in the same framework. Based on this extension to time series, we review the score-based methods following a similar paradigm from combinatoric search to continuous constrained optimization.


\subsubsection{Combinatorial search approaches}
To conduct the combinatorial search based on scoring function from MTS data efficiently, researches have developed various approaches including structural expectation-maximization~\cite{MTS/SB/learnDBN/uai/FriedmanMR98}, cross-validation~\cite{MTS/SB/CV_DBN_prl/PenaBT05}, and the decomposition of score functions~\cite{MTS/SB/learnDBN/jmlr/CamposJ11}.

In \cite{MTS/SB/learnDBN/uai/FriedmanMR98}, the author first utilizes Structural Expectation-Maximization (\textbf{Structural EM}) algorithm \cite{SB/structural_EM/icml/Friedman97, SB/structural_EM/uai/Friedman98}, which is originally a standard algorithm for inferring BN, to learn DBN from longitudinal data.
The Structural EM algorithm, combining structural and parametric modification with a single EM process, can be shown to find local optima defined by score functions.

In \cite{MTS/SB/CV_DBN_prl/PenaBT05}, the $K$-fold \textit{cross-validation (CV)} is leveraged as a computationally feasible scoring criterion for learning DBN. Given the observational data $D$, which is randomly split into $K$ folds $D^1,...,D^K$ of approximately equal size, the CV value of a model $G$ is formulated as $\frac{1}{T} \sum_{k=1}^K \mathrm{log} p(D^k | G, \hat{\theta}^k)$. And a greedy hill-climbing search is used to estimate $E[\mathrm{log}p(D_{T+1} | G, \hat{\theta} )]$.
The procedure starts from the empty graph and updates it gradually by applying the highest scoring single edge additional or removal available. Experiments show that the scoring methods based on cross-validation lead to models generalizing better than those based on BIC of BDe for a wide range of sample sizes. 

Based on the score functions that are decomposable, the paper \cite{MTS/SB/learnDBN/jmlr/CamposJ11} uses structural constraints to cast the problem of structure learning in DBN into a corresponding augmented BN, and presents a branch-and-bound algorithm to guarantee global optimality. The decomposed form of the optimal goal can be formalized as:
\begin{equation}
    (G^{0*}, G'^*) = \mathrm{argmax}_{G^0, G'}  (S_{D_0}(G^0)  + S_{D_{1:T}}(G') ) =( \mathrm{argmax}_{G^0}S_{D_0}(G^0) + \mathrm{argmax}_{G'}  S_{D_{1:T}}(G')    ), \nonumber
\end{equation}
where $G^0$ and $G'$ correspond to the prior network and the transition network respectively.
Structural constraints, as a way to reduce the search space, specify where arcs may or may not be included.
Because of the branch-and-bound properties, the algorithm can be stopped at the best current solution and an upper bound for the global optimum.
The proposed method is shown to be able to handle larger data sets than before, benefiting from the branch-and-bound algorithm and structural constraints.

\subsubsection{Continuous optimization approaches}

Owing to the recent contribution of NOTEARS \cite{MTS/SB/linear_NOTEARS/nips/ZhengARX18}, the score-based learning of DAGs can be reformulated as a continuous constrained optimization problem, which inspires various works \cite{MTS/SB/nonlinear_NOTEARS/aistats/ZhengDARX20, NOTMTS/GNN/DAG_GNN/icml/YuCGY19, NOTMTS/GNN/DAG_CAN/icassp/GaoSX21, NOTEARS/son/sdm/Ng0FLC022, NOTEARS/SON/corr/abs-1911-07420, NOTEARS/SON/iclr/LachapelleBDL20} in structure learning. 
At the heart of this line of the method is an algebraic characterization of acyclicity expressed as a constraint function, which is further leveraged to minimize the least square loss while enforcing acyclicity.
In the context of time series, some works have also adopted this continuous constrained formulation to support structure learning and causal discovery \cite{MTS/SB/Dynotears_aistats_PamfilSDPGBA20, MTS/SB/NTS_NOTEARS, MTS/Others/SrVARM_www_HsiehSTWH21, MTS/SB/Dynotears_interventionalDATA/icml/GaoBNLY22}.

\textbf{DYNOTEARS}, introduced in \cite{MTS/SB/Dynotears_aistats_PamfilSDPGBA20}, captures linear relations from time-series data via a continuous optimization approach.
It models the data in the following standard SVAR way:
\begin{equation}
    \mathbf{x}^t = \mathbf{x}^t \mathbf{W} + \mathbf{x}^{t-1} \mathbf{A}^1 + ... + \mathbf{x}^{t-p} \mathbf{A}^p + \mathbf{u}^t , \nonumber
\end{equation}
where $p$ is the order of SVAR model, $\mathbf{u}$ is a vector of centered error variables.
To guarantee the identifiability in SVAR models, the error terms $\mathbf{e}^t$ are assumed either non-Gaussian or standard Gaussian, i.e., $\mathbf{u}^t \sim \mathcal{N}(0, I)$, as the identifiability is proven to hold on the two cases \cite{NOTEARS/optimization/jmlr/HyvarinenZSH10, peters2017elements}.
$\mathbf{W}$ and $\mathbf{A}$ are weighted adjacency matrices, which correspond to intra-slice edges (contemporaneous relationship) and inter-slice edges (time-lagged relationship), respectively. The SEM can further takes the compact form:$\mathbf{X}^t = \mathbf{X}^t \mathbf{W} + \mathbf{X}^{(t-p):(t-1)} \mathbf{A} + \mathbf{U}$.
The procedure of structure learning revolves around minimizing the least-squares loss subject to an acyclicity constraint, which gives the following optimization problem:
\begin{equation}
    \begin{aligned}
        \mathrm{min}_{ \mathbf{W}, \mathbf{A} }\ \ f(\mathbf{W}, \mathbf{A}) \ \ \mathrm{s.t.}& \ \mathbf{W}\ \ \mathrm{is\ \ acyclic},  \nonumber \\
        \mathrm{where}\ \ f(\mathbf{W}, \mathbf{A}) = 
        \frac{1}{2n}|| \mathbf{X}^t - \mathbf{X}^t \mathbf{W} - \mathbf{X}^{(t-p):(t-1)} \mathbf{A}&||_F^2
        + \lambda_{\mathbf W}|| \mathbf{W} ||_1 + \lambda_{\mathbf{A}}||\mathbf{A}||_1.  \nonumber
    \end{aligned}
\end{equation}
To sidestep the key difficulty of solving the optimization problem under the acyclicity constraint, DYNOTEARS follow the work in \cite{MTS/SB/linear_NOTEARS/nips/ZhengARX18}, where the trace exponential function $h(\mathbf{W}) =  \mathrm{tr}(e^{\mathbf{M} \odot  \mathbf{M}   }) - d $ is leveraged as an equivalent formulation of acyclicity. The continuous constrained optimization problem is translated via the augmented Lagrangian method into unconstrained problems of the form:
\begin{equation}
    \mathrm{min}_{\mathbf{W},\mathbf{A}} F(\mathbf{W},\mathbf{A}),\  \mathrm{where} \ 
F(\mathbf{W}, \mathbf{A}) = f(\mathbf{W}, \mathbf{A}) + \frac{\rho}{2} h(\mathbf{W})^2 + \alpha h(\mathbf{W}). \nonumber
\end{equation}
Towards the optimization of the above smooth augmented objective, two solving approaches are presented separately. 
The first approach is to use standard solvers such as L-BFGS-B \cite{NOTEARS/LBFGSB/toms/ZhuBLN97}. An alternative approach is a two-stage procedure similar to those in \cite{NOTEARS/optimization/jmlr/HyvarinenZSH10}, where we can rewrite the equation as $\mathbf{z} = \mathbf{z}\mathbf{W} + \mathbf{U} $ and derive the estimate of $\mathbf{W}$ by using static NOTEARS to the error term $\mathbf{z}$.

\textbf{NTS-NOTEARS} \cite{MTS/SB/NTS_NOTEARS} is a recent advance that adopts the continuous constrained formulation. Compared to DYNOTEARS, which is a linear autoregressive model, NTS-NOTEARS is able to extract both linear and non-linear relations among variables. It achieves this by leveraging 1D convolutional neural networks (CNNs), which exploit a sequential topology in the input data and are thus well-suited neural function approximation models for temporal data. 
$d$ CNNs, each of which the first layer is a 1D convolutional layer with $m$ kernels, are trained jointly where the $j$-th CNN predicts the expectation of targeted variable $x_t^j$ at the specific time $t$ given preceding and contemporaneous input variables.
Each CNN can be viewed as a Markov blanket of the target variable.
The dependence of child variables on their parents in DBN is given as follows:
\begin{equation}
    \mathbb{E}[ x^t_j | Pa(x^t_j) ] = \mathrm{CNN}_j ( \{ \mathbf{x}^{t-k}: 1 \leq k \leq K   \},  \mathbf{x}^{t}_{-j}            )  , \nonumber
\end{equation}
where parents $Pa(x^t_j)$ are derived from the trained CNNs, and $\mathbf{x}^{t}_{-j}$ denotes all variables at time step $t$ except $x_j$. 
In light of NOTEARS-MLP \cite{MTS/SB/nonlinear_NOTEARS/aistats/ZhengDARX20} (a non-linear and NN-based extension of NOTEARS), the dependency strength of an edge in DBN is estimated in the following way:
\begin{equation}
    W^{k}_{ij} = || \phi^k_{i,j} ||_{L}^2\ \mathrm{for} \ k=1,...,K+1 . \nonumber
\end{equation}
In detail, the $ x^{t-k}_i$ belongs to the parent set $Pa(x^t_j)$ on the condition that the estimated dependency strength is larger than threshold weight $ W^{k}_{ij} > W^{k}_{thres}  $.
The optimization procedure follows a similar way as DYNOTEARS. It's also worth noticing that NTS-NOTEARS shows prior knowledge of variable dependencies that can be transformed as additional optimization constraints and incorporated into the L-BFGS-B solver.

To handle both observational and interventional data, an algorithm, called \textbf{IDYNO} \cite{MTS/SB/Dynotears_interventionalDATA/icml/GaoBNLY22}, is proposed recently. 
It first introduces a non-linear objective through neural networks to model complex dynamics, then modifies an objective and general solution approach to handle different distributions on intervention targets.

We can find that it's a powerful methodology for score-based structure learning to use continuous optimization and avoid the explicit combinatoric traversal of possible causal structures. The past several years have also witnessed numerous applications and extensions of this methodology. However, some boundaries and limitations are further discussed in \cite{Discussion/scale/npl_impairNOTEARS_KaiserS22, Discussion/scale/bewareof_nips_ReisachSW21, MTS/SB/aistats_NgLKL022}, including the influence of data scale and the convergence condition of the augmented Lagrangian method. 
We recommend you take these issues into consideration for further developments and applications of this family of methods.

\subsection{FCM-Based Approaches}  
\label{subsection:FCM} 

The two families of methods above either face the inseparability of the MEC or the need for large samples to confirm causal faithfulness.   
Causal discovery can also be conducted based on Functional Causal Models (FCM) \cite{General/pearl2000models}, which is also known as SCM in \ref{subsection:key_concepts} and describes a causal system via a set of equations. 
Recent years have witnessed the proliferation of FCM-based approaches for both temporal and non-temporal data.
In this subsection, we first introduce the main ideas of FCM-based approaches, including the functional causal model and the usage of noise in orienting causal relations.
Then two families of FCM-based approaches, \textit{i.e.,} methods using independent component analysis and additive noise model, will be reviewed, respectively.

In FCM, each variable is explained by an equation in terms of its direct causes and some additional noise. For example, the function $x_j = f_j(x_i, u_j) $ explains the causal link $x_i\to x_j$ with some additional noise $u_j$. 
One basic idea of the FCM-based causal discovery approaches is that statistical noise can be a \textit{valuable} source of insight, which caters to recent discoveries \cite{climenhaga2021causal} challenging the orthodoxy that the noise should be treated as a nuisance. To be specific, causal relationships can be identified and estimated with the help of noise.

\subsubsection{Methods using independent component analysis}
In this part, we first introduce the basic idea of this family of methods by reviewing the original algorithm in non-temporal setting \cite{MTS/FCM/LiNGAM_jmlr_ShimizuHHK06}.
Then, methods for MTS data will be detailed~\cite{MTS/FCM/VAR_LINGAM_icml_HyvarinenSH08, MTS/FCM/VAR_LINGAM_v2_jmlr_HyvarinenZSH10, MTS/FCM/ijcai2013/SchaechtleSB13,MTS/FCM/cikm_WuWWLC22}.

LiNGAM \cite{MTS/FCM/LiNGAM_jmlr_ShimizuHHK06} is a typical FCM-based causal discovery algorithm in non-temporal setting, and has the following assumptions: (1) a linear data generation process, (2) non-Gaussian disturbances, (3) no unobserved confounders. 
In the LiNGAM model, the relations among observations can be formulated as $\mathbf{x} = \mathbf{B}\mathbf{x} + \mathbf{u}$, where $\mathbf{x},\mathbf{B},\mathbf{u}$ respectively denote the vector of variables, the adjacency matrix of the causal graph and the noise vector. The equation can be rewritten as $\mathbf{x}=\mathbf{A}\mathbf{u}$, where $\mathbf{A} = (\mathbf{I}-\mathbf{B})^{-1}$. For the equation, the \textit{independent component analysis (ICA)} method \cite{stone2004independent_ICA} can be used to estimate $\mathbf{A}$, and causal relationships $\mathbf{B}$ can be derived. Along this line, DirectLiNGAM \cite{MTS/FCM/DirectLiNGAM_jmlr_ShimizuISHKWHB11} further leverages the regression model to ensure the original models to converge to the correct solution in a controlled number of steps. Extensions of LiNGAM to time series are as follows.

As a temporal extension of LiNGAM, \textbf{VAR-LiNGAM} \cite{MTS/FCM/VAR_LINGAM_icml_HyvarinenSH08, MTS/FCM/VAR_LINGAM_v2_jmlr_HyvarinenZSH10} estimates the structural autoregressive (SVAR) models by leveraging non-Gaussianity property. SVAR models reflect both instantaneous and time-delayed causal effects and are among the most prevalent tools in empirical economics to analyze dynamic phenomena \cite{MTS/FCM/VAR_LINGAM_econ_explain_moneta2013causal}. In VAR-LiNGAM, a representation of time series is a combination of SVAR and SEM, which is defined as:
\begin{equation}
   \mathbf{x}^t = \sum_{k=0}^\tau \mathbf{B}^k \mathbf{x}^{t-k} + \mathbf{u}^t   \tag{SVAR} 
\end{equation}
where $\mathbf{B}^k$ is the $n\times n$ matrix of the causal effects between the variables $\mathbf{x}$ with time lag $k$. And $\mathbf{u}^t$ are random processes modeling the external influences or `disturbances', which are assumed to be independent, temporally uncorrelated and non-Gaussian.
To estimate the above model, a classic least-squares estimation of the autoregressive (AR) model (time lag $k>0$) is combined, which is formalized as:
\begin{equation}
\mathbf{x}^t = \sum_{k=1}^\tau \mathbf{M}^k \mathbf{x}^{t-k} + \mathbf{n}^t   \tag{VAR}    
\end{equation}
Based on the SVAR and VAR formalization, the basic idea of VAR-LiNGAM is that we can estimate $\mathbf{M}^k$ of VAR model in a classic least-square fashion consistently and efficiently. And we can deduce the estimate of instantaneous causal effect through LiNGAM analysis. As for the time-delayed effect, it can be derived from reparametrization. The ensuing method in detail is defined as follows in four steps:
(1) Firstly, fit the regressions and denote the least-squares estimates of the AR matrices by $\hat{\mathbf{M}}^k$.
(2) Secondly, compute the residuals, i.e., $\hat{\mathbf{n}}^t = \mathbf{x}^t -  \sum_{k=1}^\tau  \hat{\mathbf{M}}^k  \mathbf{x}^{t-k}  $.
(3) Thirdly, perform LiNGAM analysis \cite{MTS/FCM/LiNGAM_jmlr_ShimizuHHK06} based on the equation $\hat{\mathbf{n}}^t = \mathbf{B}^0 \hat{\mathbf{n}}^t + \mathbf{e}^t$ to derive the estimate of instantaneous causal effect $\hat{\mathbf{B}}^0$.
(4) Finally, compute the estimates of the time-delayed causal effect $\hat{\mathbf{B}}^k (k>0)$ as $\hat{\mathbf{B}}^k= (\mathbf{I} - \hat{\mathbf{B}}^0 )\hat{\mathbf{M}}^k$.
The VAR-LiNGAM model degenerates to the LiNGAM model if the order of the autoregressive part is set to zero, i.e., $\tau = 0$. And an intensive application of this approach in empirical economics can be found in \cite{MTS/FCM/VAR_LINGAM_econ_explain_moneta2013causal}.

The VAR-LiNGAM is extended to the identification and estimate of causal models under time-varying situations \cite{MTS/FCM/VAR_LINGAM_extend1_ijcai_HuangZS15}, where Gaussian Process regression is further leveraged to automatically model how the causal model change over time. 
In \cite{MTS/FCM/VAR_LINGAM_extend3_lanne2017identification}, the initial VAR-LiNGAM is generalized to the condition where the inferred graphs can contain cycles. And the proposed model is demonstrated theoretically to be identifiable.
Another algorithm based on LiNGAM, called the Multi-Dimensional Causal Discovery (MCD), is proposed in~\cite{MTS/FCM/ijcai2013/SchaechtleSB13}.
MCD can efficiently discover causal dependencies in multi-dimensional settings, such as time-series data, by integrating data decomposition and projection.

To get rid of constraints of linear \cite{MTS/FCM/VAR_LINGAM_icml_HyvarinenSH08, MTS/FCM/VAR_LINGAM_v2_jmlr_HyvarinenZSH10} or additive assumptions\cite{MTS/FCM/nips_PetersJS13}, an FCM-based algorithm named Nonlinear Causal Discovery via HM-NICA (\textbf{NCDH}) is recently proposed in \cite{MTS/FCM/cikm_WuWWLC22} to extract general nonlinear relations from time series.
At the heart of this algorithm, a nonlinear ICA algorithm is leveraged as a measurement of nonlinear relationships. The observations are assumed to be generated by mutually independent latent components:
\begin{equation}
    \mathbf{x} = \mathbf{f}(\mathbf{S}) \ \mathrm{where} \ \mathbf{f}=(f_1,f_2,...,f_d)^T \ \mathrm{and} \ \mathbf{S}=(S_1,S_2,...,S_d)^T. \nonumber
\end{equation}
Similar to that in linear ICA, $\mathbf{S}$ contains components that are independent of each other, and the goal of nonlinear ICA is to recover $\mathbf{S}$ from $\mathbf{x}$. 
NCDH first leverages the nonlinear ICA combined with HMM \cite{Discussion/Nonstation/maybe/uai/HalvaH20} to separate latent noises. As a remedy for the permutation uncertainty of ICA, a series of independence tests are conducted to determine the corresponding relations between the observed variables and the separated noises. A recursive search algorithm is finally taken to extract the causal relations.

\subsubsection{Methods using additive noise model}

In reality, there're many non-linear causal relationships that violate the assumption of LiNGAM family methods. Despite recent advances (such as NCDH) extracting causal relations in general nonlinear conditions, their usages are restricted. 
Another family of FCM-based approaches is based on the \textit{additive noise model (ANM)} with nonlinear function, which is suitable in more general settings.
In this part, the main ideas of methods using ANM will be given firstly. 
Then we will introduce the detailed methods for MTS data.

It's demonstrated in~\cite{MTS/FCM/ANM/nips/HoyerJMPS08} that the true causal structure can be identified in the ANM with nonlinear functions if the causal minimality condition holds. In ANM, if $x_i\to x_j$, we have $x_j = f(x_i)+u_j$, and the cause $x_i$ and additive noise $u_j$ are independent. 
If the noise $u$ is subject to non-Gaussian distribution and $f(\cdot)$ is a linear function.
In the bivariate case $x_i\to x_j$, we can fit regression models in causal and anti-causal directions, the true orientation can be inferred by testing the independence with residuals. As for the multivariate case, a pairwise strategy can be adopted \cite{DBLP:conf/ANM/P2MSetting_icml_MooijJPS09}. The correctness of this algorithm is discussed in \cite{Peters1314/jmlr/PetersMJS14}.



In \cite{MTS/FCM/nips_PetersJS13}, the Time Series Models with Independent Noise (\textbf{TiMINo}) is proposed, which is a causal discovery method for time series based on ANM. 
It inputs time-series data and outputs either a summary time graph or remains undecided, which avoids leading to wrong causal conclusions when the model is mis-specified or the  data is insufficient.
It leverages a similar method as that in non-temporal and multivariate setting \cite{DBLP:conf/ANM/P2MSetting_icml_MooijJPS09}.
In detail, it tries to fit the structural equation models for time series, which can be formulated as follows:
\begin{equation}
    x^t_j = f_j(  Pa( x^{\tau}_j  )^{t - \tau}  , ... , Pa( x^{1}_j  )^{t - 1}, Pa( x^{0}_j  )^{t} ,  u^t_j  ), \nonumber
\end{equation}
where 
error terms $u^t_j$ are jointly independent over variable index $j$ and time index $t$.
There are several options for fitting methods $f$ such as linear models, generalized additive models, and Gaussian process regression models. For inferring causal relations in the additive noise model, independence tests such as cross-correlations and HSIC \cite{DBLP:conf/nips/GrettonFTSSS07} can be leveraged.

There are some drawbacks to those functional causal models, such as VAR-LiNGAM and TiMINo. It's illustrated that those methods are not well scalable across the increase of node numbers \cite{intro/nonts_surveys/glymour2019review}, and those performances are not promising without a large sample size \cite{Discussion/practical_guide/malinsky2018causal}. To overcome those drawbacks, a Noise-Based / Constraint-Based (\textbf{NBCB}) approach is proposed in \cite{MTS/FCM_maybe/pkdd/AssaadDGA21}, where the constraint-based approach is further leveraged based on the original additive noise model for time-series data. 
In detail, the potential causes of each time series are detected by an additive noise model which is similar to that in TiMINo. Unnecessary causal relations are pruned using temporal causal entropy, which is an extension to causation entropy \cite{MTS/CB/oCSE/siamads/0007TB15} measuring the (conditional) dependencies between two-time series.

\subsection{Granger Causality Based Approaches}
\label{subsection:Granger_base} 

Granger causality is a popular tool for analyzing time-series data in many real-world applications. 
There exist many causal discovery approaches developed on the basis of Granger causality.
In this subsection, we first introduce definitions of Granger causality.
Before delving into detailed methods, two categories of Granger causality models for MTS (model-free and model-based) will be given and compared. 
Due to the superiority of model-based approaches in more general conditions, the rest of this part will focus on two recent advances in model-based approaches: (1) methods based on kernels (\ref{subsection:kernel_Granger}), and (2) methods based on neural networks (\ref{subsection:NN_Granger}).

\subsubsection{Basics of Granger causality}

Granger causality analysis, which is first proposed in \cite{granger1969investigating}, is a powerful method that determines cause and effect based on predictability.
A time series $x_i$ Granger-causes $x_j$ if past values of $x_i$ provide unique, statistically significant information about future values of $x_j$.
According to this proposition, $x_i$ is defined to be `causal' for $x_j$ if
\begin{equation}
    \mathrm{var}[x^t_j - \mathcal{P}(x^t_j | \mathcal{H}^{<t})] < \mathrm{var}[x^t_j - \mathcal{P}(x^t_j | \mathcal{H}^{<t} \textbackslash x^{<t}_i)],  \nonumber
\end{equation}
where $\mathcal{P}(x^t_j | \mathcal{H}^{<t})$ denotes the optimal prediction of $x^t_j$ given the history of all relevant information $\mathcal{H}^{<t}$.
Here $\mathcal{H}^{<t} \textbackslash x^{<t}_i$ indicates excluding the information of $x_{<t}^p$ from $\mathcal{H}_{<t}$.
The above definition seems general and does not have specific modeling assumptions, whereas there are also various forms of definition for Granger causality based on different model specifications and statistical tools for better representation power and the convenience of inference, such as autoregression model (in Granger's original paper~\cite{granger1969investigating}) and so on.
And if all relevant variables are observed and no instantaneous connections exist, Granger causal relations are equivalent to causal relations in underlying DAGs \cite{MTS/FCM/nips_PetersJS13, peters2017elements}.

\subsubsection{Early approaches for MTS}
\label{subsection:Granger_early} 

Earlier methods for identifying Granger causality were limited to bivariate settings. 
Specifically, a well-documented \cite{lutkepohl1982non} issues for Granger causal analysis in bivariate settings is that the causal findings may be misleading without adjusting for all relevant covariates.    
On the one hand, it's necessary to account for more variables to prevent identifying incorrect Granger causal relations~\cite{intro/ts_surveys/AliGranger21}.
On the other hand, MTS widely exist among various fields. 
Inferring Granger causal relations in MTS, which is also termed graphical Granger causality or network Granger causality in some literature, has become a hot research topic.
Various graphical Granger causal analysis models for MTS can be divided into two categories, namely model-free and model-based approaches.

\textbf{Model-free Methods.} The mainstream of model-free approaches for multivariate Granger causality are based on predictability and need to estimate the conditional probability density functions (CPDFs) \cite{bai2010multivariate}. 
In \cite{diks2016nonlinear}, the estimates of the CPDFs are provided, and the the bivariate Diks-Panchenko nonparametric causality test is extended to the multivariate case.
By introducing conditional variables into the marginal probability density functions, the copula-based Granger causality model \cite{hu2014copula, kim2020copula} can also be extended to multivariate case.
Besides, model-free measures such as transfer entropy and directed information \cite{MTS/Others/information/directedInform/jcns/AmblardM11}, are able to detect nonlinear dependencies. The definitions and some properties of these model-free estimators will be detailed in \ref{subsection:TE}. 
Model-free methods can deal with nonlinear Granger causal relations well.
However, these estimators suffer from high variance and require large amounts of data for reliable estimation, and also suffer from curse of dimensionality when the number of variables grows. 
Thus, in the complex real-word scenarios where is nonlinear and high-dimensional, the utilization of model-free methods to some extend are limited.

\textbf{Model-based Methods.} In contrast to model-free counterparts, model-based methods are computationally efficient and therefore more suitable for inferring Granger causal relations in high-dimensional conditions.   
The model-based inference approach is adopted by the vast majority of Granger causal whereby the measured time series is modeled by a suited parameterized data generative model. And the inferred parameters ultimately reveal the true topology of Granger causality.
Earlier methods along this line are typically using the popular vector autoregressive (VAR) model under the assumption of linear time-series dynamics. For $d$-variate time series $\mathbf{x}$, the VAR model is defined as:
\begin{equation}
    \mathbf{x}^t = \sum_{k=1}^\tau A^k \mathbf{x}^{t-k} +\mathbf{u}^t, \nonumber
\end{equation}
where $A^k$ is a $d \times d$ matrix that specifies how lag $k$ affects the future evolution of the series and $\mathbf{u}^t$ denotes zero mean noises.
In the VAR model, as a straightforward extension form the bivariate case \cite{granger1969investigating}, time series $i$ does not Granger-cause time series $j$ if and only if for all time lag $k$, the component $(j,i)$ of $A^k$ equals zero.
Thus the Granger causal analysis reduces to determine which entries in $A^k$ are zero over all lags.
There are also abundant research works \cite{intro/ts_surveys/maybe_kdd_ArnoldLA07, MTS/Granger/GGM_kdd_LozanoALR09, MTS/Granger/GGM_bioinformatics_shojaie2010discovering, MTS/Granger/LassoGC/jmlr/BasuSM15} reducing the computational complexity via the Lasso penalty and its variants for Granger causal analysis in high-dimensional time series, which are also termed as Lasso Granger causality (\textbf{Lasso-GC}). For these methods, the problem of Granger causal series selection can be  generally formulated as follows based on least square loss:  
\begin{equation}
    \mathrm{min}_{A^1,...,A^{\tau} \in \mathbb{R}^{d \times d}}   \sum_{t=\tau + 1}^T || \mathbf{x}^t - \sum_{k=1}^\tau A^k \mathbf{x}^{t-k} ||_2^2  + \lambda R(\mathbf{A}), \nonumber
\end{equation}
where $R(\cdot)$ is the sparsity-inducing regularizer and has various implementations as shown in table \ref{tab:lasso_sparsity_overview}.
Different penalty terms induce different sparsity patterns in $A^1,...,A^{\tau}$, thus inducing different heuristics and constraints in the Granger causal series selection.
Except for Lasso-GC, another line of works based on VAR models in multivariate setting worth mentioning is the conditional Granger causality index (CGCI) \cite{MTS/Granger/geweke1982measurement}.
For variable $X,Y$ and conditional variables $Z$, by comparing residuals errors of the reduced and full models $\mathrm{CGCI}_{X\to Y|Z} = \mathrm{ln} \frac{ \mathrm{var}(\epsilon_{Y|Z} ) }{ \mathrm{var}  (\epsilon_{Y|XZ}) }$, a distinction between direct and indirect causality in multivariate systems can be made based on CGCI.
Along this line of works, mBTS-CGCI is proposed in \cite{MTS/Granger/mBTS_CGCI/tsp/SiggiridouK16} based on a modified backward-in-time selection (mBTS) to limit the order of VAR models, thus can be better applied to high-dimensional scenarios.

\begin{table}[t]
    \centering
    \caption{Common sparsity-inducing penalty terms, described by \cite{Lasso_penalty_review/nicholson2017varx, MTS/Granger/iclr21_GVAR_MarcinkevicsV}}
    \label{tab:lasso_sparsity_overview}
    \scalebox{0.95}{
    \begin{tabular}{c|c}
    \toprule
    Model Structure & Penalty Function \\ 
    \midrule
    Basic Lasso &  $|| \mathbf{A} ||_1$    \\
    Elastic net &  $\alpha || \mathbf{A} ||_1   + (1 - \alpha) || \mathbf{A} ||_2^2, \alpha \in (0,1)  $  \\
    Lag group Lasso &  $\sum_{k=1}^{\tau} || \mathbf{A}^k  ||_F   $  \\
    Component-wise Lasso & $ \sum_{p=1}^d \sum_{k=1}^{\tau} || {(\mathbf{A}^{k:\tau}     )}_p    ||_2       $ \\
    Element-wise Lasso &  $\sum_{p=1}^d \sum_{q=1}^d \sum_{k=1}^{\tau}  ||  {(\mathbf{A}^{k:\tau}     )}_{p,q}    ||_2    $\\
    Lag-weighted Lasso & $ \sum_{k=1}^{\tau} k^{\alpha} || \mathbf{A}^k ||_1 , \alpha \in (0,1) $ \\
    \bottomrule
    \end{tabular}}
    \vspace{-4ex}
\end{table}

Although the model-based approaches, compared to the model-free counterparts, take advantage of efficiently processing high-dimensional time series,
the fundamental issue of these approaches is the model misspecification. 
Especially, the notion of multivariate Granger causality based on the vanilla VAR model assumes time series follows linear dynamics, whereas many interactions in real-world applications are inherently nonlinear.
Recently, many model-based approaches, which are compatible with nonlinear causal relations, have emerged and can be grouped into two categories: methods based on \textbf{kernel}, and methods based on \textbf{neural networks}.
As the generation on Granger causality, fundamental venation and development orientations have been reviewed above and prospected from classic documents.
In the following part of this subsection, due to their ability to be leveraged in complex real-world scenarios, we will detail the recent advances of model-based methods in nonlinear and high-dimensional settings, especially new perspectives from neural networks.   

\subsubsection{Recent advances based on kernels}
\label{subsection:kernel_Granger}
To extract nonlinear causal relations in a model-based approach, establishing a nonlinear parameter model is a common strategy.
A line of works extend Granger causality to kernel methods \cite{MTS/Granger/Kernel_PRE_ancona2004radial,    MTS/Granger/Kernel_PRL_marinazzo2008kernel, MTS/Granger/Kernel_PRE_marinazzo2008kernel, MTS/Granger/Kernel_uai_SindhwaniML13, MTS/Granger/Kernel_ren2020novel}.
In \cite{MTS/Granger/Kernel_PRE_ancona2004radial}, Granger causality is extended to bivariate nonlinear cases by means of radial basis functions.
Furthermore, a Granger causality analysis model is put forward \cite{MTS/Granger/Kernel_PRL_marinazzo2008kernel} based on the theory of \textit{reproducing kernel Hilbert spaces (RKHS)}. The key idea is to embed data into a Hilbert space and search for nonlinear relations in that space. This method is then generalized to the multivariate case in \cite{MTS/Granger/Kernel_PRE_marinazzo2008kernel}.
In \cite{MTS/Granger/Kernel_uai_SindhwaniML13}, a matrix-valued extension of the kernel method is proposed, imposed on a dictionary of vector-valued RKHS. The algorithm is for high-dimensional nonlinear multivariate regression, and can naturally lead to nonlinear generalization of graphical Granger Causality.
Recently, an algorithm based on Hilbert-Schmidt independence criterion Lasso Granger causality (\textbf{HSIC-Lasso-GC}) \cite{MTS/Granger/Kernel_ren2020novel} is proposed.

\subsubsection{Recent advances based on neural networks}
\label{subsection:NN_Granger}

\begin{figure*}
    \centering
	\includegraphics[width=1.0\textwidth]{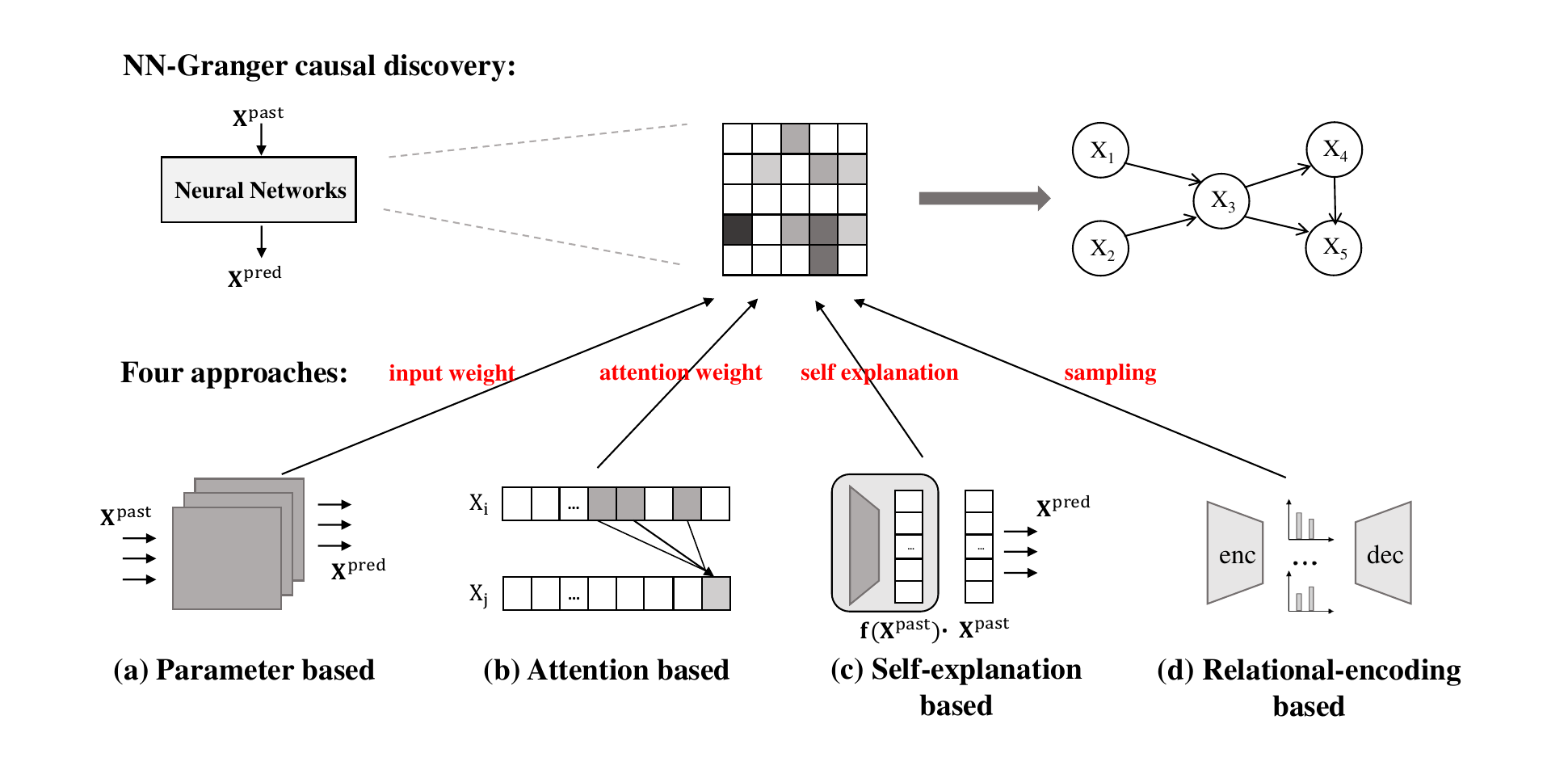}
	\caption{Four types of NN-based Granger causal discovery methods according to the means to derive causal weights.}
	\label{fig:newGC}
\end{figure*}

Neural networks are able to represent nonlinear, complex, and non-additive interactions between variables.
In this part, recent advances of Granger causal methods  based on neural networks will be reviewed, including non-uniform embedding~\cite{MTS/Granger/NN_GC_MontaltoSFTPM15, MTS/Granger/RNN_GC_WangLQLFWP18}, information regularization~\cite{MTS/Granger/MPIR_ICML19TS_workshop}, component-wise neural network modeling~\cite{MTS/Granger/NGC_origin17, MTS/Granger/pamiNGC22, MTS/Granger/iclr20_esru}, low-rank approximation~\cite{MTS/Granger/SCGL_CIKM_XuHY19}, self-explaining networks~\cite{MTS/Granger/iclr21_GVAR_MarcinkevicsV}, attention mechanisms~\cite{MTS/Attention/TCDF_NautaBS19, MTS/Attention/aaai_AME_SchwabMK19}, recurrent variational autoencoders~\cite{MTS/Granger/CR_VAE2023}, and inductive modeling~\cite{MTS/Attention/icdm_InGRA_ChuWMJZY20, Discussion/NewForm/ACD_LoweMSW22}.
Besides, as illustrated in Fig~\ref{fig:newGC}, existing NN-based Granger causality approaches can be categorized into four groups: parameter-based~\cite{MTS/Granger/pamiNGC22, MTS/Granger/iclr20_esru}, attention-based~\cite{MTS/Attention/TCDF_NautaBS19, MTS/Attention/icdm_InGRA_ChuWMJZY20}, self explanation-based~\cite{MTS/Granger/iclr21_GVAR_MarcinkevicsV}, and relational encoding-based~\cite{Discussion/NewForm/ACD_LoweMSW22}.

\textbf{DL-extensions with Non-uniform Embedding.}
A feature selection procedure termed as a non-uniform embedding (NUE) is proposed in \textbf{NN-GC} \cite{MTS/Granger/NN_GC_MontaltoSFTPM15} to identify the significant Granger causes in the MLP model. 
By greedily adding lagged components of predictor time series as input, an MLP is updated iteratively.
A predictor time series is claimed a significant Granger cause of the target time series if at least one of its lagged components is added when the procedure is terminated.
In \textbf{RNN-GC} \cite{MTS/Granger/RNN_GC_WangLQLFWP18}, the NUE is extended by replacing MLPs with gated RNN models,
However, as this technique requires training and comparing many candidate models, it's costly in high-dimensional settings.

\textbf{DL-extensions with Information Regularization.} For extracting nonlinear dynamics,
a method with Minimum Predictive Information Regularization (\textbf{MPIR}) \cite{MTS/Granger/MPIR_ICML19TS_workshop} is introduced.
It leverages learnable corruption for predictor variables and minimizes a mutual information-regularized risk, which combines the benefits of the Granger causality paradigm with deep learning models. 
In MPIR, the author states that the naive way to combine neural nets with Granger causality suffers from two major drawbacks: instability and inefficiency.
The solution is to encourage each $\mathbf{x}_i^{t-K:t-1}$ to provide as little information to $x^t_j$ as possible while maintaining good prediction via learned corruption, replacing the naive way which predicts $x^t_j$ with one $\mathbf{x}_i^{t-K:t-1}$ missing at a time. The risk is defined as follows: 
\begin{equation}
   R_{\mathbf{X}, x_j} [f_{\theta}, \mathbf{n}] =  E_{\mathbf{X}^{t-1}, x^t_j, \mathbf{u}}[ ( x^t_j - f_{\theta}(  \Tilde{\mathbf{ X }}^{t-K:t-1}_{ (\mathbf{n}) } )               )^2          ]     + \lambda \cdot    \sum_{p=1}^d I(\Tilde{X}^{t-K:t-1}_{i(n)}; X^{t-K:t-1}_i), \nonumber
\end{equation}
where $\Tilde{\mathbf{ X }}^{t-K:t-1}_{ (\mathbf{n}) } := \mathbf{X}^{t-K:t-1} + \mathbf{n} \odot \mathbf{e}$ (or its element-wise representation,  $\Tilde{X}^{t-K:t-1}_{ i(n) } := {X}^{t-K:t-1}_i + n_i \cdot e_i, i=1,2,...,d$) are the noise-corrupted inputs with learnable noise amplitudes $\mathbf{n}$ and $u_j \sim N(\mathbf{0},\mathbf{I})$.
And $W_{pq} = I(\Tilde{X}^{t-K:t-1}_{i(n^*)}; X^{t-K:t-1}_i)$ is the minimum predictive information at the minimization of $R_{\mathbf{X}, x_j} [f_{\theta}, \mathbf{n}]$, which contains causal information and measures the predictive strength of variable $i$ for predicting variable $j$, conditioned on all the other observed variables. 
To be specific, $W_{ij} = 0$ if $ x_i \perp \!\!\! \perp x_j $.
Besides, as it's inefficient to estimate the mutual information term with a large dimension, an upper bound is derived as an alternative optimization goal. 
Instead of training many candidate models and suffering from instability and inefficiency, this framework only requires training $d$ models separately.


\textbf{DL-extensions with Component-wise NN Modeling.} Another NN-based approach to measure nonlinear Granger causality is component-wise modeling.
A component-wise framework is proposed in \cite{MTS/Granger/NGC_origin17}, which can be viewed as a generalization of the linear VAR model. In detail, the generation procedure of each variable can be written as follows:
\begin{equation}
    \mathbf{x}^t_j := g_j( \mathbf{x}^{1:(t-1)}_1,... , \mathbf{x}^{1:(t-1)}_i, ..., \mathbf{x}^{1:(t-1)}_d ) + u^t_j,\ \mathrm{for}\ 1\leq j \leq d,  \nonumber
\end{equation}
where $g_j(\cdot)$ is a continuous function, based on regularized neural networks implementation, specifying how the past values of $\mathbf{x}$ determine the future values of variable $\mathbf{x}_j$.
In this context, the time series $\mathbf{x}_i$ is Granger non-causal for time series $\mathbf{x}_j$ ($\mathbf{x}_i  \nrightarrow  \mathbf{x}_j$) if and only if $g_j(\cdot)$ is invariant to $\mathbf{x}^{1:(t-1)}_i$, which can be defined as:
\begin{equation}
   g_j( \mathbf{x}^{1:(t-1)}_1,... , \mathbf{x}^{1:(t-1)}_i, ..., \mathbf{x}^{1:(t-1)}_d ) = g_j( \mathbf{x}^{1:(t-1)}_1,... , \mathbf{x}^{1:(t-1)}_{i'}, ..., \mathbf{x}^{1:(t-1)}_d ), \nonumber
\end{equation}
for all $ ( \mathbf{x}^{1:(t-1)}_1,... ,\mathbf{x}^{1:(t-1)}_d )  $ and all $ \mathbf{x}^{1:(t-1)}_i \neq  \mathbf{x}^{1:(t-1)}_{i'}$.
We will introduce two methods~\cite{MTS/Granger/pamiNGC22, MTS/Granger/iclr20_esru} based on this framework, respectively.

Neural Granger Causality (\textbf{NGC}) is proposed in \cite{MTS/Granger/pamiNGC22} to infer nonlinear Granger causality using structured MLP and LSTM with sparse input layer weights, which are termed as component-wise MLP (\textbf{cMLP}) and component-wise LSTM (\textbf{cLSTM}), respectively.
In the cMLP, each nonlinear output $g_j$ is modeled with a separate MLP as to easily disentangle the effects from inputs to outputs.
The input matrix of the first layer provides information for penalized selection of Granger causality. 
To be specific, in the first layer of $g_j(\cdot)$
\begin{equation}
   h_1^t = \sigma(\sum_{k=1}^{\tau}W^{k}_1 \mathbf{x}^{t-k} + b_1), \nonumber
\end{equation}
if the $i$-th column of weight matrix $W^k_1$ contains zeros for all time lag $k$, then time series $i$ does not Granger-cause series $j$. 
Analogously to the VAR type methods, the Granger causal series are selected by the following encoding selection\cite{MTS/Granger/NGC_origin17} procedure:
\begin{equation}
    \mathrm{min}_{\mathbf{W}} \sum_{t=\tau}^T (x^t_j - g_j(x_{(t-1):(t-\tau)})) + \lambda \sum_{i=1}^{d}  R((W_1)_{:i}), \nonumber
\end{equation}
where sparse inducing penalty $R(\cdot)$ is implemented through group lasso penalty, which extracts causal relations without requiring precise lag specification.
As for the cLSTM, it sidesteps the lag selection problem and the Granger causal information can also be easily interpreted in the vanilla LSTM model.
The input matrix, which is slightly different from that in MLP, is defined as $W^1 =( (W^f)^{\top}, (W^{in})^{\top},(W^o)^{\top},(W^c)^{\top} )^{\top}$, controlling how the past time series affect the forgot gates, input gates, output gates, and cell updates. 
Granger-causal series can be selected based on a group lasso penalty across columns of $W^1$.
In the end, to optimize the non-convex optimization objectives in either cMLP or cLSTM, the proximal gradient descent\cite{parikh2014proximal} is used, which leads to the exact zeros in the input matrix. This property in the optimization procedure meets the requirement for interpreting Granger non-causality in the framework.
To infer the network topology of Granger causality, $d$ models need to be trained with each variable as a response.

Another sample-efficient architecture economy-SRU (\textbf{eSRU}) is proposed in~\cite{MTS/Granger/iclr20_esru}. 
It leverages Statistical Recurrent Units (SRUs) \cite{MTS/Granger/icml17_sru} to model the observed time-series data. 
Here SRUs are a special type of RNNs designed for MTS with time-delayed and nonlinear dependencies and therefore also suited for extracting the network topology of nonlinear Granger causal relations. To be specific, it suffers less from the vanishing and exploding gradient issues owing to an ungated architecture and is able to model both short and long-term temporal dependency among multivariate time series by maintaining multi-time scale summary statistics.
Similar to model-based approaches like cLSTM, the measure of Granger causal relationships can be derived from the input-layer weight parameters of the SRUs.
However, due to the common issue of data scarcity in the causal inference problem, the original framework suffers from overfitting.
Additionally, two modifications are implemented as a remedy for overfitting in eSRU.

\textbf{DL-extensions with Low-rank Approximation.}
The scalable causal graph learning (\textbf{SCGL}) framework is proposed in~\cite{MTS/Granger/SCGL_CIKM_XuHY19}.
The authors first deconstruct data nonlinearity into two types (\ie univariate-level and multivariate-level nonlinearity), which are modeled separately.
The key idea of SCGL is that learning the full size of the adjacency matrix $A \in \mathbb{R}^{d \times d}$ would be unscalable when the size of variables $d$ is quite large.
In practice, the relationship of variables is low-rank in hidden space~\cite{MTS/Granger_confounder/SPLN_automatica_ZorziC17, DBLP:journals/automatica/ChiusoP12}. 
Therefore, it's natural to approximate $A$ via a $k$-rank decomposition, where $k<d$.
The low-rank approximation reduces the noise influence in causal discovery and provides interpretability in downstream time series analysis~\cite{Application/SCGL_use_cikm_HuangXYYWX20}.


\textbf{DL-extensions with Self-explaining Networks.} For better interpretability, the generalized vector autoregression (\textbf{GVAR}) model \cite{MTS/Granger/iclr21_GVAR_MarcinkevicsV} is proposed.
It's based on an extension of self-explaining neural networks \cite{MTS/Granger/nips18_SENN_Alvarez-MelisJ}.
The self-explaining neural networks are inherently interpretable models motivated by restricted properties, and follow the form:
\begin{equation}
    f(\mathbf{x}) = g( \theta (\mathbf{x})_1h(\mathbf{x})_1, ..., \theta (\mathbf{x})_kh(\mathbf{x})_k), \nonumber
\end{equation}
where $g(\cdot)$ and $\mathbf{h}(\mathbf{x})$ denote a link function and the interpretable basis concepts, respectively.
Combined with the vector autoregression model, which is often specified in Granger causal inference, the GVAR model is given by
\begin{equation}
   \mathbf{x}^t = \sum_{l=1}^{\tau} \Psi_{\theta_l}(\mathbf{x}^{t-l})\mathbf{x}^{t-l} + \mathbf{u}^t, \nonumber
\end{equation}
where $\Psi_{\theta_l}: \mathbb{R}^d \to \mathbb{R}^{d\times d}$ is a neural network parameterized by $\theta_l$, of which the output is the matrix corresponding to the strength of influence. In detail, the strength of influence $x^{t-l}_i \to x^t_j$ is measured by the component $(j,i)$ of $\Psi_{\theta_l}(\mathbf{x}^{t-l})$.
The loss function consists of three terms: the MSE loss, a sparsity-inducing regularization (can be chosen from \ref{tab:lasso_sparsity_overview}), and the smooth penalty, which is defined as follows: 
\begin{equation}
   \frac{1}{T-\tau} \sum_{t=\tau+1}^T || \mathbf{x}^t - \widehat{\mathbf{x}}^t||_2^2 +  \frac{\lambda}{T-\tau}    \sum_{t=\tau+1}^T R(\Psi_t) + \frac{\gamma}{T-\tau-1} \sum_{t=\tau+1}^{T-1} || \Psi_{t+1} - \Psi_t  ||_2^2 , \nonumber
\end{equation}
here $\{ \mathbf{x}^t \}_{t=1}^T$ is the observed d-variate time series whereas $\widehat{\mathbf{x}}^t$ is the one-step forecast made by the GVAR model.
Now that the interpreting matrices for each time point $t$ can be derived via $\Psi_{\widehat{\theta}_k}( \mathbf{x}^t)$, the signs of Granger causal effects and their variability in time can also be assessed. 
Furthermore, a procedure of GVAR based on the heuristics of time-reversed Granger causality \cite{MTS/Granger/timereversal_GVAR_WinklerPBMH16}, which expects the relationships to be flipped on time-reversed data, is leveraged to improve the stability of the inferred structures.   
Compared to the aforementioned methods, such as cMLP, cLSTM, eSRU and MPIR, another key difference is that these methods require training $d$ neural networks, whereas GVAR requires training $2\tau$ networks.

\textbf{DL-extensions with Attention Mechanisms.} The temporal causal discovery framework (\textbf{TCDF}) is introduced in \cite{MTS/Attention/TCDF_NautaBS19}, which utilizes attention-based dilated CNN. 
This framework consists of $d$ independent attention-based CNNs with the same architecture but different target variable $X_j$.
For each target variable, a neural network is proposed to derive prediction, attention scores and kernel weights. Intuitively, a high attention score on $X_i$ while forecasting $X_j$ indicates the former contains prediction information towards the latter.
A permutation-based procedure is additionally provided for evaluating variable importance and identifying significant causal links.
TCDF can discover self-causation and time delays between cause and effect. Besides, by assuming that the bidirectional causal relations can not be instantaneous, it can also detect the presence of hidden confounders with equal delays.

Besides, an interpretable multi-variable LSTM with mixture attention is proposed in IMV-LSTM \cite{MTS/Attention/IMV_LSTM19_0002LA19, MTS/Attention/IMV_LSTM18_GuoLL18} to extract variable importance knowledge.
And it's widely used as a baseline for causal discovery in multivariate time series.
However, the topic on attention and its interpretation is to some extent still a controversial and inconclusive topic \cite{DBLP:conf/naacl/JainW19, DBLP:conf/emnlp/WiegreffeP19, DBLP:conf/lrec/GrimsleyMB20}.  
Especially in the context of Granger causal explanation, the naive-trained soft attention mechanisms are noted \cite{DBLP:conf/icml/SundararajanTY17,  MTS/Attention/aaai_AME_SchwabMK19, MTS/Attention/icdm_InGRA_ChuWMJZY20} to provide no incentive to yield accurate attributions.
In \cite{MTS/Attention/aaai_AME_SchwabMK19}, Granger causal attention weights are introduced based on the measures named as the mean Granger-causal error.
The decrease in error when adding $i$ can be computed as:$ \Delta \varepsilon_{X, i} =    \varepsilon_{X \textbackslash \{i\}} - \varepsilon_X$, given the auxiliary prediction error $\varepsilon_X, \varepsilon_{X \textbackslash \{i\}}$  with and without any information from the $i$-th variable. Then the Granger-causal attention factor can be computed as:$ \omega_i(X) = \frac{ \Delta \varepsilon_{X, i} }{  \sum_{j=1}^d  \Delta \varepsilon_{X, j} } $.
The attention factor $\omega_p(X)$ is able to capture Granger causality, which is zero if $p$-th time series is Granger noncausal for the target series.

\textbf{DL-extensions with Recurrent Variational Autoencoders.}
Recently, the causal recurrent variational autoencoder (\textbf{CR-VAE})~\cite{MTS/Granger/CR_VAE2023} is proposed, where a generative model incorporates Granger causal learning into the data generation process.
By preventing encoding future information before decoding, the encoder of CR-VAE obeys the principle of Granger causality
To be specific, given time lag $\tau$, a CR-VAE model can be written as:
\begin{equation}
    \hat{\mathbf{x}}^{t-\tau : t} = D_{\theta}(\mathbf{x}^{t-\tau:t-1}, E_{\psi}(\mathbf{x}^{t-2\tau-1:t-\tau-1})  ) + \epsilon^t,
   \nonumber
\end{equation}
where $E_{\psi}, D_{\theta}$ represent encoder and decoder.
Another distinct to the classical recurrent VAE is that the CR-VAE leverages a multi-head decoder where the $i$-th head is designed for generating $\mathbf{x}_i$.
Besides, an error-compensation module is leveraged to capture instantaneous effects.
The CR-VAE is not only able to extract causal relations, but also conduct the data-generating process in a transparent manner benefiting from the learned causal matrix.

\textbf{DL-extensions with Inductive Modeling.} The problem of methods with inductive modeling is slightly different from the above methods, where MTS data from massive individuals, which entails different causal mechanisms but shares common structures, is collected.
The goal is to train a model on samples with heterogenous structures to discover Granger causal relations from each individual.
Two approaches with inductive modeling are reviewed here.

An inductive Granger causal modeling (\textbf{InGRA}) is proposed in \cite{MTS/Attention/icdm_InGRA_ChuWMJZY20}, combined with Granger causal attention \cite{MTS/Attention/aaai_AME_SchwabMK19} and prototype learning. 
As there often exist real-world scenarios where massive multivariate time series data is collected from heterogeneous individuals sharing commonalities.
Instead of training one or a set of models for each individual, InGRA trains a global model for individuals potentially having different Granger causal structures, devoid of sample inefficiency and over-fitting issues.
Firstly, the Granger causal attention mechanism is leveraged to quantify variable-wise contributions toward prediction. As the Granger causal attention is not robust enough to reconstruct Granger causal topology from limited data of a single individual,  
InGRA secondly leverages prototype learning, of which the key idea is to solve problems for new inputs based on similarity to prototypical cases, to detect common causal structures.
As a result, the Granger causal relations and strengths between the $d-1$ exogenous variables and the target variable are inferred.

A framework termed amortized causal discovery (\textbf{ACD}) is proposed in \cite{Discussion/NewForm/ACD_LoweMSW22}, which aims to train a single model to infer causal relations across samples with different underlying causal graphs but shared dynamics.
It's an encoder-decoder framework, in which the encoder function is defined to infer Granger causal relations of the input sample whereas the decoder function learns to predict the next time-step given the inferred causal relations. 
In the implementation, a graph neural network is applied to the amortized encoder, and ACD models the functions using variational inference, which is based on the widely used neural relational inference (NRI) model \cite{Discussion/NewForm/ACD_NRI_Kipf18}. Besides, to derive a causal interpretation of the inferred edges, the proof is provided in ACD to relate the zero-edge function to Granger causality.
As a result, the causal relations of previous unseen samples can be inferred without refitting the model.

\subsection{Others}
\label{subsection:MTS_others} 

The aforementioned four categories of approaches have been the subjects of many endeavors in causal discovery research.
For the sake of completeness, we present five types of methods that are distinct from the above approaches in this subsection, including causality based on information-theoretic statistics, causal models based on differential equations, nonlinear state-space methods, logic-based methods, and hybrid methods.

\subsubsection{Causality based on information-theoretic statistics}
\label{subsection:TE} 

Causal relationships in MTS can be measured based on information-theoretic statistics.
As a model-free measure, it's widely used in constraint-based approaches (\ref{subsection:CB_mwcs}) and Granger causal models (\ref{subsection:Granger_early}).
However, its definitions and characteristics have not been detailed.
In this part, we will first introduce Transfer Entropy~\cite{TE_origin/schreiber2000measuring}, which is the original concept of information-theoretic statistics of causality, and then its variants.

Transfer Entropy \cite{TE_origin/schreiber2000measuring} is a measure of information flow or effective coupling between two processes, regardless of the actual functional relationship. 
Instead of model-based criterion, which shares the problem that the model might be misspecified, as a model-free measure, it can be combined with a variant of specific structure learning methods. 
In detail, the Transfer Entropy from $i$ to $j$ (with time lag) can be expressed as:
\begin{equation}
  \mathrm{TE}(X^t_i \to X^{t+1}_j)  = h(X^{t+1}_j | X^t_j) - h(X^{t+1}_j | X^t_j, X^t_i),  \nonumber
\end{equation}
where $h(\cdot| \cdot)$ denotes the conditional entropy. Here the term $h(X^{t+1}_j | X^t_j)$ measures the uncertainty of $X^{t+1}_j$ given information about $X^t_j$, and $h(X^{t+1}_j | X^t_j, X^t_i)$ measures the uncertainty of $X^{t+1}_j$ given information about both $X^t_j$ and $X^t_i$. Therefore, we can understand Transfer Entropy $\mathrm{TE}(X^t_i \to X^{t+1}_j)$ in the causal view of reduction of uncertainty about future dynamics of $X_j$ when the current dynamics of $X_i$ is given addition to that of $X_j$.
For Gaussian variables, the equivalence between Transfer Entropy and Granger causality is demonstrated in \cite{Back/Granger_TE_equivalence_barnett2009granger}.
Furthermore, Transfer Entropy is reformulated into a decomposition form and embedded into the framework of graphical models for multivariate in \cite{MTS/others/information/PRE_runge2012escaping}. 
In \cite{MTS/others/information/PRE_runge2012quantifying}, the causal coupling strength for multivariate time series is quantified based on a variant of transfer entropy.

Although some utilization in multivariate scenarios, Transfer Entropy suffers from the pairwise limitation. And it's reported to fail to distinguish between direct and indirect causality in networks \cite{MTS/CB/oCSE/siamads/0007TB15} .
As a remedy to pairwise limitation, Causation Entropy \cite{MTS/Others/concepts/causationEntropy/sun2014causation}, a model-free information theoretic statistic for inferring causality, is introduced.
In detail, the Causation Entropy from the set of nodes $I$ to the set of nodes $J$ conditioning on the set of nodes $C$ is defined as follows:
\begin{equation}
   \mathrm{CE}(X^t_{\mathbf{I}} \to X^{t+1}_{\mathbf{J}} | X^t_{\mathbf{C}} )  = h(X^{t+1}_{\mathbf{J}} | X^t_{\mathbf{C}}  ) - h(  X^{t+1}_{\mathbf{J}} | X^t_{\mathbf{C}}, X^t_{\mathbf{I}} ), \nonumber
\end{equation}
here $I,J,C$ are all subset of nodes $\{1,2,...,d\}$.
As a type of conditional mutual information, Causation Entropy is a generalization of Transfer Entropy for measuring pairwise relations to network relations of many variables. And similar to the equivalence relations between Transfer entropy and Granger Causality, Causation Entropy also generalizes Granger Causality and Conditional Granger Causality when applied to Gaussian variables.
However, according to its definition, this concept assumes that the hidden dynamics follow a stationary first-order Markov process as the Causation Entropy only models causal relations with time lags equal to one. 
Recently, to measure any lagged or instantaneous relations, an extension of Causation Entropy, named Greedy Causation Entropy, is proposed in \cite{MTS/others/information_criterion/uai/AssaadDG22}.

\subsubsection{Causal models based on differential equations}

Differential equations are a commonly used modeling tool in many fields, and are especially useful if measurements can be done on the relevant time scale.
Compared to the aforementioned causal models, this type of approach is specifically designed to model systems that can be well represented by differential equations~\cite{intro/ts_surveys/peters2022causal}.  
In this part, we will first review the relationships between differential equations and causal models, for both discrete and continuous time.
The first difference-based causal discovery framework will be introduced.
Then, we will give the recent advances in this type of method.

There is abundant literature \cite{intro/ts_surveys/peters2022causal, bongers2018causal} \cite{DBLP:journals/corr/abs-1911-10500, DBLP:conf/uai/MooijJS13,DBLP:conf/uai/RubensteinBMS18} discussing the relationship between differential equations and structural causal models.  
For discrete time, a difference-based causal discovery framework is first proposed in \cite{MTS/others/difference_based/uai/VoortmanDD10}.
The cross-temporal restriction is satisfied, where all causation across time is due to a derivative $\dot{x}$ causing a change in its integral $x$. 
This characteristic makes the difference-based causal model a restricted form of dynamic SEMs.
And difference-based causality learner (\textbf{DBCL}) is leveraged to extract difference-based causal models from data, which is proven to be able to identify the presence or absence of feedback loops. 
For continuous time, several theoretical endeavors have also been made to derive a causal interpretation of dynamic systems by both ordinary differential equations (ODEs) \cite{DBLP:conf/uai/MooijJS13, DBLP:conf/uai/RubensteinBMS18, UAI19/DBLP:conf/uai/BlomBM19, MTS/others/difference_based/PNAS/pfister2019learning} and stochastic differential equations (SDEs) \cite{hansen2014causal, MTS/others/difference_based/uai18/MogensenMH18}.  

More recently, under a dynamic causal system where the multivariate time series are irregularly-sampled (in infinitesimal interval of time), an algorithm called neural graphical model (\textbf{NGM}) is proposed in \cite{MTS/Others/NGM_neuralode_iclr_BellotBS22}.
In many applications, the underlying causal system of interest can be represented as a dynamic structural model as follows:
\begin{equation}
   d \mathbf{x}(t) = \mathbf{f}( \mathbf{x}(t) )dt + d\mathbf{w}(t), \ \ \ \mathbf{x}(0)=\mathbf{x}_0, \ \ \ t\in [0,T],  \nonumber
\end{equation}
where $\mathbf{w}(t)$ is a $d$-dimensional standard Brownian motion, $\mathbf{x}_0$ is a Gaussian random variable independent of $\mathbf{w}(t)$, and the function $\mathbf{f}$ describes the causal graph $G$. 
NGM is a learning algorithm based on penalized Neural Ordinary Differential Equations (neural-ODE).
The recovery of causal graph can be cast to penalized optimization problems of the form: 
\begin{equation}
   \mathrm{min}_{\mathbf{f}_\theta}  \frac{1}{n}\sum_{i=1}^n ||  \mathbf{x}(t_i)  - \hat{\mathbf{x}}(t_i)  ||_2^2,\ \ \mathrm{subject}\ \mathrm{to}\ \rho_{n,T}(\mathbf{f}_{\theta}) \ \mathrm{and} \   \hat{\mathbf{x}}(t) = \mathbf{f}_{\theta}(\hat{\mathbf{x}}(t_i))dt,   \nonumber
\end{equation}
where the observation of the systems are at irregular time points $0 \leq t_1 < ... < t_n \leq  T$.


\subsubsection{Nonlinear state-space methods}

In this part, we will first introduce the basics of nonlinear state-space methods, including the Takens theorem and Convergent Cross Mapping algorithm.
Then variants and recent advances of the original algorithm will be given, to tackle the challenges such as high sensitivity to noise, large sample demands, inconsistent results, and misidentifications.

\begin{figure*}
    \centering
	\includegraphics[width=1.0\textwidth]{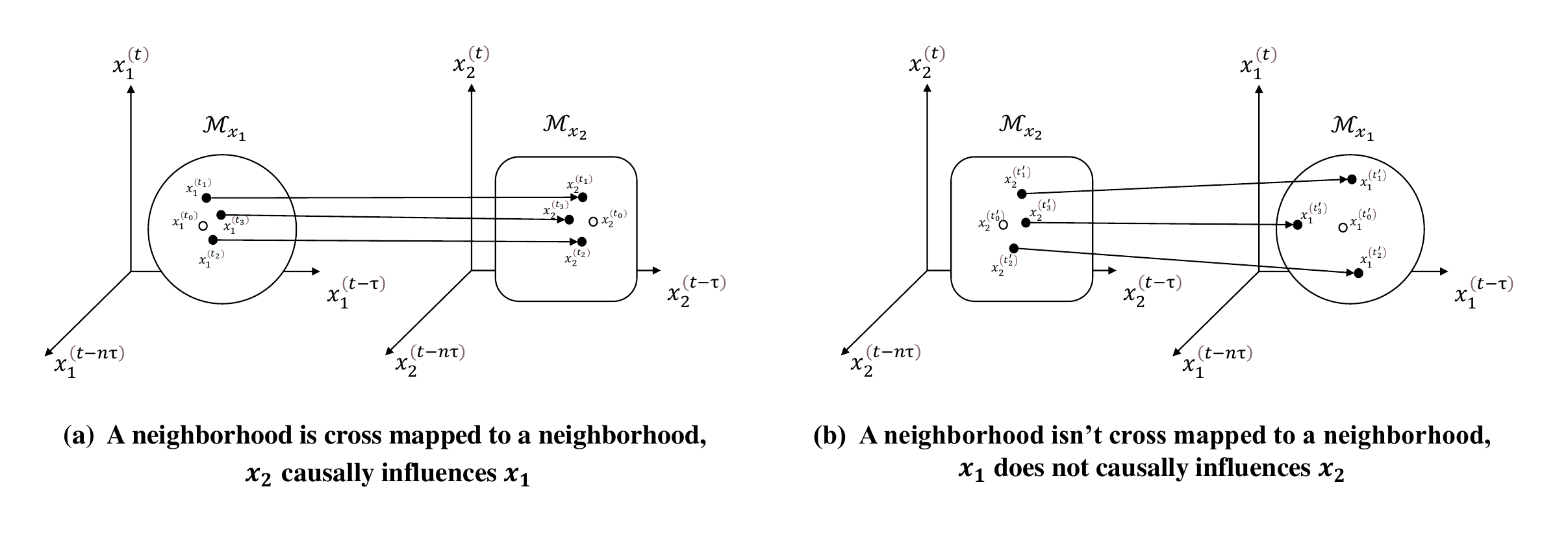}
	\caption{The illustration of the cross convergent mapping procedure.}
	\label{fig:CCM_illustration}
\end{figure*}

The state space reconstruction theory proposed by Takens \cite{takens1981detecting} provides a theoretical basis for analyzing the dynamic characteristics of nonlinear systems. 
Based on this theory, another approach for determining causality, known as Convergent Cross Mapping (\textbf{CCM}), was first proposed in \cite{MTS/CCM/work2_main_science_sugihara2012detecting}.
Developed for coupled time series, this method leverages Takens' theorem via state space reconstruction. 
In detail, given two time series $x_1^t$ and $x_2^t$, the attractor manifolds $\mathcal{M}_{x_1}, \mathcal{M}_{x_2}$ are first reconstructed using $x_1^t$ and $x_2^t$, respectively.
Secondly, causality can be detected by measuring the correspondence between $\mathcal{M}_{x_1}$ and $\mathcal{M}_{x_2}$, to be specific, by testing whether every local neighborhood defined on one manifold is preserved in the other.
Figure \ref{fig:CCM_illustration} gives the illustration of CCM. 
This methodology has been successfully applied in many fields \cite{MTS/CCM/brain_ecology_hirata2016detecting, MTS/CCM/ecology_ye2015distinguishing} where nonlinear systems are dynamically coupled.

However, there exist issues for the original CCM method, such as high sensitivity to observation noise, a requirement for a relatively large number of observations, and inconsistent results under different optimal algorithms. 
To overcome these challenges, variants of CCM based on time-lagged analysis \cite{MTS/CCM/ecology_ye2015distinguishing}, deep Gaussian process \cite{MTS/CCM/feng2019detecting} reservoir computing \cite{MTS/CCM/RCC_huang2020detecting} and neural ODE \cite{MTS/CCM/latent_CCM_iclr_BrouwerASM21} were proposed. 
Besides, most CCM-based approaches have been originally developed for bivariate analysis.
Although the same procedures may be used multiple times to ascertain the causal network among multivariate time series, the performance is not guaranteed under high-dimensional conditions \cite{MTS/CCM/huang2020systematic}.
Misidentifying indirect causations as direct ones performs one of the key challenges in multivariate settings. 
Recently, partial cross mapping (PCM), which combines CCM with partial correlation, was proposed \cite{MTS/CCM/leng2020partial} to eliminate indirect causal influences.

\subsubsection{Logic-based methods}

Another type of methodology, used for causal inference and causal discovery in time-series data, is based on logic formulas.
The original algorithm of this type of approach will first be introduced and combined with its semeiology and definition of potential causality.
Then we will give its variants and recent advances. 

In logic-based methods, temporal data can be thought of as observations of the sequence of states the system has occupied and is referred to as traces in model checking.
This line of research originates from work in \cite{MTS/logic/uai/KleinbergM09}, where causal relationships are described in terms of temporal logic formulas.
To be specific, it first leverages logic, Probabilistic Computation Tree Logic (\textbf{PCTL}), to define \textit{prima facie (potential) causality} based on temporal priority and the uplift of conditional probability. 
Given the notation in the original work, the prima facie cause is defined if the following conditions all hold: (1) $F_{>0}^{\leq \infty}c$, (2) $c \rightsquigarrow_{\geq p}^{\geq 1, \leq \infty} e$, and (3) $F_{<p}^{\leq \infty}e$, implying that there may exist any number of transitions between $c$ and $e$ and the sum of a set of path probabilities are at least $p$. 
To separate the underlying prima facie (potential) causes into genuine and spurious causes, the notion of $\epsilon$-\textit{insignificant cause} is introduced by computing the average difference in probabilities for each prima facie cause of an effect in relation to all other prima facie causes of the effect:
\begin{equation}
  \epsilon_{avg}(c,e) = \frac{ \sum_{x \in X \textbackslash  c}  \epsilon_x(c,e)  }{|X|},   \nonumber
\end{equation}
where $\epsilon_x(c,e) = P(e|c \wedge x) - P(e| \lnot c \wedge x)$. 
A prima facie cause $c$ is an $\epsilon$-\textit{insignificant cause} of $e$ if $\epsilon_{avg}(c,e) < \epsilon$.
The value of $\epsilon$ is chosen based on empirical null hypothesis testing by assuming: (1) data contains two classes, significant and insignificant, (2) the significant class is relatively small to the insignificant class.  
And false discovery rate control is implemented simultaneously. 
This methodology has also applications in fields \cite{MTS/KleinbergS/kleinberg2013causality}. 

To expand the methodology to the condition where both discrete and continuous components exist, PCTLc is introduced in \cite{MTS/logic/ijcai/Kleinberg11} to express temporal and probabilistic properties involving discrete and continuous variables, and the significance of relationship in the continuous case is validated via conditional expectation of an effect instead of conditional probability. 
Besides, a variant \cite{MTS/logic/flairs/HuangK15} of this logic-based approach was proposed to improve the accuracy of causal discovery and enable faster computation of causal significance, by showing the computational complexity can be reduced under several conditions.
Following this line of temporal logic form, a recent work \cite{MTS/logic/jair/CostaD21} combines the idea of decision trees and reconsiders the problem of causal discovery to extract temporal causal sequence relationships from real-time time series.

\subsubsection{Hybrid methods: combining score-based and constraint-based approaches}

Hybrid approaches are proposed for the benefit of combining the strengths of both constraint-based (\ref{subsection:CB}) and score-based (\ref{subsection:SB}) approaches.
We cover two parts of hybrid methods, including methods based on max-min hill-climbing heuristics, and methods incorporating the conditional independence tests to improve the local search.

Some researchers develop hybrid approaches based on max-min hill-climbing heuristics \cite{SB/MMH_DB/ml/TsamardinosBA06,MTS/SB/MMHO_DBN/origin/cibcb/LiN13, MTS/SB/MMHO_DBN/tcbb/LiCZN16}. As hybrid local learning methods, Max-Min approaches fuse concepts from both constraint-based techniques to limit the space of potential structures and search-and-score Bayesian methods to search for an optimal structure. They are originally leveraged in the structure learning of BN for static data \cite{SB/MMH_DB/ml/TsamardinosBA06}. The Max-Min hill-climbing Bayesian network (MMHO-DBN), introduced in \cite{MTS/SB/MMHO_DBN/tcbb/LiCZN16}, learns the structure of DBN based on an extension of the max-min hill-climbing heuristic and is leveraged in the modeling of real gene expression time-series data. 

There are also hybrid approaches that combine conditional-independence tests and local search to improve the criterion score \cite{NONMTS/hybrid/pgm/OgarrioSR16,MTS/CB_FCI_SVAR_FCI_MalinskyS18}. Greedy FCI (GFCI) \cite{NONMTS/hybrid/pgm/OgarrioSR16} is a hybrid score that combines features of GES with FCI. SVAR-GFCI \cite{MTS/CB_FCI_SVAR_FCI_MalinskyS18} extends this method to causal structure learning from time series. In \cite{MTS/SB_or_hybrid/FASK_sanchez2019estimating}, both a variant of the PC-stable algorithm referred to as Fast Adjacency Skewness (FASK), and a hybrid two-step algorithm is proposed for extracting causal relations for time-series data.

\section{Causal Discovery from Event Sequences}\label{sec:event}

An important assumption in multivariate time series is that the timestamps are discrete and the time intervals are fixed. However, in the real-world scenario, the vast majority of events will not occur at fixed intervals. In consequence, we need to come up with some methods to deal with these irregular and asynchronous data. We can construct event sequences as $\{(t_1, e_1),(t_2,e_2),...\}$, where the first dimension represents the time at which the corresponding event happens, and the second dimension stands for the corresponding event type. In this section, we will focus on inferring causal relationships in event sequences.
First, the multivariate point process is introduced, which is preliminary for causal discovery in event sequences.
Then, we review approaches based on the Granger causal model, which are well-developed. 
Lastly, other approaches including constraint-based and score-based methods are given.

\subsection{Multivariate Point Process}
\label{subsec:MPP}

    An event sequence records the occurrence of one specific type of event (or `event type' for simplicity). Meanwhile, we can characterize an event sequence through a point process. To discover the relationships between different types of events, we consider its high-dimensional cases, which is to model event sequences through \textit{Multivariate Point Processes (MPPs)}. Therefore, our problem can be defined as inputting a set of point processes, where each point process represents an event sequence, and outputting a causal graph $G$ established by different processes. In the causal graph $G$, each node represents a point process, and each directed edge captures a directed interaction from one point process to another.
    In this part, we will detail MPPs, including their intensity functions and log-likelihood functions.

    \textbf{Intensity Functions of MPPs.} A temporal point process is a stochastic or random process composed of a time series of binary events that occur in continuous time~\cite{daley2003introduction}. MPPs are high-dimensional point processes, implying that they can involve multiple types of events. 
    $\mathcal{E}=\{1, …, E\}$ is the set of event types. 
    The occurring time of these events $\{t_1, t_2, …, t_n|t_i\in[0,T]\}$ are unevenly-distributed. 
    The multivariate point process with $E$ types of events can be represented by $E$ counting processes $\{N_e\}_{e=1}^E$, where $N_e = \{ N_e(t)|t \in [0,T]  \}$. 
    The core of a point process is its conditional intensity function, in which the process’s pattern is captured. A type-$u$ intensity function can be defined as the expected instantaneous rate of type-$e$ event’s occurrence given the history:
    \begin{equation}
       \lambda_e(t)=\frac{\mathbb{E}[dN_e(t)|\mathcal{H}_t]}{dt} \nonumber
    \end{equation}
    Here $\mathcal{H}_t = \{(t_i,e_i)|t_i\textless t, e_i\in\mathcal{E}\}$ represents all types of events happened before time $t$.

    \textbf{Log-likelihood Functions of MPPs.} Next, we show the relationship between the intensity function and the Probability Density Function(PDF) of the joint distribution: $f((t_1,e_1),...,(t_n,e_n)|(t_0,e_0))$. Using the chain rule, there is $f((t_1,e_1),...,(t_n,e_n)|(t_0,e_0))=\prod_{j=1}^{n}f((t_j,e_j)|\mathcal{H}_{t_j})$. Then, we can set up the likelihood function for estimating the joint distribution:
    \begin{equation} \label{eq2}
        \widetilde{L_0} \triangleq \sum_{j=1}^{n}ln f(t_j|e_j,\mathcal{H}_{t_j})+\sum_{j=1}^{n}ln f(e_j|\mathcal{H}_{t_j})
    \end{equation}
    Since the goal is to infer the causal relationship between different events, here we focus on the first term and omit the second term: $L_0 \triangleq \sum_{j=1}^{n}ln f(t_j|e_j,\mathcal{H}_{t_j})$. The intensity function reflects the expectation of the event happening in $[t, t+dt]$ given the information of $\mathcal{H}_{n-1} \triangleq \mathcal{H}_{t_{n-1}}$. Similar to the calculation of the force of mortality in survival analysis, there holds,
    \begin{equation}
    \begin{aligned}
        \lambda_e(t|\mathcal{H}_{n-1})&=\frac{\mathbb{E}[N_e(t+dt)-N_e(t)|\mathcal{H}_{n-1}]}{dt}\\
        &=\frac{\int_{t_{n-1}}^{t+dt}f(l|e,\mathcal{H}_{n-1})dl-\int_{t_{n-1}}^{t}f(l|e,\mathcal{H}_{n-1})dl}{1-\int_{t_{n-1}}^{t}f(l|e,\mathcal{H}_{n-1})dl}\\
        &=-\frac{d}{dt}ln(1-\int_{t_{n-1}}^{t}f(l|e,\mathcal{H}_{n-1})dl)\nonumber
    \end{aligned}
    \end{equation}
    Integrating the equation above and substituting $f$'s expression into~\ref{eq2}, we have,
    \begin{equation} \label{eq3}
        L_0 = \sum_{j=1}^{n}\{ln\lambda_{e_j}(t_j|\mathcal{H}_{j-1})-\int_{t_{j-1}}^{t}\lambda_{e_j}(l|\mathcal{H}_{j-1})dl\} 
    \end{equation}
    We have briefly introduced Multivariate Point Processes and constructed likelihood functions to characterize MPPs in the above. Next, we aim to discover the causal relationships within MPPs using Granger-based, as well as constraint-based and score-based methods.\\

\subsection{Granger Causality Based Approaches}
\label{subsec:event_Granger}

In this subsection, we consider the task to infer Granger causalities in event sequences. Similar to that in MTS, we say $e_j$-type events Granger cause $e_i$ if $\{e_j(t)|t<t_0\}$ is useful in forecasting $e_i(t)$.
The detailed methods can be categorized according to the following model specifications, \ie, GLM point process, Hawkes process, Wold process, and neural point process.

    \subsubsection{Methods for GLM point processes}
    We first introduce causal discovery approaches for event sequences which are modeled via \textit{Generalized Linear Model (GLM)} of point processes~\cite{doi:10.1152/jn.00697.2004}.
    The GLM assumes that the logarithm of the intensity function has a linear format, i.e., $ln\lambda_e(t|\mathcal{H}_{n-1})=\beta_0+\beta_1 X_1+\beta_2 X_2$. Specifically, in our mission, the intensity functions follow,
    \begin{equation} \label{eq4}
        ln\lambda_i(t|\gamma_i,H_i(t)) = \gamma_{i,0}+\sum_{j=1}^{J}\sum_{m=1}^{M_i}\gamma_{i,j,m}R_{j,m}(t)
    \end{equation}
    Here $\gamma_{i,0}$ can be interpreted as the background intensity of event $e_i$, $\gamma_{i,j,m}$ is the intensity on type-$e_i$ events triggered by type-$e_j$ events and $R_{j,m}(t)$ is the number of occurrence of $e_j$-type events happened in $[t-mW,t-(m-1)W]$ ($W$ is a small number which refers to the length of time range). By looking at the sign of $\sum_{m=1}^{M_i}\gamma_{i,j,m}$, we can distinguish whether type-$e_j$ events have excitatory or inhibitory effects on type-$e_i$ events.
    
To infer the Granger Causality between type-$e_j$ and type-$e_i$ events, we substitute~\ref{eq4} into the likelihood function~\ref{eq3}. Next, we follow a simple thought that we can exclude a certain type of event and then infer the Granger causality by comparing its intensity with the original case. Specifically, we obtain both the likelihood of type-$e_i$’s occurrence with and without type-$e_j$’s effect: $L_i(\gamma_i)$, $L_i(\gamma_i^j)$. Then, consider that $\sum_{m=1}^{M_i}\gamma_{i,j,m}$ is an indicator of the effection type, the Granger causality from type-$e_j$ to type-$e_i$ events can be proposed as~\cite{10.1371/journal.pcbi.1001110}:
    \begin{equation}
    \begin{aligned}
        \phi_{ij} &= -sign(\sum_{m=1}^{M_i}\gamma_{i,j,m})\Gamma_{ij}\\
        &= -sign(\sum_{m=1}^{M_i}\gamma_{i,j,m})log\frac{L_i(\gamma_i^j)}{L_i(\gamma_i)} \nonumber
    \end{aligned}
    \end{equation}
    Apparently, there exists $L_i(\gamma_i) \geq L_i(\gamma_i^j)$, hence, $\Gamma_{ij} = log\frac{L_i(\gamma_i^j)}{L_i(\gamma_i)} \leq 0$. Only if '$<$' is satisfied, type-$e_j$ events will be the Granger cause to type-$e_i$ events. In the next step, Kim et al.~\cite{10.1371/journal.pcbi.1001110} presented a significance test of these causal interactions by conducting $H_0$ hypothesis: $\theta_0=\gamma_i^j$ and $H_1$ hypothesis: $\theta_1=\gamma_i$. 
    Passing through an FDR significance test, the final causal relationships could be estimated by $\widetilde{\phi_{ij}}$: (1) type-$e_j$ events are an excitatory cause of type-$e_i$ events when $\widetilde{\phi_{ij}}>0$, (2) the cause is inhibitory when $\widetilde{\phi_{ij}}<0$, (3) there exists no causal relationship between type-$e_j$ and type-$e_i$ events when $\widetilde{\phi_{ij}}=0$.



    \subsubsection{Methods for Hawkes processes}
    In this part, we review methods for Hawkes process. As a particular type of point process, the basics of the Hawkes process are first given.
    Then we detail approaches based on MLE to infer causal relations, including (1) parameterization strategies and (2) regularization methods. Next, we review other estimation approaches, including (1) graphical event models, (2) generalized method of moments, (3) event sequence separation, and (4) minimum description length.
    We note that there exists a plethora of literature in this category since a natural match-up between Granger causality and Hawkes processes.

    The Hawkes process is a type of point process that has a fixed form of intensity function:
    \begin{equation} \label{eq6}
        \lambda_{e_i}(t) = \mu_{e_i} + \sum_{e_j=1}^{E}\int_{0}^{t}\phi_{e_ie_j}(s)dN_{e_j}(t-s)
    \end{equation}\\
    Here, $\mu_{e_i}$ is called the baseline intensity, which can only be affected by exogenous events, hence, is a constant over time. And $\phi_{e_ie_j}(s)$, the impact function, measures the decay of the excitement on future type-$e_i$ events triggered by historical type-$e_j$ events. That is to say, it captures the endogenous intensity from $e_j$ to $e_i$. Considering the similarity between definitions of $\phi$ and the Granger causality, we can directly infer Granger causality by analyzing $\phi$:
    
    \begin{prop} \label{prop1}
        (Eichler et al.~\cite{https://doi.org/10.1111/jtsa.12213}, 2017)
        \begin{equation}
            e_j \ \text{does not Granger-cause} \ e_i \iff \phi_{e_ie_j}(s)=0, \forall s \in R  \nonumber
        \end{equation}  
    \end{prop} 
    
    Therefore, we aim to model $\phi_{e_ie_j}(t)$ for each event and all $t\in R$. However, due to the complexity and heterogeneity of event sequences, this mission could be extremely difficult to accomplish.
    Zhou et al.~\cite{pmlr-v31-zhou13a} parameterize $\phi_{e_ie_j}(s)$ as $a_{e_ie_j}g(s)$. By this means, $\phi$ is split into events-interaction and time-decaying parts. 
    
    \textbf{MLE Approaches.} The Maximum Likelihood Estimation (MLE) can be performed for estimating parameters in~\ref{eq6}. We take $\lambda$'s expression into~\ref{eq3}, which results in the corresponding likelihood function: $L(A,\mu)$. Here, $A$ is composed of $(a_{e_ie_j})$, and $\mu$ is built up by $\mu_{e_i}$. Next, consider that in real-world scenarios, most events can only influence a small fraction of other events, and the community structures in the influence networks tend to be low-ranked~\cite{pmlr-v31-zhou13a}, we should add penalty entries to the MLE loss function. Specifically, the following objective function can be constructed in order to achieve matrix $A$'s low rank and sparsity:
    \begin{equation}
        \underset{A\geq0,\mu\geq0}{min}-L(A,\mu)+\lambda_1\Vert A \Vert_*+\lambda_2\Vert A \Vert_1  \nonumber
    \end{equation}
    
    Here, $\Vert \cdot \Vert_*$ is the nuclear norm, of which performance in reducing the matrix's rank has been proven. And $\Vert \cdot \Vert_1$ is the L1 norm. It can enforce matrix A to gain more sparsity. $\lambda_1$, $\lambda_2$ are parameters to control the strength of these two penalties. We denoted the object function as $f(A,\mu)$. Apart from this, an EM-based algorithm could be conducted to solve the optimization problem for $A$ and $\mu$. In details, Zhou used the surrogate function $Q(A,\mu;A^{(m)},\mu^{(m)})$ as a tight upper bound of $f(A,\mu)$. By optimizing $Q(A,\mu;A^{(m)},\mu^{(m)})$ iteratively, $f(A,\mu)$ was forced to decrease and, thus, successfully optimized. We summarize this in algorithm \ref{alg:Algorithm1}.


    \textit{(1) Parameterization Strategies}: The parameterization method $\phi_{e_ie_j}(s)$ = $a_{e_ie_j}g(s)$ mentioned above may suffer from bad performance if the data do not fit its strong assumption. Therefore, to gain robustness in different types of event sequences, Xu et al.~\cite{pmlr-v48-xuc16} came up with a strategy to choose a family of basic functions and used their linear combination, $\sum_{m=1}^{M}a_{e_ie_j}^m \kappa_m(s)$, to model the targeted intensity function. 
    
    However, in \textbf{NPHC}~\cite{pmlr-v70-achab17a}, Achab et al. put forward that the \emph{basic function} strategies would have extraordinary computing complexities when there exist too many types of events (i.e., $E$ is large). Given that our goal is only to infer the Granger causality, there is no need to totally parameterize the Hawkes process, and thus, we only need to estimate the corresponding integral $\int_{0}^{+\infty}\phi_{ee'}(s)ds$. Achab denoted the integral as $g_{ee'}$ while $(g_{ee'})$ formed matrix G. Then, from Eichler's proof~\cite{https://doi.org/10.1111/jtsa.12213} as well as $\phi_{ee'}(s)>0, \forall s>0$, it is clear that $(g_{ee'})=0 \iff e' \ \text{does not Granger-cause} \ e$.
    
    Other works considered the underlying topological relationships within event sequences. In THP~\cite{Cai2021THPTH}, Cai et al. assumed the existence of a hidden undirected graph structure $G_N$ between events. And the corresponding intensity function is formed as $\lambda_{e_i}(n,t)=\mu_{e_i}+\sum_{e_j\in E}(g_{e_ie_j}*s_{e_i,e_j,t})_{G_N}(n)$. Here, $g_{e_ie_j}$ is the graph convolution kernel that could capture the effect from the graph neighbors. And $s_{e_i,e_j,t}$ is the time convolution kernel representing the sum of the past impact function $\phi_{e_ie_j}(s), s<t$. This is based on the assumption that the hidden topological structure will not change during the process.

    \textit{(2) Regularization Methods}: In an aforementioned method we presented $A$'s nuclear norm $\Vert A \Vert_*$ and L1 norm $\Vert A \Vert_1$ as regularizers. And in \emph{basic function} methods, a special sparse-group-lasso regularizer~\cite{simon2013sparse} is applied to fit their summation parameterization. Specifically, Xu et al.~\cite{pmlr-v48-xuc16} conducted a group-lasso penalty as well as a lasso penalty simultaneously in order to enforce $a_{e_ie_j}^m=0$ for all $m$, i.e., group sparsity, in addition to a regular sparsity for all the entries $a_{e_ie_j}^m$. Nevertheless, in $L_0$ Hawkes~\cite{NEURIPS2021_15cf7646}, Ide et al. proved that EM-based MLE algorithms with L1-regularization cannot offer sparse solutions mathematically. Hence, their sparse solutions can appear only as numerical artifacts. Sequentially, Ide presented an L0-regularized EM-MLE algorithm to circumvent this problem. Here, the L0-norm $\Vert A\Vert_0$ indicates the number of non-zero entries in matrix $A$.
    
    Similar to the aforementioned topological parameterization strategies, Xu et al.~\cite{pmlr-v48-xuc16} considered the underlying topological relationships between event types in constructing our regularizers. Specifically, pairwise similarity $\sum_{e_i=1}^{E}\sum_{e_j\in C_e}\Vert a_{e_i\cdot} - a_{e_j\cdot}\Vert_F^2 + \Vert a_{\cdot e_i} - a_{\cdot e_j}\Vert_F^2$ could be presented in order to enforce that similar events could have similar intensity functions. However, we must add that this regularizer requires a predefined cluster structure and thus can be optimized.
    
    Prior domain knowledge could be of great use when discovering the Granger causality. Due to the event sequences' high dimensionality and heterogeneity, existing algorithms regularly suffer from underfitting and poor interpretability. Hence, it is natural to consider adding domain knowledge from humans to the causal-inferring model. In specific, a bottom-up visualization model with user feedback was established~\cite{9222294}. Jin et al. set up their based model with the traditional MLE method in MLE-SGLP. During the training process, the user could either confirm or remove a causal relation depending on their domain knowledge from the network. And the model will change its optimization target corresponding to the user's choice. For example, in accordance with the idea in MLE-SGLP~\cite{pmlr-v48-xuc16}, Jin~\cite{9222294} constructed their intensity function as $\phi_{e_ie_j}=\sum_{m=1}^{M}a_{e_ie_j}^m \kappa_m(s)$, and set $a_{e_ie_j}$ as $[a_{e_ie_j}^1,...,a_{e_ie_j}^n]$. Correspondingly, their objective function could be: $\underset{\mu,\alpha}{argmin}\quad -L+\alpha\sum_{e_i,e_j}\Vert a_{e_ie_j}\Vert_2$. After the user made their choice to either confirm or delete edges in the causal graph $\hat{G}$, Jin updated the object function as follows:
    \begin{equation} \label{eq9}
    \begin{split}
        \underset{\mu,\alpha}{argmin}\quad -L+\alpha_v\sum_{e_i,e_j}\Vert a_{e_ie_j}(\hat{G})\Vert_2 \\
        \text{s.t.} \quad a_{e_ie_j}=0 \quad\text{for}\quad (e_j\rightarrow e_i) \notin \hat{G}
    \end{split}
    \end{equation}
    
    \begin{equation} \label{eq10}
        a_{e_ie_j}(\hat{G})=
        \begin{cases}
            0; \quad \text{if} \quad (e_j\rightarrow e_i) \quad \text{is confirmed}\\
            a_{e_ie_j}; \quad \text{otherwise}
        \end{cases}
    \end{equation}
    
    Here the constraints in~\ref{eq9} fit the removal operations, and the updates in~\ref{eq10} represent the user's confirmations.
    
    \textbf{Other estimation approaches}
    
    \textit{(1) Graphical Event Models}: The aforementioned methods use Maximum Likelihood Estimation to model the Hawkes processes of event sequences. However, these attempts lack interpretability and require fine-tuning processes for parameters to achieve a good performance. Therefore, entirely data-driven, graph-based, and dependency-captured Graphical Event Models (GEMs) could be presented to infer Granger causalities in the event sequences.
    
    We will elaborate more on GEM's attributes in~\ref{subsec43}. Here, we only focus on its relationship with the Granger causality. Suppose there is a directed graph $\mathcal{G}=(\mathcal{E},\mathcal{A})$, in which the edges represent dependencies between different event types. For each event-type $e$, we assumed that its conditional intensity could only be affected by its parent type, i.e., it follows $\lambda_e(t|h_t)=\lambda_e(t|[h_t]_{P_a(e)})$, where $P_a(e)\subseteq\mathcal{E}$ is $e$'s parent event in a graph $\mathcal{G}$, and $[h_t]_{P_a(e)}$ is the history of events which types are listed in the set $P_a(e)$. In accordance with~\ref{prop1}, there holds,

    \begin{prop} \label{prop2}
        \text{(Granger Causality in GEMs, Yu et al., 2020~\cite{pmlr-v138-yu20a})}\\
        For two event types $e_i$ and $e_j$ in $\mathcal{G}=(\mathcal{E},\mathcal{A})$, $e_j$ does not Granger-cause $e_i \iff \phi_{e_ie_j}(t)=0, \forall t>0 \iff e_j \notin Pa(e_i)$
    \end{prop}

    Hence, one can apply traditional score-based structure learning methods to discover the Granger causality. For example, BIC scores can be presented for learning the optimized graph $\mathcal{G^*}$. The optimization approach is consistent. At the same time, Yu conducted a Forward-Backward Search (FBS) to learn the parent types of a certain event type independently~\cite{pmlr-v138-yu20a}. The Forward-Backward Search with BIC scores is proved to be sound and complete for a family of GEMs~\cite{pmlr-v51-gunawardana16}.
    
    \textit{(2) Generalized Method of Moments}: In NPHC, the optimization object is a matrix $G=(g_{ee'})=(\int_{0}^{+\infty}\phi_{ee'}(s)ds)$. Therefore, the Generalized Method of Moments (GMM) can be used to address this problem~\cite{hall2004generalized}. Achab et al. presented a GMM-based NPHC algorithm to model the first, second, and third-order cumulants of matrix $G$~\cite{pmlr-v70-achab17a}. Afterward, the Granger causality could be directly attained from $G$. This moment estimation approach is proven consistent and robust to certain observation noise~\cite{pmlr-v139-trouleau21a}. However, this approach might receive poor results in specific datasets, e.g., datasets with long tails. That is mainly due to GMMs' general issue: they can only capture the information within a statistical distribution's moments.
    
    \textit{(3) Event Sequences Separation}: Another intriguing idea is separating the event sequences into multiple sub-sequences and applying the Hawkes Process model in each sub-sequence correspondingly. In GC-nsHP~\cite{CHEN202222}, Chen et al. divided the event sequences $\mathcal{H}_{n}=[(t_1,e_1),...,(t_n,e_n)]$ into $K$ different patterns, where $K$ should be predefined according to its applying scenario. 'Events' in the same pattern are supposed to build up a stationary sub-process of $\mathcal{H}_{n}$. Then, $K$ different Hawkes processes were established specifically for $K$ patterns, and the Granger causality can only be learned inside each pattern. Within each iteration, a Viterbi-path-based pattern reassignment algorithm and an EM-MLE-based parameter-updating algorithm were conducted alternately. In the parameter updating part, consider that $X_{t-1}$ and $X_{t}$ are more likely to be in the same pattern, Chen added a penalty term to help to put the adjacent sequences into the same pattern.

    \textit{(4) Minimum Description Length}: Following the Minimum Description Length (MDL) principle~\cite{rissanen1998stochastic, grunwald2019minimum}, Jalaldoust et al. conducted a trade-off between the goodness-of-fit and the model complexity~\cite{Schindler_Plant_2022}. In detail, they partitioned the parameter space $\Theta$ into $\{\Theta_\gamma:\gamma \in \Gamma\}$, defined a luckiness function $v:\Theta \rightarrow \mathbb{R}$, and set the normalized maximum likelihood distribution for each model $\gamma \in \Gamma$ to be:
    \begin{equation}
        p_{v|\gamma}^{NML}(x)=\frac{max_{\theta \in \Theta_\gamma}p(x|\theta)v(\theta)}{\int_{x\in\mathcal{X}}max_{\theta \in \Theta_\gamma}p(s|\theta)v(\theta)ds}
    \end{equation}
    The logarithm of the integral can be seen as the model complexity:
    \begin{equation}
        COMP(M_\gamma;v)=log\int_{x\in\mathcal{X}}max_{\theta \in \Theta_\gamma}p(s|\theta)v(\theta)ds
    \end{equation}
    Jalaldoust picked the optimized model $\hat{\gamma}^{MDL}\in\Gamma$ using 
    \begin{equation} \label{eq13}
        \hat{\gamma}^{MDL}=\mathop{\arg\min}_{\gamma\in\Gamma}L_v(\gamma;x)
        =\mathop{\arg\min}_{\gamma\in\Gamma}(-log\pi(\gamma)-r_v(\hat{\theta}_{v|\gamma}(x);x)+COMP(M_\gamma;v))
    \end{equation}
    where $\pi$ is a uniform distribution and $r_v(\hat{\theta}_{v|\gamma}(x);x)$ is the goodness-of-fit relevant to $p$, $v$, and $\Theta_\gamma$.

    Moreover, consider a one-to-one mapping from $\gamma \in \Gamma$ to a $p\times p$ adjacent matrix of a causal graph within the set of all binary $p\times p$ matrices. By optimizing the~\ref{eq13}, one can choose the most appropriate model from their predefined model family, hence, infer the Granger causal relationships between event types.\\
    
    \subsubsection{Methods for Wold processes}
    While most of the existing algorithms concerning discovering Granger causality from event sequences are based on Hawkes Processes, we can also model these relationships on another type of process - Wold Processes, which bear less complexity in nature. Suppose we denote $\delta_i=t_i-t_{i-1}$ to be the waiting time for $i$-th event from the occurrence of $(i-1)$-th event. Wold Processes are built upon a simple assumption that the current waiting time $\delta_i$ is only related to the closest past waiting time $\delta_{i-1}$. That is to say, the set $\{\delta_i, i\in \mathrm{N}\}$ forms a Markov chain. The inherent Markov property within the Wold processes makes them suitable for modeling the dynamics of certain complex systems. Besides, Figueiredo et al.~\cite{NEURIPS2018_aff0a6a4}) have measured the correlation between $\delta_i$ and $\delta_{i-1}$ on certain datasets. The result shows that in most of their datasets, the median Pearson correlation is above 0.7, which is a sign of the adequacy of the Wold model. Accordingly, following Alve et al.~\cite{10.1145/2939672.2939852} and Figueiredo et al.~\cite{NEURIPS2018_aff0a6a4}'s idea, the intensity function can be performed as
    \begin{equation}
        \lambda_{e_i}(t)=\mu_{e_i}+\sum_{e_j\in E}\frac{\alpha_{e_ie_j}}{\beta_{e_j}+\Delta_{e_ie_j}(t)}  \nonumber
    \end{equation}
    based on the BuSca model. Here, $\mu_{e_i}$ is the base intensity as in Hawkes Processes. $\Delta_{e_ie_j}(t)$ denote the time interval between the last $e_i$ type occurrence and $e_j$ type occurrence on time t. That is, if we define the closest $e_i$ type event before time $t$ happened at time $t_{e_i}$, and $e_j$ type before time $t_{e_i}$ happened at time $t_{e_j}$, there is $\Delta_{e_ie_j}(t)=t_{e_i}-t_{e_j}$. Hence, the cross-type entry $\sum_{e_j\in E}\frac{\alpha_{e_ie_j}}{\beta_{e_j}+\Delta_{e_ie_j}(t)}$ in our intensity function will be larger if $\Delta_{e_ie_j}(t)$ decrease. This perfectly matches the fact that if type-$e_i$ events always happen just before type-$e_j$ events occur, we see a greater probability that $e_j$ has a certain effect on $e_i$. $\alpha_{e_ie_j}$ is the normalizing entry satisfying $\sum_{e_j\in E}\alpha_{e_ie_j}=1$, while $\beta_{e_j}$ is the base rate such that when the time interval $\Delta_{e_ie_j}(t)$ between two types are infinitesimal at time $t$, the cross-type entry will converge to $\frac{\alpha_{e_ie_j}}{\beta_{e_j}}$.
    
    The Granger causality can be learned through this Wold-based model by inspecting $\alpha_{e_ie_j}$. Specifically, if $\alpha_{e_ie_j}\neq0$, it is considered that $e_j$ Granger-cause $e_i$. Since the approaches to learning the processes may not have an adequately sparse solution, Figueiredo tested the statistical significance of these possible Granger causal relationships and discard those with low significance. Moreover, the Wold-based model can be learned through MCMC, Expectation Maximization (EM)~\cite{NEURIPS2018_aff0a6a4} and Variational Inference~\cite{pmlr-v130-etesami21a} approaches. The task is to infer the parameters $\{\alpha_{e_ie_j}, \beta_{e_j}, \mu_{e_i} |\forall e_i, e_j\}$ in the intensity functions, which could reveal all the properties inside event sequences. Here we do not elaborate on the details of these learning methods.\\

    \subsubsection{Methods for Neural Point Processes}
    
    With the rapid development of neural networks, \textit{Neural Point Processes(NPPs)} have gradually been utilized to model event sequences and infer causal relations.
    The core idea of these NPP algorithms is to use neural networks to infer the intensity function $\lambda_e(t)$. In specific, they encode an event sequence into the hidden state, during which they capture the feature of the sequence. Then, they use decoders to infer the future intensity function. And there are two major types of NPPs. One is based on the autoregressive(AR) model; its hidden states $h_i$ only update when an event occurs. The other follows the hypothesis that the hidden state $h(t)$ changes continuously in time. The continuous-time models hold advantages that they are natural and more suitable for estimating attributes at any time $t$ because of their continuous traits. Nonetheless, this flexibility comes at a cost. Continuous hidden models could suffer from a slower training speed compared with AR-based discrete models. This is because the evolution, likelihood, and sampling processes might demand numerical approximations.
    In this part, we will first give the basics of NPPs and then introduce how to learn the Granger causality in NPPs.

    \textbf{Basics of NPPs.} The general process of using the AR-based NPP model to infer Granger causality~\cite{pmlr-v119-zhang20v} is presented as follows. First, we embed each event into a vector $v_i=[\theta(t_i-t_{i_1});V^T z_i$], where $\theta(\cdot)$ is a predefined function, V is the embedding function for events' type, $[\cdot;\cdot]$ could be concatenation and $z_i$ could be the one-hot coding for event type $u_i$. Then, we utilize a sequence encoder(e.g. LSTM or GRU) to encode $\{v_j; j\leq i\}$ to $h_i=Enc(h_{i-1}, v_i)$. Also, there exists a different encoding method in which the encoding is done independently for each i using, e.g., self-attention. This can better capture the long-range dependencies between events but have heavy computing complexity as well.
    
    Next, we aim to decode the hidden state $h_i$ into the intensity function $\lambda_e(t)$. To do this, we need to make some assumptions about $\lambda$. For example, we could predict that the intensity function can be divided into the sum of some interaction-related and time-related functions $\lambda_e(t)=\sum_{m=1}^{M}a_{em} \kappa_m(t)$ similar to what Xu did~\cite{pmlr-v48-xuc16}. Sequentially, we could only infer the $a_{em}$ entries since $\kappa_m(s)$ can be chosen from a large function family in which the functions can represent a wide variety of time-varied patterns. Hence,
    \begin{equation}
        \mathbf{\alpha}:\mathbb{R}^{rank(h_i)}\rightarrow \mathbb{R}_+^{K*S} \nonumber
    \end{equation}
    is the corresponding decoder for this model. Here, $k$ and $S$ are the numbers of event types and basic functions correspondingly. However, the aforementioned method could not fit the continuous hidden state model. Under this circumstance, since $h(t)$ is continuous, thus carrying a lot more time-varying information than $h_i$ do, we can simply define the intensity as:
    \begin{equation}
        \lambda_e(t)=g_e(h(t))
    \end{equation}
    Here, $g_e:\mathbb{R}^{rank(h_i)}\rightarrow R_{>0}$ is a non-linear function (e.g. softplus function) which maps $h(t)$ to the corresponding intensity function for event-type $e$ at time $t$.
    
    As for the training process, most of the NPPs nowadays use the Maximum Likelihood Estimation(MLE) method as most of the traditional PP methods do. They take the negative log-likelihood of MLE as the objective function and use the neural network to optimize it. Besides, there exist alternative methods to use for learning the MLE. For example, if we set the objective as $\mathbb{E}_{X\sim p(X)}[f(X)]$, we can model the point process $P(X)$ using $f(X)$ by variational inference or reinforcement learning.

    \textbf{Inferring Granger Causal Relations from NPPs.} When it comes to inferring the Granger causality, attribution methods need to be used~\cite{pmlr-v119-zhang20v}. That is because, in neural methods, most of the algorithms do not follow the parametrization in the Hawkes processes. On the contrary, their goal is to directly model the processes' intensity function in order to loosen the strictness of Hawkes process and thus gain more accuracy. Since those intensity functions captured all the characteristics of event sequences, we should take full advantage of them. To do that, Zhang et al. first denoted $x_p=[t_1,e_1,...,t_p,e_p,t_{p+1}]$, $\underline{x_p}=[t_1,0,...,t_p,0,t_{p+1}]$ as the baseline input, and $f_k(x_p)=\int_{t_p}^{t_{p+1}}\lambda_e(s)ds$ as the impact function~\cite{pmlr-v119-zhang20v}. For each event type k, we have:
    \begin{equation}
        f_k(x_p)-f_k(\underline{x_p})=\sum_{q=1}^{p}A_q(f_k,x_p,\underline{x_p}) \nonumber
    \end{equation}
    where $A_q(f_k,x_p,\underline{x_p})$ is the attribution(e.g. Integrated Gradients) for the event type of $z_q$. Hence, $A_q(f_k,x_p,\underline{x_p})$ can be regarded as the contribution of $z_j$-type events to the prediction of k-type events given the history $x_p$. Next, Zhang conducted a normalization on $A_q(f_k,x_p,\underline{x_p})$ as
    \begin{equation}
        Y_{e_i,e_j}=\frac{\sum_{s=1}^{s}\sum_{p=1}^{n_s}\sum_{q=1}^{i}\mathbb{I}(k_q^s=e_j)A_q(f_{e_i},x_p^s,\underline{x_p^s})}{\sum_{s=1}^{s}\sum_{p=1}^{n_s}\mathbb{I}(k_q^s=e_j)} \nonumber
    \end{equation}
    Consecutively, the Granger causality between $u_i$ and $u_j$-type events can be inferred from $Y_{e_ie_j}$. This method can measure not only the inhibitive causality but also the magnitude of the causality.
    
    Interestingly, some other neural algorithms just model the intensity function as in Hawkes processes. They set $\mu$ and $\alpha$ as matrices and directly put $H$ and $A$ into the neural networks. Since the input structure is much easier, we could add other hypotheses, like the topological structure between events, and let the neural network(in this case, GCN) optimize $H$ and $A$ iteratively. Then, we could directly infer Granger causality from the matrix $A$.\\

\subsection{Other Inferring Approaches}
\label{subsec43}
    In this section, we will not directly model intensity functions in point processes. Instead, we focus on discovering the relationships between different processes (i.e., different types of events). To do that, we could utilize the Graphical Event Model mentioned before and loosen the assumption that each node follows the Hawkes Processes. Historically, Didelez et al. and Meek et al. first introduced the Graph Event Models to capture dependencies among events. Based on common graph methods, they assumed that an event type's intensity function is only related to its parental type. GEMs capture dependencies between various types of events over time, providing a general framework to model the dependency in graph methods. Therefore, similar to the stationary as well as the discrete-time case, constraint-based and score-based approaches can be utilized.\\
    
    \subsubsection{Constraint-based methods}
    Just like the notion \textit{independence} between different random variables, we can define \textit{process independence} for point processes:
    \begin{Definition} (Didelez, 2008~\cite{https://doi.org/10.1111/j.1467-9868.2007.00634.x}; Bhattacharjya et al., 2022~\cite{Bhattacharjya2022ProcessIT})\\
        For processes $X, Y, Z$, s.t. $Y\cap Z=0$, $X$ is a process independent of $Y$ given $Z$ if all events in $X$ have conditional intensities such that if historical information of events in $Z$ is known, then those events in process $Y$ do not provide any further information.
    \end{Definition}
    Meek et al.~\cite{article} and Bhattacharjya et al.~\cite{Bhattacharjya2022ProcessIT} introduced the notion of $\delta^*$-separation, which is based on $d$-separation but released its restriction of not having self-loops and made each self-loop independent of their own history. Then, they proposed a causal dependence assumption with $\delta^*$-separation analogous to the faithfulness assumption. Based on the causal dependence assumption, several constraint-based methods, such as the PC and max-min parents algorithms, are proposed to learn the causal relationships between different types. There are several Process Independence testers to choose from. For example, we have the NI tester:
    \begin{equation}
        \text{NI score} = \frac{1}{2}\frac{\sum_{z}(\lambda_{x|y,z}-\lambda_{x|\hat{y},z})^2}{\sum_{z}\lambda_{x|y,z}+\lambda_{x|\hat{y},z})^2}
    \end{equation}
    where $y$ and $\hat{y}$ indicate the parental state where $Y$ has or has not appeared in its window. We also have the LR tester:
    \begin{equation}
        \text{LR score} = F_{\chi_{2^{|Z|}}^2}(-2[logL^*(X|Y,Z)-logL^*(X|Z)])
    \end{equation}
    Here, $F(\cdot)$ is the cumulative distribution function of a chi-squared random variable with $2^{|Z|}$ degrees of freedom. Then, we apply a threshold $\tau$ for each tester, that is, when the score is less than $\tau$, there is no causal relationship between type $X$ and type $Y$.\\
    
    \subsubsection{Score-based methods}
    Similarly, there are score-based methods that can be applied to GEMs. Bhattacharjya et al.~\cite{NEURIPS2018_f1ababf1} proposed PGEM - a model that assumed its intensity functions are only influenced by whether or not parent types happened in some recent time window. In addition, they used the BIC criterion on conditional intensities $\lambda_{x|u}$ to search for the optimal parent sets for each event type, that is, to infer the graph structure in their PGEM model. The graph structure is a representation of the causal relationships between different types of events.
    
    \subsubsection{Transfer Entropy}
    Recall that Transfer Entropy (TE) can be used to discover causal relationships in discrete-time cases. Here, we can also apply TE to event sequences (i.e., point processes) to identify our continuous-time causal relationships. Specifically, Spinney et al.~\cite{PhysRevE.95.032319} constructed a continuous-time pairwise Transfer Entropy:
    \begin{equation}
        \mathbf{T}_{Y\rightarrow X}=\lim\limits_{\tau\to\infty}\frac{1}{\tau}\sum_{i=1}^{N_X}ln\frac{\lambda_{x|\mathbf{x}<t,\mathbf{y}<t}[\mathbf{x}_{<x_i},\mathbf{y}_{<y_i}]}{\lambda_{x|\mathbf{x}<t}[\mathbf{x}_{<x_i}]}
    \end{equation}
    where $N_X$ is the number of events in the target process and $\tau$ is the length of time when there holds the corresponding intensity function $\lambda_{x|\mathbf{x}<t,\mathbf{y}<t}[\mathbf{x}_{<x_i},\mathbf{y}_{<y_i}]$ and $\lambda_{x|\mathbf{x}<t}[\mathbf{x}_{<x_i}]$. The processes are independent when $\mathbf{T}_{Y\rightarrow X}=0$. We can define the conditional TE similarly. There are some existing consistent methods for estimating the continuous-time TE and its conditional form~\cite{10.1371/journal.pcbi.1008054}.\\

\section{Applications}\label{sec:app}

Temporal causal discovery has been widely used in many areas, such as scientific endeavors (earth science~\cite{Resources/datasets_surveys/runge2019inferring}, neuroscience~\cite{Applications/neuroscience/Nature_neuroscience_reid2019advancing, Applications/neuroscience/recent_good_summary/jocn_WeichwaldP21, Applications/neuroscience/nature_review/siddiqi2022causal}, bioinformatics~\cite{Applications/Gene_Science_sachs2005causal}), industrial implementations (anomaly detection~\cite{Applications/anomaly/work1_icdm_QiuLSL12}, root cause analysis~\cite{industiral_app_review/vukovic2022causal, Applications/CPSs/kbs_LiuLJS21, Applications/RCA/Assaad23EasyRCA}, business intelligence in online systems~\cite{Applications/interest_social_nx/cikm_ArabzadehFZNB18}, video analysis~\cite{Applications/video/iclr_YiGLK0TT20}). 
Table~\ref{tab:application_overview} summarizes the application areas and corresponding studies.
For scientific research, the learned causal relations should not usually be considered end results but rather starting points and hypotheses for further studies~\cite{Discussion/knowledge/makela2022incorporating}.
As a facilitator, causal discovery can play a supporting role in a multi-stage approach in an industrial setting~\cite{industiral_app_review/vukovic2022causal}.
In the rest part of this section, we will review three areas including earth science, anomaly detection and root cause, to explain these main workflows of incorporating temporal causal discovery into both scientific endeavors and industrial implementations, respectively.

\begin{table*}[t]
    \centering
    \caption{Major studies in temporal causal discovery applications.}
    \label{tab:application_overview}
    \tiny
    \scalebox{0.95}{
    \begin{tabular}{llp{8cm}}
    \toprule
    Groups & Application areas & Studies \\
    \midrule
    \multirow{10}{*}{Scientific endeavors}   
    & Earth science & Climate change detection and attribution (\eg,~\cite{Applications/climate_attr/kdd_LozanoLNLPHA09}); Quantifying climate interactions (\eg,~\cite{Applications/earthS/climate_runge2014quantifying}); Latent driving force detection (\eg,~\cite{Applications/anomaly/work3_dagm_TrifunovSRERD19, EarthSci/abnorm/causal_intens/DBLP:conf/dagm/ShadaydehDGM19}); Causality validation between temperature and greenhouse gases (\eg,~\cite{Applications/earthS/climate_change_van2015causal}). \\
    & Neuroscience &
    Dynamic causal models for neural connectivity (\eg,~\cite{Applications/neuroscience/work1_neuroimage_PennySMF04, Applications/neuroscience/work1_ploscb_PennySDRFSL10, Applications/neuroscience/work1_neuroimage_JafarianLCFZ20}); Granger causal models for neural connectivity (\eg,~\cite{Applications/neuroscience/work2_bc_KaminskiDTB01, Applications/neuroscience/PNAS/stokes2017study, Applications/neuroscience/PNAS/sheikhattar2018extracting, 10.1371/journal.pcbi.1001110}); Causal inference from noninvasive brain stimulation (\eg,~\cite{Applications/neuroscience/jocn/BergmannH21}). \\

    & Bioinformatics & Modeling gene regulatory network (\eg,~\cite{MTS/SB/MMHO_DBN/origin/cibcb/LiN13, MTS/SB/MMHO_DBN/tcbb/LiCZN16, DBLP:journals/ploscb/VernySASI17, Applications/Gene_patil2022learning, DBLP:conf/iclr/Wu0B22}). \\
    \midrule
    \multirow{22}{*}{Industrial implementations}   
    & Anomaly detection & Causal structure as detection reference (\eg,~\cite{Applications/anomaly/work1_icdm_QiuLSL12, Applications/anomaly/work2_icdm_BehzadiHP17, Applications/anomaly/work4_ijdsa_ApteVP21, Applications/anomaly/work5_22}); Detection from imbalanced data (\eg,~\cite{Application/SCGL_use_cikm_HuangXYYWX20}).   \\
    & Root cause analysis & Oscillation propagation tracing in the control loop (\eg,~\cite{landman2014fault, landman2016hybrid, chen2017root, lindner2018diagnosis}); Alarm flood reduction (\eg,~\cite{wang2015data, rodrigo2016causal, wunderlich2017structure}); Industrial knowledge combined analysis (\eg,~\cite{landman2016hybrid, industrial_knowledge/cao2022causal, thambirajah2009cause, 9964900}).  \\
    & Business intelligence in online systems & User interest prediction (\eg,~\cite{Applications/interest_social_nx/cikm_ArabzadehFZNB18, Applications/interest_social_nx/asunam_HauffaBG19}); Social media analysis (\eg,~\cite{Applications/twitter/wsdm/ChangWML13, Applications/social_media_financial_data/tsapeli2017non, Applications/socialnx/complexnetworks/KuzmaCC21, Chuzhe_app/nca/ChenCHYX20}; Online advertising (\eg,~\cite{Applications/marketing/www_NuaraSTZ0R19, Applications/marketing/kdd_CausalMTA_YaoGZCB22, MTS/Attention/icdm_InGRA_ChuWMJZY20}); User-item interaction in recommendation (\eg,~\cite{Chuzhe_app/kais/ShangS20}); User activity modeling (\eg,~\cite{Chuzhe_app/cikm/LiGBC17, Chuzhe_app/yao2022high}). \\

    & Video analysis & Video analysis and reasoning (\eg,~\cite{Applications/video/iclr_YiGLK0TT20, Applications/video/nips_Li0AFG20}); Interpretable Gait Recognition (\eg,~\cite{Applications/video/icpr/BalaziaHSP22}). \\

    & Urban data analysis & Trajectory pattern mining (\eg,~\cite{Applications/traj/jcss/ChuWCFH16, Chuzhe_app/yang2021individual}); Traffic flow prediction (\eg,~\cite{Applications/traffic_flow_pre/li2015robust}); Visual urban and causal analytics (\eg,~\cite{Chuzhe_app/tvcg/DengWXBZXCW22}). \\
    & Clinical data analysis & Causal chain discovery (\eg,~\cite{Chuzhe_app/wei2022granger}); Hypothesis testing (\eg,~\cite{Chuzhe_app/pandey2021multimodal}); Stable causal structure learning (\eg,~\cite{Applications/clinic_data/rahmadi2019finding}). \\
    & Signal processing & Blind source separation (\eg,~\cite{Chuzhe_app/crowncom/TestiFG20, Chuzhe_app/tcom/TestiG21}); Compressed sensing (\eg,~\cite{Chuzhe_app/abs-2210-11420}).   \\ 
    & Financial analysis & Causal discovery for financial news (\eg,~\cite{Chuzhe_app/tetereva2018financial, Chuzhe_app/rambaldi2015modeling}). \\
    & Military & Battlefield sequential events analysis (\eg,~\cite{Chuzhe_app/li2022discover}).   \\ 
    & Robotics and dynamic control systems & Identifying causal structure (\eg,~\cite{Applications/dynamic_control_systems/TMLR_corr_abs-2006-03906}); Causal generalization (\eg,~\cite{Applications/dynamic_control_systems/agi_SheikhlarET21}). \\

    \bottomrule
    \end{tabular}}
    \vspace{-4ex}

\end{table*}

\textbf{Earth science and climate change research:} Temporal causal discovery approaches have been widely used in the community of earth science and climate change research \cite{Applications/climate_attr/kdd_LozanoLNLPHA09, Applications/earthS/climate_ebert2012causal, Applications/earthS/climate_runge2014quantifying, Applications/earthS/climate_change_van2015causal, Applications/earthS/climate_attr_hannart2016causal, Resources/datasets_surveys/runge2019inferring, Applications/anomaly/work3_dagm_TrifunovSRERD19}.
Climate is a complex and chaotic system, incorporating spatio-temporal information.
Traditional climate models based on forward simulations have inherent limitation in describing such system due to uncertainties, simplifications, and discrepancies from observed data \cite{Applications/climate_attr/kdd_LozanoLNLPHA09}. 
Whereas, commonly used data centric methods such as lagged cross-correlation and regression analysis, aiming at deriving insights into interaction mechanisms between climate process, may lead to ambiguous conclusions in the field \cite{Applications/earthS/climate_runge2014quantifying}.  
To overcome the aforementioned issues, it's reasonable to meaningfully characterize causal relationships among parameters of interest and make assertions.
Specifically, spatio-temporal Granger modeling via group elastic net is proposed in \cite{Applications/climate_attr/kdd_LozanoLNLPHA09} to conduct climate change detection and attribution, where the extreme-value theory to model and attribute extreme events in climate, such as severe heatwaves and floods. 
In \cite{Applications/earthS/climate_runge2014quantifying}, a graphical Granger model followed by a causal interaction strength measure is proposed to quantify the strength and delay of climate interactions and overcome the possible artifacts from vanilla correlation or regression methods.  
Another challenge is the existence of unobserved confounders, which may either lead to incorrect attribution or perform as a nonnegligible driving factor.  
A line of work \cite{Applications/anomaly/work3_dagm_TrifunovSRERD19, EarthSci/abnorm/causal_intens/DBLP:conf/dagm/ShadaydehDGM19} detect the latent driving force of abnormal event in climate by estimating the causal link intensity between confounded variables.  
Besides, in climate system some parameters of interest show strong coupling, thus impose difficulties for identification of causal orientation. The convergent cross mapping (CCM) technique, which is designed for strong coupling dynamic systems, is used in \cite{Applications/earthS/climate_change_van2015causal} to identify the causality between temperature and greenhouse gases, between which the statistical association is well documented while the causality is different to extract from the observed data. 
A recent overview of time series causal discovery in the earth system is also provided in \cite{Resources/datasets_surveys/runge2019inferring}, where avenues for future work in both method developments and scientific endeavors are depicted.

\textbf{Industrial temporal anomaly detection:} In industrial systems, detecting anomalies in massive temporal data, which is derived from sensors, logs, physical measurements, system settings, etc, is meaningful while challenging.
The anomalies can be roughly categorized into univariate anomaly, which has been extensively studied, and dependency anomaly, which is much more challenging to detect but common in real-world applications.
As the challenges mainly come from high dimensions and complex dependency in data, methods \cite{Applications/anomaly/work1_icdm_QiuLSL12, Applications/anomaly/work2_icdm_BehzadiHP17, Applications/anomaly/work4_ijdsa_ApteVP21, Applications/anomaly/work5_22, Application/SCGL_use_cikm_HuangXYYWX20} based on temporal causal discovery have played a nonnegligible role in the dependency anomaly detection by providing efficient, robust and interpretable results.
Causal discovery can facilitate the detection of the generative mechanisms of an underlying system.
The key idea of this family of work is first to construct causal graphs from multivariate time series, and then detect anomalies according to the extracted causal relations. 
To be specific, in \cite{Applications/anomaly/work1_icdm_QiuLSL12, Applications/anomaly/work2_icdm_BehzadiHP17}, Granger graphical models are built on a reference data set and a testing data set respectively, the distribution differences (such as KL-divergence and Jensen-Shannon divergence) between the two learned models are computed as anomalous measures.  
In \cite{Applications/anomaly/work4_ijdsa_ApteVP21}, the inferred relation based on Granger causality is termed causally anomalous if it violates the domain knowledge or the frequently observed forms.
Recently, a causal perspective is also taken in \cite{Applications/anomaly/work5_22} to detect multivariate time series anomalies and leveraged in AIOps applications. In this work, the computation cost is reduced because instead of modeling joint distribution directly, it models factorized distribution modules from learned causal structures, where each corresponds to a local causal mechanism.    
Besides, as for the imbalanced flight data where the anomalous data points are rare, a time series classification method is proposed in \cite{Application/SCGL_use_cikm_HuangXYYWX20} based on nonlinear Granger causality learning.

\textbf{Root cause analysis in manufacturing process:} The root cause analysis is a vital task to ensure process safety and productivity in the industrial context, where the manufacturing processes are temporal and complex scenarios usually composed of multiple process units and a large number of feedback control loops. However, the acceptance of powerful ML methods in this field is hindered due to increasing requirements of fairness, accountability, and transparency (a.k.a., FAT principle \cite{FAT_principle/DBLP:journals/chb/ShinP19}), especially in sensitive-use cases \cite{industiral_app_review/vukovic2022causal}. To alleviate this issue, extracting knowledge such as causal relationships is paramount in this field. The last decade has witnessed the proliferation of the causal discovery methods for root cause analysis \cite{landman2014fault, wang2015data, rashidi2018data, Applications/CPSs/kbs_LiuLJS21, industiral_app_review/vukovic2022causal}. For instance, temporal causal discovery approaches such as Granger causality, transfer entropy, and their variants are leveraged to trace the oscillation propagation in the control loop \cite{landman2014fault, landman2016hybrid, chen2017root, lindner2018diagnosis}. The reduction of alarm flood, which has been recognized as a major cause of industrial incidents, is another aspect of industrial root cause analysis. Among three typical nuisance alarms (i.e., repetitive alarms, standing alarms and consequence alarms \cite{ henningsen1995intelligent}), it's challenging to suppress the consequence alarms and to provide a proper on the condition that the abnormality occurs and propagates. To identify all causal relations between alarms is of help \cite{hollender2007intelligent}, and a line of work \cite{wang2015data, rodrigo2016causal, wunderlich2017structure} leverages causal discovery approaches in this task.
Besides, profound industrial knowledge, such as information flow and energy flow, can be combined with causal discovery to eliminate spurious relations \cite{landman2016hybrid, industrial_knowledge/cao2022causal, thambirajah2009cause}.

\section{Performance Evaluation}\label{sec:res}

In this section, we give an overview of the benchmark datasets and evaluation metrics used in temporal causal discovery.

\subsection{Datasets}

We will briefly introduce some of the datasets used in temporal causal discovery, including MTS datasets and event-sequence datasets.

Datasets for MTS causal discovery range from health data to financial data. We discuss some of the commonly used datasets, which are publicly available and with the ground truth of causal graphs.
\begin{itemize}
    \item \textbf{Lorenz-96 simulated data}: It's a nonlinear model formulated in~\cite{lorenz1996predictability} to simulate climate dynamics. The continuous dynamics in a $d$-dimensional Lorenz model are given by $    \frac{\partial \mathbf{x}_i^t}{\partial t} = -\mathbf{x}_{i-1}^t (\mathbf{x}_{i-2}^t - \mathbf{x}_{i+1}^t  ) - \mathbf{x}_i^t + F, \ i \leq i \leq d$. The system dynamics become increasingly chaotic for higher values of forcing constant $F$. As a standard benchmark, it's used by~\cite{MTS/Granger/pamiNGC22, MTS/Granger/iclr20_esru, MTS/Granger/iclr21_GVAR_MarcinkevicsV, MTS/Attention/icdm_InGRA_ChuWMJZY20, MTS/Granger/CR_VAE2023}.

 
    \item \textbf{Linear VAR simulated data}: Time series measurements are generated according to the linear VAR model. In~\cite{MTS/Granger/pamiNGC22, MTS/Granger/iclr20_esru}, it's used to analyze methods' performance when the true underlying dynamics are linear.

    \item \textbf{CMU Human motion capture (CMU MoCap) data}: It's a data set from CMU MoCap database\footnote{\url{http://mocap.cs.cmu.edu/}}, containing data about joint angles, body position.
    Causal discovery methods can be leveraged to extract nonlinear dependencies between different regions of the body~\cite{MTS/Granger/pamiNGC22}.

    \item \textbf{DREAM-3 in Silico Network Inference Challenge}: In DREAM-3 IN Silico Network Challenge~\cite{prill2010towards}, time-series data is simulated using continuous gene expression and regulation dynamics.
    Five gene regulation networks are to be inferred from gene expression level trajectories recorded.
    This dataset has been used to evaluate causal discovery algorithms in~\cite{MTS/Granger/pamiNGC22, MTS/Granger/iclr20_esru}.

    \item \textbf{Blood-oxygenation-level dependent (BOLD) imaging data}: In this dataset\footnote{\url{https://www.fmrib.ox.ac.uk/datasets/netsim/index.html}}~\cite{DBLP:journals/neuroimage/SmithMKWBNRW11}, time-ordered samples of the BOLD signals measure different brain regions of interest in human subjects. It's generated using the dynamic causal modeling functional magnetic resonance imaging (fMRI) forward model. In~\cite{MTS/Granger/iclr20_esru, MTS/Attention/TCDF_NautaBS19}, causal discovery methods are applied to estimate the connections in the human brain based on BOLD imaging data.

    \item \textbf{Simulated financial time series}: The dataset\footnote{\url{http://www.skleinberg.org/data.html}}~\cite{MTS/KleinbergS/kleinberg2013causality} is created using factor model to describe portfolio's return depending on three factors and a portfolio-specific error term. Thus the true relationships are known. It's used by~\cite{MTS/Attention/TCDF_NautaBS19}.


\end{itemize}

As for event sequences, datasets range from online behavior to electricity.
However, the true information on causal relationships is not accessible under all scenarios.

\begin{itemize}

    \item \textbf{MemeTracker}: 
    It is a dataset\footnote{\url{http://memetracker.org}} that captures online articles' website, publication time, and all the hyperlinks within. This data set originally represents how a meme flow on different websites. The domain of the website and the publication time are considered an event type and its occurring time. And the hyperlinks between different websites can be seen as the ground truth of causal relationships. It's used by~\cite{pmlr-v70-achab17a, NEURIPS2018_aff0a6a4, pmlr-v119-zhang20v}.

    \item \textbf{IPTV viewing records}: 
    This dataset\cite{6717182} records the user's viewing behavior, i.e., what program and when they watch in the IPTV systems. The type of program and the time of watching the program can be deemed as an event type and its occurring time, respectively. It's used by~\cite{pmlr-v48-xuc16, CHEN202222, pmlr-v119-zhang20v}. However, ground-truth causal relationships are not included in this dataset.

    \item \textbf{Power grid failure event data}: 
    This dataset includes abrupt changes in the voltage or current signals within Phasor Measurement Units (PMUs) as well as each PMU's ID. The mission of the causal diagnosis task with this dataset is to infer the causalities within the grid~\cite{NEURIPS2021_15cf7646}. Since the network topology is not given out of privacy concerns, this is a non-ground-truth task.

    \item \textbf{G-7 bonds}: 
    This dataset\cite{https://doi.org/10.1002/jae.2585} includes the daily return volatility of sovereign bonds of countries in the Group of Seven. The goal of dealing with this dataset is to discover the causal network underneath sovereign bonds~\cite{Schindler_Plant_2022}. Expert knowledge from the domain can be deemed as ground truth.
\end{itemize}

\subsection{Evaluation Metrics}

In this part, we will explain different metrics used in the literature.
Given the inferred probability of an edge $p(A_{ij})$ thresholded by $thre \in (0,1)$, the set of ground truth edges in causal graph $E_{GT} = \{(i,j): A_{ij}^* = 1\} $, and the set of ground truth missing edges in causal graph $E_{MS} = \{(i,j): A_{ij}^* = 0\} $, the definition and description of commonly used metrics is provided as follows:
\begin{itemize}
    \item \textbf{True Positive Rate (TPR)}: As a ratio of common edges found in the causal discovery results and the ground truth adjacencies over the total number of ground truth edges, the TPR metric is defined as $TPR = \frac{ | \{ (i,j): p(A_{ij}) \geq thre  \}   \cap     E_{GT} |     }{ | E_{GT} |  } $.
    \item \textbf{False Positive Rate (FPR)}: Similar to that in TPR, FPR refers to the ratio of common edges found in the causal discovery results and the ground truth missing adjacencies over the number of ground truth missing edges, which is defined as $FPR = \frac{ | \{ (i,j): p(A_{ij}) \geq thre  \}   \cap     E_{MS} |     }{ | E_{MS} |  }   $.
    \item \textbf{Area Under the Receiver Operator Curve (AUROC)}: The Receiver Operator Curve (ROC) is defined as the ratio of TPR and FPR given the threshold $thre$ varies between $0$ and $1$. The area under the ROC (AUROC) is then widely used to assess the performance of causal discovery algorithms.
    \item \textbf{Structural Hamming Distance (SHD)}: SHD is a metric describing the number of edge edition that need to be made to turn the discovered graph to its ground truth counterpart, which sums the number of missing edges, extra edges, and incorrect edges.   
\end{itemize}


\section{Discussion and New Perspectives}\label{sec:discuss}

In this section, we first discuss challenges and practical considerations, including non-stationarity, heterogeneity, unobserved confounders, subsampling, and expert knowledge.
Then, two new perspectives of temporal causal discovery are provided, which in our opinion will be a promising avenue for future research.

\subsection{Challenges and Practical Considerations}
\label{subsec:challenges}

\textbf{Non-stationarity of data:} We are often faced with non-stationarity in practical scenarios, where the probability distributions of temporal variables conditional on their causes or even the causal relations may change across time, especially for temporal data.
In this condition, causal discovery approaches presuming a fixed causal model may give misleading results. 
Whereas, several types of research have shown that non-stationarity contains information for causal discovery \cite{CD_from_change/conf/uai/TianP01, CD_from_change/peters2016causal, Discussion/Nonstation_hetero/ijcai_ZhangHZGS17, Discussion/Nonstation/state_space_icml_Huang0GG19}.
Thus, it's important to properly tackle the non-stationarity in applications.
Non-stationarity may result from the change of underlying systems and can be seen as a soft intervention \cite{soft_interv/korb2004varieties} done by nature. 
Following this idea, a line of work \cite{Discussion/Nonstation_hetero/ijcai_ZhangHZGS17, Discussion/Nonstation_hetero2/jmlr/Huang0ZRSGS20} leverages a surrogate such as time and domain index to account for nonstationarity where the causal relations are changed, and the CD-NOD framework is proposed. 
Instead of leveraging informative non-stationarity to causal structure learning, another set of research focuses on modeling time-varying relationships \cite{Discussion/Nonstation/pr_GaoY22}. 
Besides, the approach for slowly varying non-stationary process, such as evolutionary spectral and locally stationary processes, is proposed in \cite{Discussion/Nonstation/slowly_varying/du2020causal}.

\textbf{Heterogeneity of data:} In causal discovery for practical applications, the heterogeneity of data lies in two levels: (1) The interacting temporal processes are heterogeneous (having different distributions), for instance, causally related meteorological observations from different stations are influenced by several major weather systems separately \cite{Discussion/heterogeneous/pakdd_BehzadiHP19}. (2) The underlying generating process changes across data sets or different domains \cite{intro/nonts_surveys/glymour2019review}, for instance stock prices from different markets \cite{Discussion/Nonstation_hetero2/jmlr/Huang0ZRSGS20} or individual behaviours in different paradigms \cite{MTS/Attention/icdm_InGRA_ChuWMJZY20}.
For the first condition where the heterogeneity exists among temporal variables, the inferred relations of the traditional causal discovery approaches, which have been designed for specific homogeneous data types, may be inaccurate. As a remedy, several variants of Granger causality, based on methods such as generalized linear models and minimum message length, are proposed in \cite{Applications/anomaly/work2_icdm_BehzadiHP17, Discussion/heterogeneous/pakdd_BehzadiHP19, Discussion/heterogeneous/entropy/Hlavackova-Schindler20}.
For the second condition, a line of work \cite{Discussion/Nonstation_hetero/ijcai_ZhangHZGS17,  Discussion/Nonstation_hetero2/jmlr/Huang0ZRSGS20} leverages the distribution shift from heterogeneity as a soft intervention to assist causal structure learning, which is similar to that in non-stationary data.  
Whereas, another line of causal discovery approaches \cite{MTS/Attention/icdm_InGRA_ChuWMJZY20, Discussion/NewForm/ACD_LoweMSW22} in the second condition focuses on inductively modeling typical structure in heterogeneous data within an end-to-end framework.

\textbf{Unobserved confounders:}
In practice, we are often met with cases where causal sufficiency is violated, \ie, there exist unobserved confounders. 
This challenging setting may lead to incorrect causal relations~\cite{MTS/FCM/VAR_LINGAM_extend2_icml_GeigerZSGJ15}.
As summarized in Table~\ref{tab:ts_category_overview}, most temporal causal discovery approaches cannot handle unobserved confounders in a straightforward way.
Several constraint-based approaches are designed without causal sufficiency and approaches
Besides, unobserved confounders are modeled by applying a structural bias in~\cite{Discussion/NewForm/ACD_LoweMSW22}.
Several recent studies termed as causal representation learning take a new perspective on unobserved confounders.
It will be detailed in subsection (\ref{subsection:causal_rep}).

\textbf{Subsampling:} In real-world applications, temporal data, especially time series, may be sampled at a rate lower than the rate of the underlying causal process due to the difficulties in data collection.
An ordinary causal discovery algorithm for sub-sampled time series may lead to spurious causal relations and missed ones. 
Several remarks and approaches~\cite{Discussion/subsample/work1, Discussion/subsample/work2_icml_GongZSTG15, Discussion/subsample/work3_nips_rateagnostic_PlisDFC15, Discussion/subsample/uai_subsample_aggr_GongZSGT17, Discussion/subsample/work5_pgm_constraintOPT_HyttinenPJED16, Discussion/subsample/biometrika/tank2019identifiability} are proposed for this issue.

\textbf{Expert knowledge: }Expert knowledge can help the causal discovery process in practice.  
The approaches of fusing expert knowledge can be categorized into three types~\cite{intro/nonts_surveys/BN21}: (1) \textit{Soft constraints}: the learning process can be influenced by the knowledge~\cite{Discussion/knowledge/ausai/ODonnellNHKAH06}. 
(2) \textit{Hard constraints}: the learnt structure must conform to the enforced requirements (\ie, conditions given with a probability $p=0$ or $p=1$). 
In~\cite{Discussion/knowledge/artmed/AsvatourianLML20}, hard constraints are leveraged in structure learning with a time dependant exposure.
Studies in~\cite{MTS/SB/NTS_NOTEARS} add prior knowledge forbidding the existence of intra-slice dependencies, which is helpful to recover edges that are not explicitly encoded by the prior knowledge.
(3) \textit{Interactive learning}: the human input is leveraged in the learning process~\cite{Discussion/knowledge/ecsqaru/MessaoudLA09, Discussion/knowledge/kdd/MelkasSCMNMP21,https://doi.org/10.48550/arxiv.2206.05420, 9222294}.

\subsection{New perspectives}
\label{subsec:newper}

\subsubsection{Extension in amortized and supervised paradigms}

In the traditional paradigms, causal discovery methods mostly either treat observational data separately or train a distinct model for each individual. 
These methods do not make full use of the common structure across different samples or supervised information from the datasets whose causal structures are clearly explored, thus suffering from several issues such as the small sample challenge and lack the inductive capability.
Recently, causal discovery is conducted in new paradigms to solve this problem. We can roughly categorize them into two groups: methods based on \textbf{amortized modeling} \cite{MTS/Attention/icdm_InGRA_ChuWMJZY20, Discussion/NewForm/ACD_LoweMSW22}, and methods based on \textbf{supervised learning} \cite{benozzo2017supervised, wang2022meta}.
We introduce them in this subsection, which we believe are a promising avenue for future research.

In amortized modeling, a global causal discovery framework is trained for individuals with different causal structures. 
As for scenarios with temporal data, these approaches have been detailed in \ref{subsection:NN_Granger} as the deep learning extension of Granger causality with inductive modeling.
InGRA \cite{MTS/Attention/icdm_InGRA_ChuWMJZY20} leverages prototype learning to extract common causal structure while ACD \cite{Discussion/NewForm/ACD_LoweMSW22} proposes an encoder-decoder framework to conduct amortized causal discovery. These methods make full use of information from massive samples and are able to infer causal relations for newly arrived individuals, which are useful in real-world applications such as e-commerce, social network, and neuroimages.

Another line of work has predominately focused on treating the inference process as a black box and learning the mapping from sample data to causal graph structures via supervised learning. Here the label information is causal structure and can be easily accessed in synthetic datasets. 
Earlier work \cite{Discussion/NewForm/RCC/jmlr/Lopez-PazMR15, DBLP:conf/aaai/TonSF21} on learning causal relations by supervised learning is restricted to learning pairwise causal direction where the problem is cast into a classification task to distinguish between $X \to Y$ and $Y \to X$ by using observed samples.
It's later extended to discovery graph structure in \cite{Discussion/NewForm/DAG_EQ/corr/abs-2006-04697,petersen2022causal}.
As the labeled information for training is often originated from synthetic data or real-world datasets which have been explored, the requirement of a supervised approach, in which the distributions of training and test data match or highly overlap, is not guaranteed. In \cite{Discussion/NewForm/ML4S/kdd/00040DJWH022, Discussion/NewForm/CSIvA_DeepMind}, methods such as vicinal graph and meta-learning are leveraged in supervised causal discovery to tackle this `domain shift' issue.  
For the temporal setting, a supervised estimation of Granger causality between time series is proposed in \cite{benozzo2017supervised}. As a recent advance, a method for learning causal discovery is proposed in \cite{wang2022meta} where the learned from large datasets with known causal relations outperform the algorithm in the traditional paradigm when testing on temporal datasets such as fMRI. 

\subsubsection{Extension in causal representation learning}
\label{subsection:causal_rep} 

Extracting the causes of particular phenomena whether explicitly or implicitly from a deep learning black box can be beneficial to the downstream tasks.
The aforementioned causal discovery methods focus on inferring relations between observed variables, or start from the premise that the causal variables are given before hand.
Although some approaches learn causal relations under unobserved variables.
There exist real-world observations (e.g., sensor measurements, image pixels in video) which are not well structured to causal variables to begin with. 
As a generalization of causal discovery from observed variables, there has recently been a growing interest in \textbf{causal representation learning} \cite{CausalRepresentation/nontemp/icml/LocatelloPRSBT20, CausalRepresentation/nontemp/towardsCRL/ScholkopfLBKKGB21, CausalRepresentation/nontemp/CausalVAE/YangLCSHW21}, which aims at learning representation of causal factors in an underlying system.
It estimates latent causal variable graphs from observations.


A line of works in causal representation learning identifies independent factors of variations based on disentanglement and Independent Component Analysis (ICA).
At the heart of this methodology is the postulation of mutually independent latent factors.
It's hard to identify true latent variables, especially in general nonlinear cases.
As a remedy, recent approaches \cite{CausalRepresentation/nontemp/icml/LocatelloPRSBT20, CausalRepresentation/iVAE_nontemp/aistats/KhemakhemKMH20, DBLP:conf/aistats/HyvarinenM17, DBLP:conf/nips/HyvarinenM16} leverage additional information in multiple views, auxiliary variables, or temporal structure, combined with deep learning methods like VAEs and contrastive learning.
A connection between ICA and causality has been recently drawn in \cite{CausalRepresentation/IMA/nontemp/nips/GreseleKSSB21, DBLP:conf/uai/Monti0H19}.
In the context of temporal data, the identifiability of causal variables from temporal sequences is discussed in latent temporal causal process estimation (\textbf{LEAP}) \cite{Discussion/latent/iclr_LEAP_YaoSHS022}. It first provides causal identifiability conditions in a nonparametric, nonstationary setting, and a parametric setting. Then it proposes a learning framework to extract latent causal relations, which extends VAE with a learned causal process network by enforcing the assumed conditions.
The non-stationary noise, modeled by flow-based estimators, can be viewed as a soft intervention to aid identification.
In line with LEAP, subsequent works \cite{TDRL_DBLP:journals/corr/abs-2210-13647} extend the identification theory to a more general case.   

Another line of work leverage intervention and data augmentation to help to identify latent causal relations. Under data augmentation, it's demonstrated in \cite{CausalRepresentation/line2/nips/KugelgenSGBSBL21} that common contrastive learning methods can block-identify causal variables that remain unchanged. 
For the temporal setting, \textbf{CITRIS} \cite{CausalRepresentation/CITRIS/icml/LippeMLACG22} is proposed. It's a VAE framework learning causal representation where latent causal factors have possibly been interved on.
By using intervention target information for identification, CITRIS is devoid of suffering from functional or distributional form constraints.
Besides, causal factors in CITRIS are considered as either scalars or potentially multidimensional vectors, which is more practical in complex scenarios. Along this line of work, instantaneous causal relations are extracted in iCITRIS \cite{CausalRepresentation/interv/iCITRIS/abs-2206-06169}.

\section{Conclusion}\label{sec:conc}

Causal discovery in temporal data is fundamental to understanding the dynamics and estimating the causal effects of interest. 
This article reviews two categories of temporal causal discovery: multivariate time series causal discovery, and event sequence causal discovery. 
Multivariate time series causal discovery can be categorized into four groups, including constraint-based, score-based, FCM-based, and Granger causal model. 
Main ideas and recent advances for each type are reviewed.
For causal discovery in event sequence, we can classify these algorithms into constraint-based, score-based, and Granger causal models, which are in accordance with multivariate time series causal discovery. 
We note that Granger causal models are especially well-developed for event sequence due to a natural match-up between Granger causality and Hawkes processes.
To bridge the gap between abundant temporal causal discovery algorithms with real-world impacts, we introduce several major studies including scientific endeavors and industrial implementations.
We also provide an extensive list of resources, including datasets and metrics, which can be used as a guideline for future research in this field.
Whilst many algorithms are offered with theoretical or empirical guarantees, the quality of the inferred relations is dependent on many issues, including non-stationarity, heterogeneity, unobserved confounders, subsampling and expert knowledge.
We discuss these challenges and practical considerations.
Lastly, we introduce new perspectives of causal discovery, where avenues for future work in amortized modeling, supervised learning, and causal representation learning are depicted.


\begin{acks}
We thank Lun Du, Wei Chen, Jin Wang, Yongjun Xu, Fei Wang, Zezhi Shao, Yueyang Su, Yongtao Xie for valuable advice.
We thank Hao Sun for his assistance in enhancing our visualizations.
Thank anonymous readers for their letters that helped us improve our paper.
\end{acks}




\end{document}